\pdfoutput=1
\documentclass[12pt]{report}
\usepackage[utf8]{inputenc}
\usepackage{graphicx} 
\usepackage{setspace}
\usepackage[margin=1in]{geometry}
\usepackage{amsmath}
\usepackage{float}
\usepackage{sectsty}
\usepackage{microtype}
\usepackage[sort&compress,numbers]{natbib}
\usepackage[none]{hyphenat}
\usepackage{hyperref}

\begin{document}

\pagenumbering{roman}

\begin{titlepage}
  \begin{center}
    \vspace*{1cm}
    
    \LARGE
    \textbf{Improving Generalization Performance of YOLOv8 for Camera Trap Object Detection}
    
    \vspace{2cm}
    \large
    A thesis submitted to the Graduate College of the University of Cincinnati in partial fulfillment of the requirements for the degree of
    
    \vspace{0.5cm}
    \LARGE \textbf{Master of Science}\\
    \vspace{0.5cm}
    \large in the\\ 
    \vspace{0.5cm}
    \LARGE\textbf{Department of Computer Science}\\
    \vspace{0.5cm}
    \large of the \textbf{College of Engineering and Applied Science}
    
    \vspace{0.25cm}
    \large
    by
    
    \vspace{1cm}
    \LARGE \textbf{Aroj Subedi}\\
    \large Masters University Of Cincinnati\\
    \large March 2024
    
    \vspace{0.5cm}
    \large
    Committee Chair: Yizong Cheng, Ph.D.
    
    \vfill
    
    \vspace{1cm}

  \end{center}
\end{titlepage}

\setcounter{page}{2}
\chapter*{Abstract}
Camera traps have become integral tools in wildlife conservation, providing non-intrusive means to monitor and study wildlife in their natural habitats. The utilization of object detection algorithms to automate species identification from Camera Trap images is of huge importance for research and conservation purposes. However, the generalization issue where the trained model is unable to apply its learnings to a never-before-seen dataset is prevalent. This thesis explores the enhancements made to the YOLOv8 object detection algorithm to address the problem of generalization. \\

The study delves into the limitations of the baseline YOLOv8 model, emphasizing its struggles with generalization in real-world environments. To overcome these limitations, enhancements are proposed, including the incorporation of a Global Attention Mechanism (GAM) module, modified multi-scale feature fusion, and Wise Intersection over Union (WIoUv3) as a bounding box regression loss function. A thorough evaluation and ablation experiments reveal the improved model's ability to suppress the background noise, focus on object properties, and exhibit robust generalization in novel environments. The proposed enhancements not only address the challenges inherent in camera trap datasets but also pave the way for broader applicability in real-world conservation scenarios, ultimately aiding in the effective management of wildlife populations and habitats.

\newpage
\ 
\thispagestyle{empty}

\tableofcontents
\listoftables
\listoffigures

\clearpage
\pagenumbering{arabic}

\chapter{Introduction}

\section{Camera Traps}
Camera traps are innovative conservation tools, equipped with sensors and get activated automatically to capture images or videos whenever a movement is sensed in and around the natural environment where it is placed. Camera traps operate during both day and night and can be left unattended for extended periods, providing an effective means of recording wildlife in their natural environment with as little human interference as possible, minimizing disturbance to the wildlife. Camera traps are cost-effective and are routinely employed for wildlife surveys \cite{rowcliffe, Tobler} and monitoring efforts \cite{kays2009}, delivering comprehensive and continuous feed from even the most remote or challenging terrains, helping to collect data/images of nocturnal, sensitive, and rare species. \\

Camera traps are not a recent innovation but have been utilized by humans for a considerable time now. In the study conducted by Kucera et al.  \cite{kucera2011}, the evolution of camera traps, technological advancements in both early and modern eras, their scientific applications, and the conservation challenges they present were explored. Swann et al. \cite{Swann2011} discussed various types of camera traps based on image resolution, trigger speed, mode of use, and battery life, some of which are designed for particular conditions and are specialized for capturing specific target species. \\

The data and images acquired through camera traps make significant contributions to ecological research  \cite{besson, burton, Nichols2011}. They aid in comprehending animal behavior for the formulation of effective conservation strategies \cite{Caravaggi2017} and allow for the estimation of animal density without the need for laborious individual recognition \cite{Rowcliffe2008}. This information can be further analyzed to monitor individual animal movements and activities, estimate population sizes, ascertain distribution patterns, and evaluate trends in wildlife populations over time. Such insights play a crucial role in safeguarding animals and their environments from extinction or human-induced harm. The adoption of camera traps has transformed the landscape of biodiversity monitoring and wildlife management, becoming synonymous with ecological research and conservation practices. Moreover, unavoidable circumstances such as the COVID-19 pandemic, which restricted researchers' access to field sites and impeded traditional data collection and analysis processes, highlighted the indispensable need to use camera traps to track population trends, study breeding activities, understand human-wildlife interactions, and adapt management strategies to mitigate potential human-wildlife conflicts \cite{BLOUNT2021108984}.

\section{Challenges with Camera Traps dataset}\label{sec:camera-trap-challenges}

While camera traps are widely adopted for the non-invasive monitoring of species, there are some inherent limitations, challenges, and issues associated with the images captured by such traps. Villa et al. \cite{GOMEZVILLA201724} explore primary challenges encompassing cost and logistics, ethical considerations, management of substantial volumes of collected raw data, biases within the collected data, manual species identification, and the recognition and tracking of individual animals. Similarly, He et al. \cite{app13106029} analyze two forms of imbalance in camera trap datasets: class-level imbalance (numerous classes with few images) and object/box-level imbalance (numerous small objects vs. few large ones), highlighting the unreliability of evaluation metrics in imbalanced settings. Additionally, Newey et al. \cite{newey2015} underscore potential drawbacks associated with the use of more affordable, ``recreational" camera traps for data collection, outlining challenges related to remote monitoring, data retrieval, and an increase in false positives, leading to inaccurate evaluations. Furthermore, Glen et al. \cite{glen2013} discuss the challenges associated with detecting and identifying small mammals using camera traps. \\

Some of such potential challenges are discussed below: 

\begin{enumerate}
    \item \textbf{False Triggers:}\label{item:false-triggers}\\
    The camera traps are susceptible to false triggers caused by various factors, including wind or upward movement of heat from the ground, swaying vegetation, falling branches, or changes in lighting conditions. These instances may lead to the capturing of empty frames. Furthermore, Empty frames may occur when an animal leaves the camera’s field of view before the trigger sequence completes. The occurrence of false triggers contributes to an inflated dataset, presenting challenges in accurately assessing animal activity or behavior.
    
    \item \textbf{Class imbalance:}\label{item:class-imbalance}\\
    It refers to the uneven distribution of various species or categories within the images captured by the camera traps. Species that are larger in size and more frequently encountered are overrepresented, whereas those that are rare and less in numbers tend to have lower representation in the datasets.
    
    \item \textbf{Object/box-level imbalance:}\label{item:obj-box-imbalance}\\
    Camera trap images exhibit significant differences in the body size and shape between the similar animals due to the variations in the alignment, and their different distances from the camera. Additionally, animal poses might vary, including standing, moving, resting in awkward poses, or interacting, among others. The other challenges include cases where the camera traps capture only a portion of an animal, making it difficult to identify the species. 
    
    \item \textbf{Environmental variability:}\label{item:env-variability}\\
    In the captured images, backgrounds like grasslands and forests, along with common objects such as tree stumps, logs, and rocks, showcase significant diversity. Moreover, environmental conditions related to lighting, shadow casting, and seasonal differences between summer, winter, spring, etc, contribute to substantial fluctuations in the captured images. This can result in the depiction of the same animal in different forms across various images.
    
    \item \textbf{Placement of camera:}\label{item:placement-of-camera}\\
    The positioning of camera traps at varying heights from the ground introduces variations in angle and perspective, altering the visual representation of individual animals. Consequently, an individual animal in camera trap images may be presented with a confusing perspective, complicating the interpretation of its appearance. 
    
\end{enumerate}

\section{Object Detection}
Object detection is a field in computer vision and image processing that involves the identification of occurrences of semantic objects belonging to specific classes (Humans, Animals, Cars, etc) in digital images or videos. Object detection is a combination of object localization and image classification to provide a comprehensive understanding of the visual content in the images and videos. Object localization entails determining the object positions within an image, while image classification involves identifying the class labels or the categories to which the detected object belongs to. The object localization is done in the form of bounding box coordinates. Object detection has a wide range of applications, including autonomous vehicles, medical image analysis, and wildlife surveillance and monitoring among a few. \\
 
The process of object detection can be approached using deep-learning-based approaches due to their ability to automatically learn relevant features from the data. Convolutional Neural Networks (CNNs) in particular have revolutionized this field due to their ability to extract hierarchical features from the input image and due to other factors that allow them to recognize patterns regardless of their positions within the image. Some of the popular object detection methods today are Region-based Convolutional Neural Networks (R-CNN) \cite{girshick2014rich}, Faster R-CNN \cite{ren2016faster}, You Only Look Once (YOLO) \cite{bochkovskiy2020yolov4, ge2021yolox, li2022yolov6, redmon2016look, redmon2016yolo9000, redmon2018yolov3, yolov5, yolov8, wang2022yolov7}, etc. \\

Here is a brief overview of how object detection is performed using a deep neural network:
\begin{enumerate}
    \item \textbf{Dataset Preparation:}\\
    The training phase requires a labeled dataset where each image contains objects of interest with corresponding bounding box annotations and class labels.
    
    \item \textbf{Architecture Selection:}\\
    The models generally consist of a backbone network like a pre-trained Convolutional Neural Network (CNN) for feature extraction and other additional layers for object detection. The input image is passed through the architecture to extract meaningful features.
    
    \item \textbf{Model Prediction:}\\
    After the extraction of the features, the features are further processed by the additional layers to predict the object class and its location in the form of bounding box coordinates.
    
    \item \textbf{Optimization of loss function:}\\
    The model’s prediction is compared with the ground truth label/annotations to calculate the loss. Using backpropagation, the gradients are calculated which updates the weights of the CNNs to minimize the detection errors/loss and to guide the model towards accurate predictions.
    
    \item \textbf{Non-Maximum Suppression (NMS):}\\
    To eliminate redundant and overlapping bounding boxes, non-maximum suppression (NMS) is applied which selects the most confident and accurate bounding boxes. 

    \item \textbf{Inference:}\\
    Once the training process is complete, the trained model can be deployed for object detection on new and unseen data.
    
\end{enumerate}

\section{Object detection in camera trap datasets}
After the images are captured by the camera traps, the subsequent analysis of collected data requires a thorough examination to extract actionable insights and information. The conventional methods, relying on human experts for species identification, present a significant bottleneck, characterized by slowness and labor intensity. MacLeod et al. \cite{macleod2010} advocate for automating this process using technologies like Machine Learning and Artificial Intelligence to significantly accelerate and enhance the accuracy of species identification, thus reducing human errors. \\

Leorna et al. \cite{LEORNA2022101876} emphasize the potential of AI-based models like MegaDetector for wildlife detection in camera trap images and propose a synergistic approach, combining automated tools with human expertise to facilitate efficient and accurate large-scale camera trap data analysis. Additionally, Fennell et al. \cite{FENNELL2022e02104} utilized MegaDetector to detect humans and animals from images, achieving 99\% and 82\% precision, and 95\% and 92\% recall, respectively, at a confidence threshold of 90\%. Tabak et al. \cite{Tabak2022.02.07.479461} introduced an R package leveraging deep learning models for the automated analysis of camera trap images. The package offers three pre-trained models to classify different animal categories at the class, species, and family levels. Furthermore, Loos et al.  \cite{loos2018} utilized state-of-the-art deep-learning-based detectors using YOLOv2 and SSD for automatic animal detection and counting in camera trap images. \\

In the realm of species identification and counting automation, Norouzzadeh et al. \cite{Norouzzadeh2021} propose a deep active learning system that significantly reduces annotation bottleneck by 99.5\%. Similarly, Carl et al. \cite{carl2020} assessed the feasibility of using existing pre-trained deep learning models for detecting and classifying European wild mammal species in camera traps, achieving a 94\% detection rate with the pre-trained FasterRCNN + InceptionResNet V2 model. Choinski et al. \cite{choinski2021step} evaluated the effectiveness of a pre-trained deep learning model (YOLOv5) for animal detection and classification of medium-sized and large mammals in camera trap images from a European temperate forest. They achieved an average accuracy of 85\% F1-score in identifying the presence of animals and an accuracy of 78\% in classifying 12 common mammal species within the study area. Furthermore, Villa et al. \cite{GOMEZVILLA201724} conducted an evaluation of very deep convolutional neural networks (CNNs) for identifying animal species in camera trap images, achieving top-1 accuracy of 88.9\% and top-5 accuracy of 98.1\% on the evaluation set using a residual network topology.
 \\

The scope of the usage of the object detection model for the camera trap images is advantageous for several reasons:
\begin{enumerate}
    \item \textbf{Automated Data Processing:}\\
    Given the large volumes of data generated by the camera traps, manually reviewing those images requires a considerable amount of time and effort. Timely processing of this vast volume of raw data is essential to derive actionable insights for effective conservation measures and wildlife population management. The prolonged intervals between data collection, transformation into meaningful information, and the generation of usable recommendations can widen the gap between research findings and practical, timely interventions. Object detection models play a pivotal role in automating the analysis of camera trap data, offering a substantial reduction in the time and effort traditionally needed for manual review.
    
    \item \textbf{Automated annotation for species identification:}\\
    Labeling species manually from camera trap images consumes a significant amount of human hours, diverting those valuable hours from being used for other impactful tasks. Automating the annotation and labeling process not only eliminates the time-intensive nature of species identification but also frees up human resources to focus on more substantial and other high-priority tasks.
    
    \item \textbf{Time-series analysis for behavior patterns:}\\
    By automating the analysis of time-series data, object detection allows for the identification of behavioral patterns over extended periods. This helps in understanding seasonal variations, migration trends, and other long-term ecological dynamics.
    
    \item \textbf{Behavioral analysis:}\\
    Object detection models allow for tracking and analyzing the behavior of individual animals within a scene. This task when done repeatedly by a human tracker could be tiresome and may result in data inaccuracies.
    
    \item \textbf{Consistent monitoring:}\\
    Leveraging object detection models facilitates ecological studies and habitat monitoring by assessing images from different camera trap models including varying environmental conditions. This is vital for maintaining accurate records and to facilitate the comparability of data collected from various sources across different locations and time durations.
    
\end{enumerate}

\section{Challenges in camera trap object detection}
There are several challenges with the camera trap datasets as discussed in \ref{sec:camera-trap-challenges}. Using the same camera trap datasets to train the neural network presents further challenges possibly affecting the performance of the trained model. Some of them are discussed below:
\begin{enumerate}
    \item The false trigger problem discussed in \ref{sec:camera-trap-challenges}.\ref{item:false-triggers} leads to an increase in false positive detections where empty frames are identified as frames containing objects, hindering the ability of the model to accurately distinguish between relevant and irrelevant instances in the images.
    
    \item The class imbalance problem \cite{Schneider2020} discussed in \ref{sec:camera-trap-challenges}.\ref{item:class-imbalance} can lead to biased model training, where the model is better at recognizing over-represented classes and may struggle with under-represented ones.
    
    \item Object/box level imbalance discussed \ref{sec:camera-trap-challenges}.\ref{item:obj-box-imbalance} can affect the performance of the model as the model may prioritize common objects like rocks, branches, or the parts of the object instead of predicting the object of interest as a single entity.
    
    \item Environment variability discussed in \ref{sec:camera-trap-challenges}.\ref{item:env-variability} may result in the inability of the models trained on images from one season or environment to perform well on the images from another season or environment.
    
    \item The placement of the camera \cite{tanwar2021} discussed in \ref{sec:camera-trap-challenges}.\ref{item:placement-of-camera} may result in perspective distortion or misinterpretation of size and proportion making it difficult for the model to accurately assess the size and proportions of animals and might hinder the ability of the model to identify species or determine specific behaviors.
    
\end{enumerate}

\section{One-stage vs Two-stage Object Detectors}
The object detectors in use today are mainly divided into two categories based on the approaches used to detect the object of interest present in the image: one is the Two-Stage detector, and the other is the One-Stage detector. The objection detection algorithms like RCNN \cite{girshick2014rich}, Fast RCNN \cite{girshick2015fast}, Faster RCNN \cite{ren2016faster} are two-stage detectors while the Single Shot Multibox Detector (SSD) \cite{ssdLiu_2016}, RetinaNet \cite{lin2017focal}, You Only Look Once (YOLO) \cite{bochkovskiy2020yolov4, ge2021yolox, li2022yolov6, redmon2016look, redmon2016yolo9000, redmon2018yolov3, yolov5, yolov8, wang2022yolov7} are the One-Stage detectors. The two-stage object detection algorithm tackles the object detection problem using a two-pass approach. Initially, Regions of Interest (ROIs), which are regions in the image potentially containing objects, are proposed using the already established methods like Selective Search \cite{selectivesearch} or Region Proposal Networks (RPNs) \cite{ren2016faster} in the first phase. Subsequently, in the second phase, the proposed regions are further processed to achieve accurate object detection. This involves tasks such as classification and refinement using bounding box regression or binary mask predictions to enhance both the accuracy of object classification and localization. \\

However, in the one-stage detectors, the prediction of coordinates for the bounding boxes and class labels for all potential objects is done in a single pass through the neural network. Unlike two-stage detectors, there is no separate stage involving the proposal of candidate regions in one-stage detectors. As a result, one-stage detectors are faster during inference and easier to train while the two-stage detectors exhibit high robustness and have comparatively higher recognition and localization accuracy. \\

\section{Generalization in Object detection models}
In the context of object detection models, Generalization refers to the capability of the trained models to effectively apply learned patterns and knowledge acquired from the training data to novel, unseen datasets that were not part of the training or the validation set. A well-generalized model should perform reasonably well in these new conditions. This aspect is pivotal for the model's performance, particularly in real-world applications where it needs to handle diverse conditions, environments, and instances of objects. \\
 
The ability to achieve strong generalization is essential for the model to accurately detect and classify objects in real-world scenarios and locations where training data is not available. The model’s capacity to generalize across various object categories in different and varied settings enhances its robustness and is critical for its utility in practical applications. \\

\section{Problem Statement}\label{sec:problem-statement}
The current object detection algorithms exhibit limitations in their depth of generalization \cite{beery2018recognition,lee2023object,zhang2022domain}. A fundamental issue arises during the training process, where the selection of training images, validation sets, and test images stems from identical distributions within the same locations or camera traps \cite{Schneider2020}. While this practice aids the model in achieving favorable predictions on the test dataset, it inadvertently hampers the model's ability to generalize effectively when deployed in real-world settings within natural environments. The consequence is a noticeable degradation in performance, leading to unsatisfactory outcomes which highlights the poor generalization ability of models trained under such conditions. This transferability, or generalizability, problem is thought to arise because different locations have different backgrounds (the part of the picture that is not the animal) and most models evaluate the entire image, including the background \cite{beery2019efficient, norouzzadeh2017automatically}. \\

Part of the problem lies in the inherent bias introduced by the similarity in distributions across the training, validation, and test datasets as the model now becomes adept at recognizing patterns specific to the training locations. However, when employed to detect the same objects in a diverse and unseen background with lighting variations and different settings, the model struggles dearly. Consequently, the model fails to generalize its learned features effectively, resulting in suboptimal performance and diminishing its utility in real-world applications.

\section{Thesis Outline}

This thesis study is organized as follows:

\begin{enumerate}
    \item \textbf{Chapter 2} \\
    Introduces the You Only Live Once (YOLO) v8 object detection algorithm, along with a detailed discussion of the YOLOv8 model structure, visualization of feature maps, and a comprehensive discussion on the challenges and issues associated with the model.
    
    \item \textbf{Chapter 3} \\
    Explores various enhancements and adjustments made by practitioners and researchers to enhance the performance of the YOLOv8 model across diverse domains.
    
    \item \textbf{Chapter 4} \\
    Provides an in-depth discussion of the architectural modifications applied to the baseline YOLOv8s model to address the issue of generalization.

    \item \textbf{Chapter 5} \\
    Explains the selected visualization and evaluation criteria for assessing the model’s performance and conducting a comparative analysis between the baseline model and the improved model.

    \item \textbf{Chapter 6} \\
    Walks through the dataset chosen for training, validation, and testing along with the experimental setup employed for the evaluation and the results for the baseline and the improved YOLOv8s model.

    \item \textbf{Chapter 7} \\
    Summarizes the main findings and the results of the thesis study and suggests potential extensions for future research that could be further investigated or expanded upon.
\end{enumerate}
\chapter{You Only Look Once (YOLO)}
You Only Look Once (YOLO) is a popular state-of-the-art single-stage object detection algorithm, first introduced by Redmon et. al \cite{redmon2016look} in 2016. YOLO uses a single neural network to make predictions of the bounding boxes and the class category that the object belongs to directly in one forward pass. The algorithm divides the input image into grids and each grid is responsible for predicting the bounding boxes and class probabilities, providing an exceptional balance of accuracy and speed, enabling accurate and precise object detection. \\

Multiple versions and improvements to the algorithm have been introduced in different forms after it was first released in 2016. Terven et. al \cite{Terven_2023} summarize the network architecture and development of different versions of YOLO model outlining the mechanism of working, training methodology, strengths, and limitations along with future perspective. Some of the notable changes include the introduction of the convolutional layers instead of fully connected layers in YOLOv2 \cite{redmon2016yolo9000}, residual connections and feature pyramid for improved feature extraction in YOLOv3 \cite{redmon2018yolov3}, the introduction of CSPDarkNet53 (Cross-Stage Partial Connection composed of 53 convolutional layers) for the backbone in YOLOv4 \cite{bochkovskiy2020yolov4}. Similarly, the YOLOv5 \cite{yolov5} embraced component-based design and involved community-driven development along with a Spatial Pyramid Pooling Fast (SPPF) layer to process the input feature map at a variety of scales, YOLOX \cite{ge2021yolox} introduced decoupled head to perform the classification and regression task independently, YOLOv6 \cite{li2022yolov6} introduced new classification and regression losses and the YOLOv7 \cite{wang2022yolov7} boasts enhanced performance based on the Extended Efficient Layer Aggregation Network (E-ELAN) which enhances gradient flow and improves feature extraction process.\\

The most recent version of the YOLO is YOLOv8 \cite{yolov8} introduced by the ultralytics company in January 2023. Along with object detection, the YOLOv8 supports multiple other computer vision tasks like Object Classification, Object Tracking, Pose Estimation, and Instance Segmentation. YOLOv8 incorporates advancements from YOLOv5 while also benefiting from the strengths found in other state-of-the-art models within the YOLO family. There are five versions of YOLOv8 available: YOLOv8n (nano), YOLOv8s (small), YOLOv8m (medium), YOLOv8l (large), and YOLOv8x (extra-large). Each version has a different purpose and trade-offs in terms of speed, accuracy, and resource utilization with the nano version being the smallest in terms of the number of parameters and fastest while the extra-large version is the most accurate but slowest during training and inference.\\

This thesis involves utilization of the compact iteration of YOLOv8, namely YOLOv8s, which is a scaled-down version to strike the balance between speed, accuracy, and resource utilization.

\section{YOLO Model Structure}

\begin{figure}[h]
    \centering
    \includegraphics[scale=0.30]{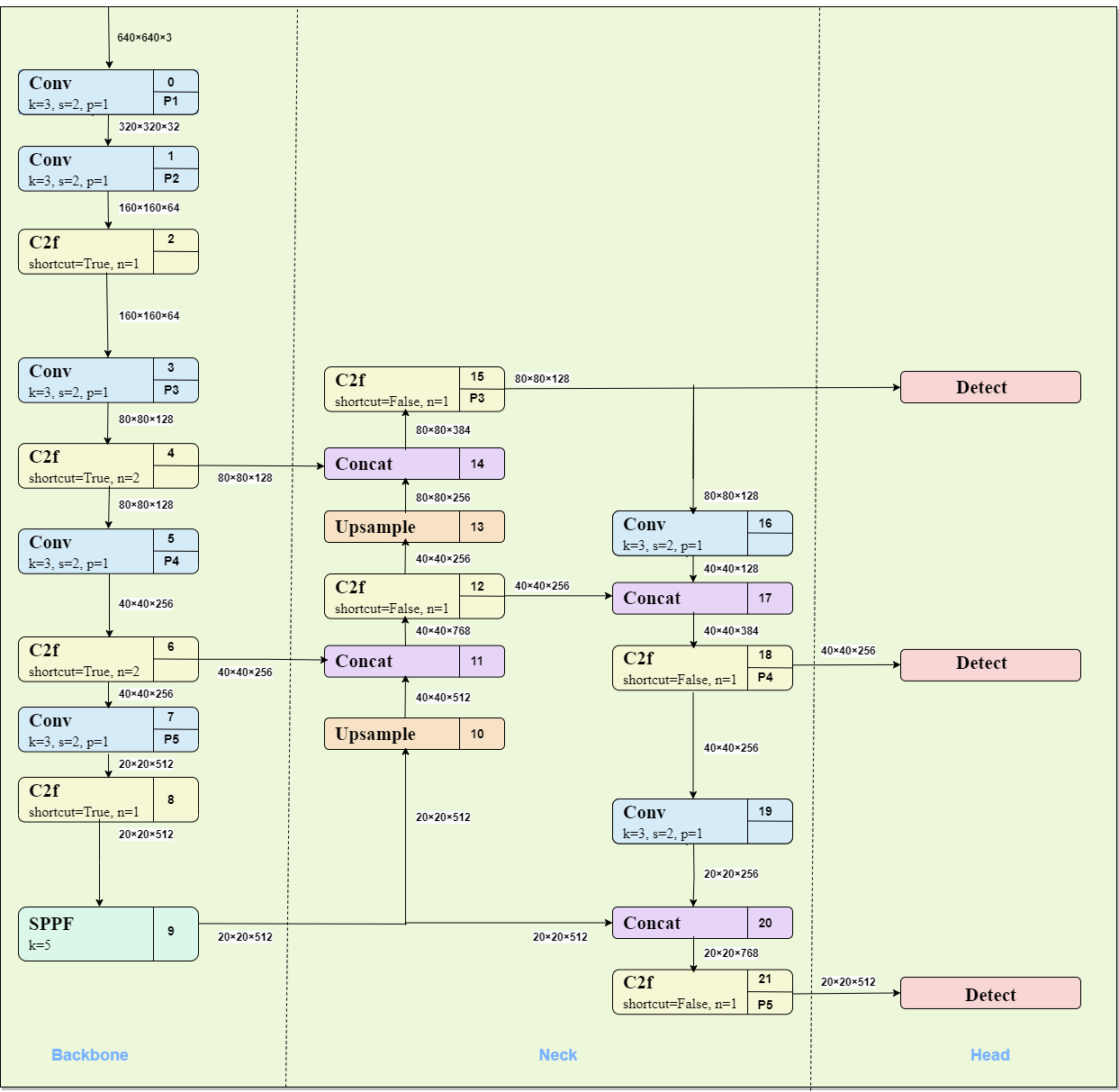}
    \caption{YOLOv8s Model Structure. Diagram inspiration taken from Github User RangeKing. https://github.com/ultralytics/ultralytics/issues/189}
    \label{fig:yolov8s-model-structure}
\end{figure}

The YOLOv8 model structure is divided into three parts: Backbone, Neck, and Head.

\subsection{Backbone}
The backbone serves as a feature extractor. The backbone consists of a Convolutional (Conv) block positioned at layers 0, 1, 3, 5, and 7. Additionally, it incorporates a C2f (Cross-Stage Partial Bottleneck with 2 Convolutions) block situated at layers 2, 4, 6 and 8, and a SPPF (Spatial Pyramid Pooling Fast) block at the end in layer 9.

\subsubsection{Convolution Block}
Convolution block in YOLOv8 is a standard 2D convolution operation that extracts features such as edges, textures, shapes, and patterns from the input image. The features are extracted hierarchically in the backbone network. The initial convolution block situated at layers 0, 1, and 3 focuses on fetching low-level features like edges and textures. As the information flows deeper into the network, the convolution block positioned at layers 5 and 7 focuses on fetching high-level features like shapes and patterns. The convolution block has different kernels (filters/feature extractors) of size 3x3 which convolve/move across the image and perform dot operations to generate output called feature maps. The number of filters in the convolutional block varies according to the layer it is present in the network. Some examples of filters are edge detection filters, sharpen filters, box blur filters, Gaussian blur filters, etc.\\

Each convolution block uses kernels (k) of size 3x3, padding (p) of 1, and stride of 2.\\
The output dimension is thus given by:

\begin{equation}
    \text{Output Dimension} = \frac{\text{Input Size} - \text{Kernel Size} + 2 \times \text{Padding}}{\text{Stride}} + 1\label{eq:output-dimension}
\end{equation}

The input and output dimensions details along with the visualization of the feature maps for each convolutional block generated for one of the dataset image are given below.

\begin{figure}[ht]
    \centering
    \includegraphics[scale=0.30]{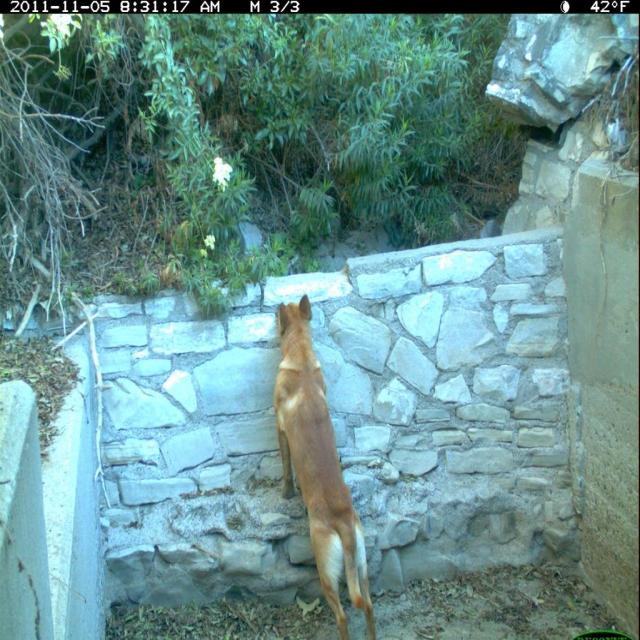}
    \caption{Dataset image used for the visualization of the feature maps.}
    \label{fig:dataset-image}
\end{figure}

For the visualization of all the feature maps below, the viridis colormap has been used which maps data values to colors in visual representation where the colormap ranges from blue/deep purple for low values to shades of yellow/green/bright yellow for high values.

\paragraph{Conv Block at Layer 0} \mbox{}\\
Input dimension:      640 x 640 x 3 [3 {=} Channels i.e Red, Green, and Blue] \\
Output dimension:     320 x 320 x 32 \\
Output Channels/Feature Maps: 32

\begin{figure}[H]
    \centering
    \includegraphics[scale=0.35]{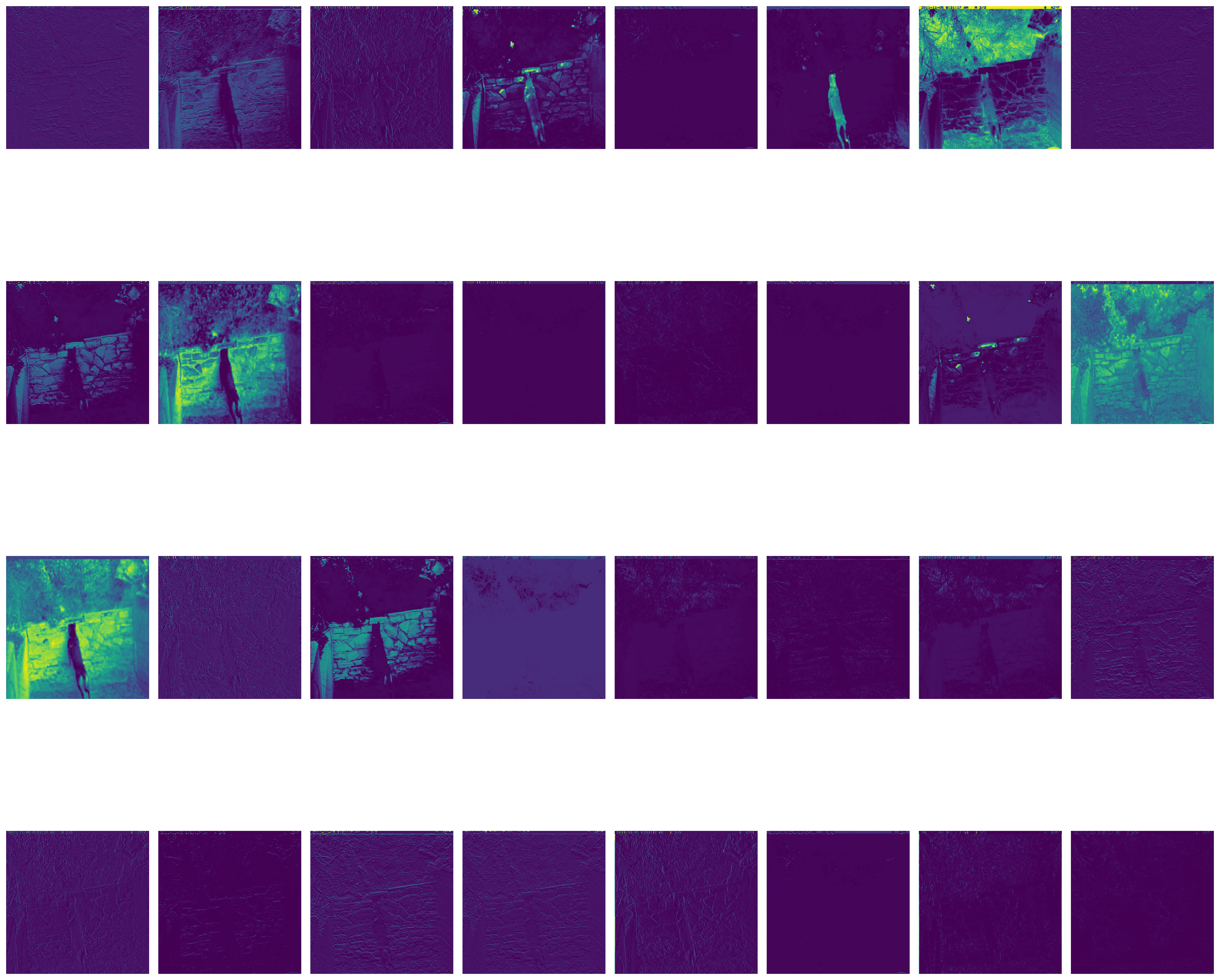}
    \caption{Visualization of feature maps produced by Conv Block at Layer 0}
    \label{fig:stage0_Conv_features}
\end{figure}

\paragraph{Conv Block at Layer 1} \mbox{}\\
Input dimension: 		320 x 320 x 32 \\
Output dimension: 		160 x 160 x 64 \\
Output Channels/Feature Maps: 64 \\

\begin{figure}[H]
    \centering
    \includegraphics[scale=0.34]{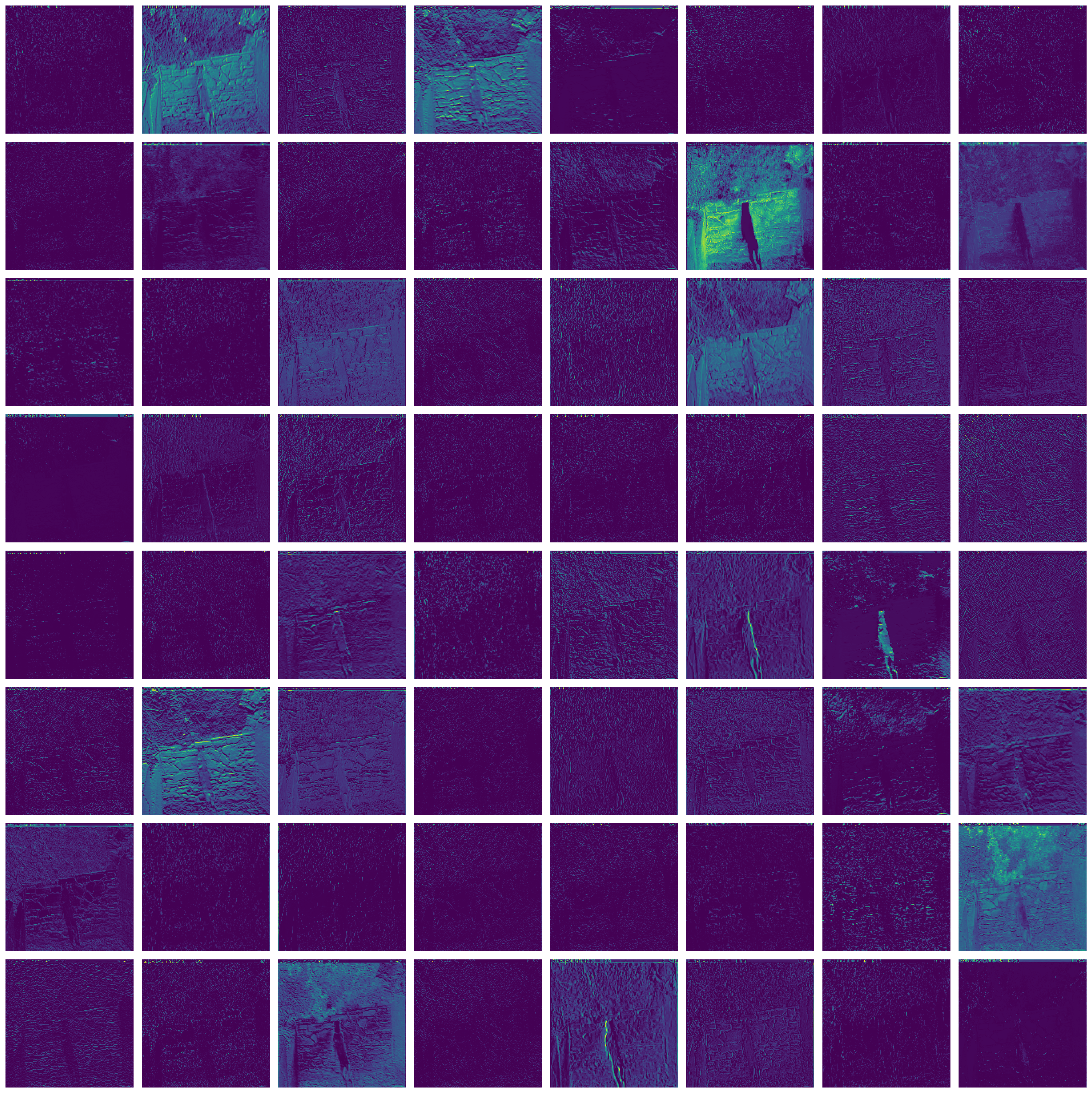}
    \caption{Visualization of feature maps produced by Conv Block at Layer 1}
    \label{fig:stage1_Conv_features}
\end{figure}

\paragraph{Conv Block at Layer 3} \mbox{}\\
Input dimension: 		160 x 160 x 64 \\
Output dimension: 		80 x 80 x 128 \\
Output Channels/Feature Maps: 128 \\

\begin{figure}[H]
    \centering
    \includegraphics[scale=0.34]{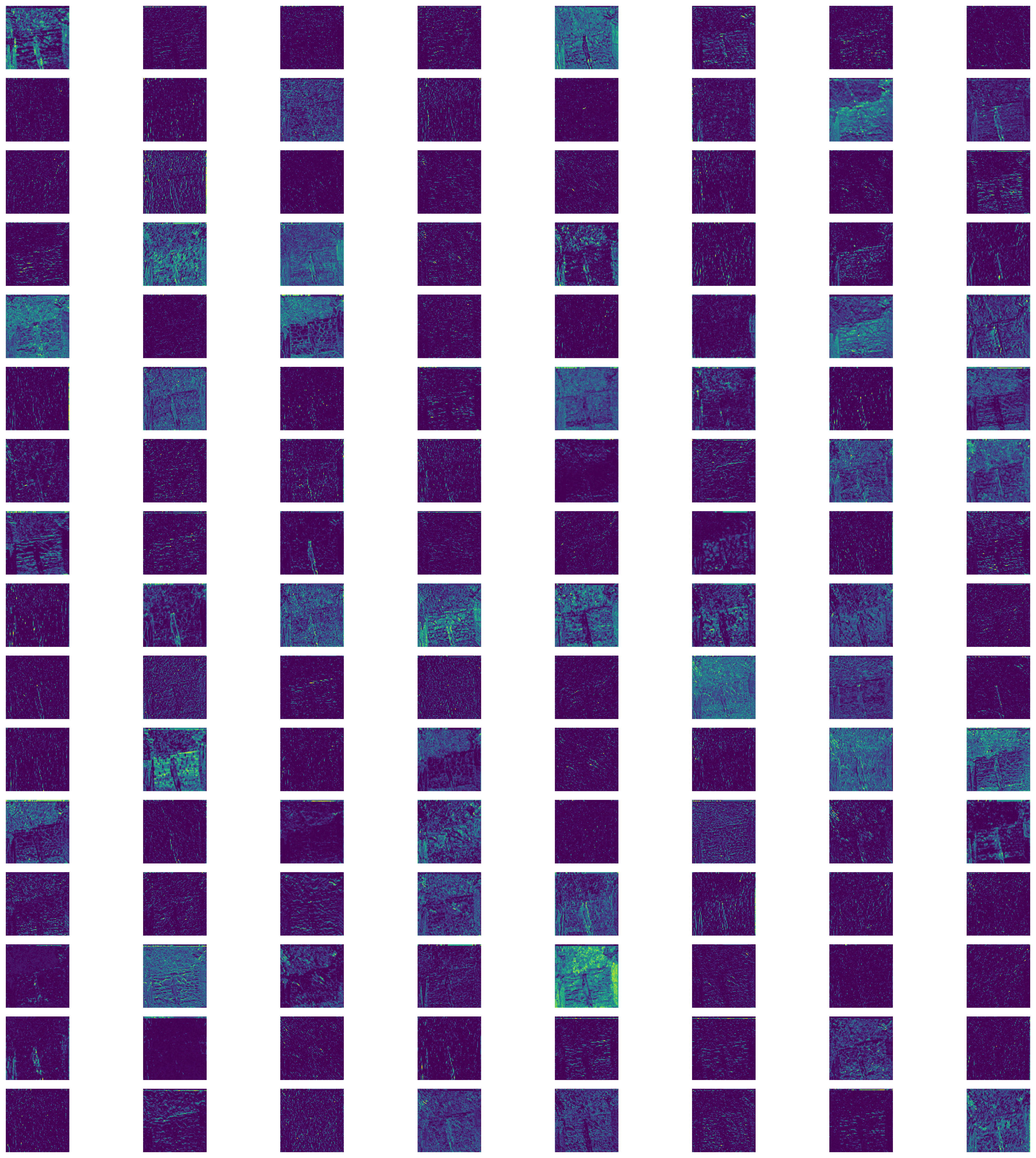}
    \caption{Visualization of feature maps produced by Conv Block at Layer 3}
    \label{fig:stage3_Conv_features}
\end{figure}

\paragraph{Conv Block at Layer 5} \mbox{}\\
Input dimension: 		80 x 80 x 128 \\
Output dimension: 		40 x 40 x 256 \\
Output Channels/Feature Maps: 256 \\

\begin{figure}[H]
    \centering
    \includegraphics[scale=0.34]{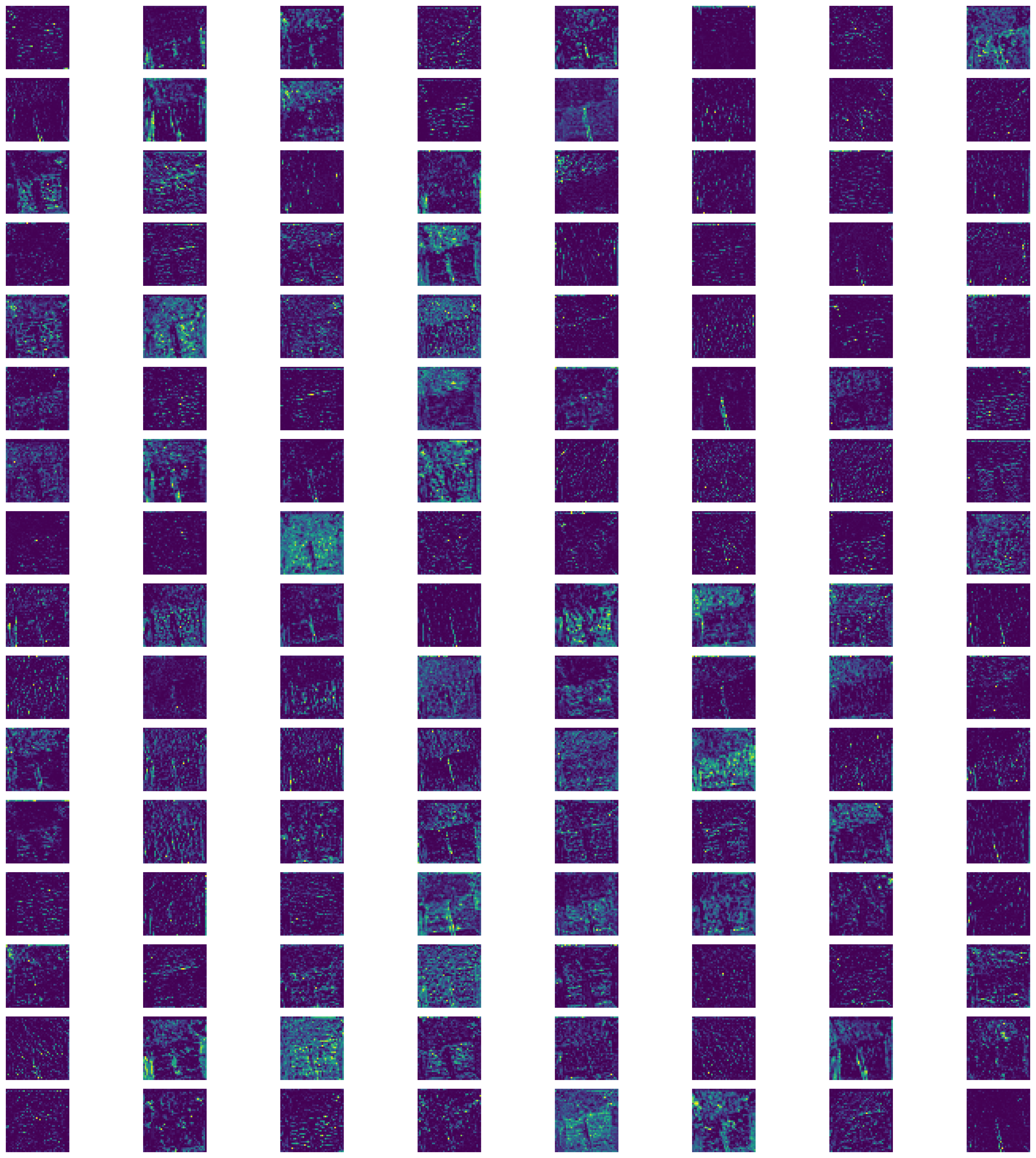}
    \caption{Visualization of feature maps produced by Conv Block at Layer 5}
    \label{fig:stage5_Conv_features}
\end{figure}

\paragraph{Conv Block at Layer 7} \mbox{}\\
Input dimension: 		40 x 40 x 256 \\
Output dimension: 		20 x 20 x 512 \\
Output Channels/Feature Maps: 512 \\

\begin{figure}[H]
    \centering
    \includegraphics[scale=0.34]{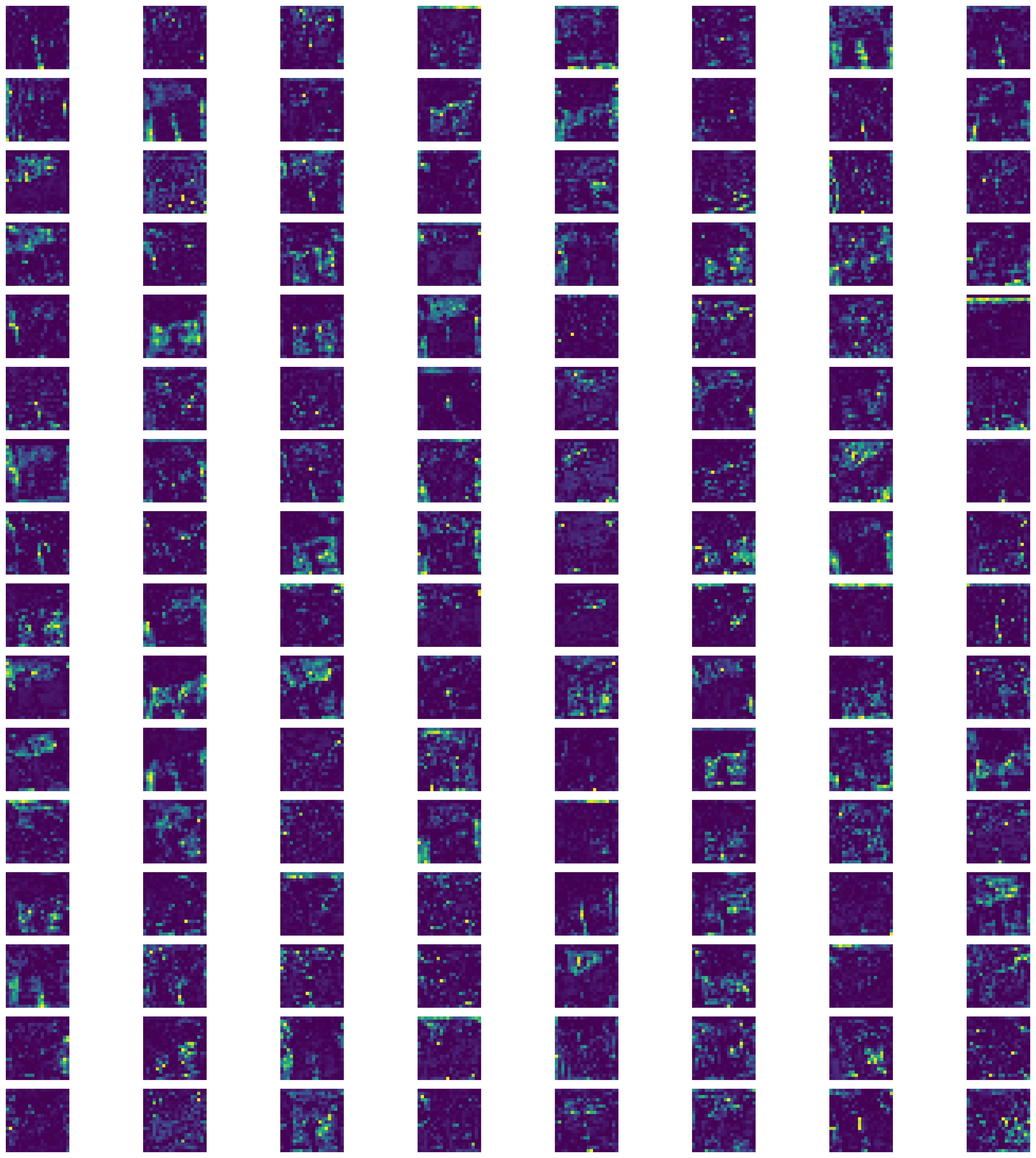}
    \caption{Visualization of feature maps produced by Conv Block at Layer 7}
    \label{fig:stage7_Conv_features}
\end{figure}

\subsubsection{C2f Block}
The C2f block (Cross-Stage Partial Bottleneck \cite{wang2019cspnet} with 2 Convolutions) combines the features extracted from the convolution block with contextual information to improve detection accuracy. It aggregates feature maps from different levels with different scales to enhance the feature representation capabilities allowing the model to be more robust.

\begin{figure}[H]
    \centering
    \includegraphics[scale=0.46]{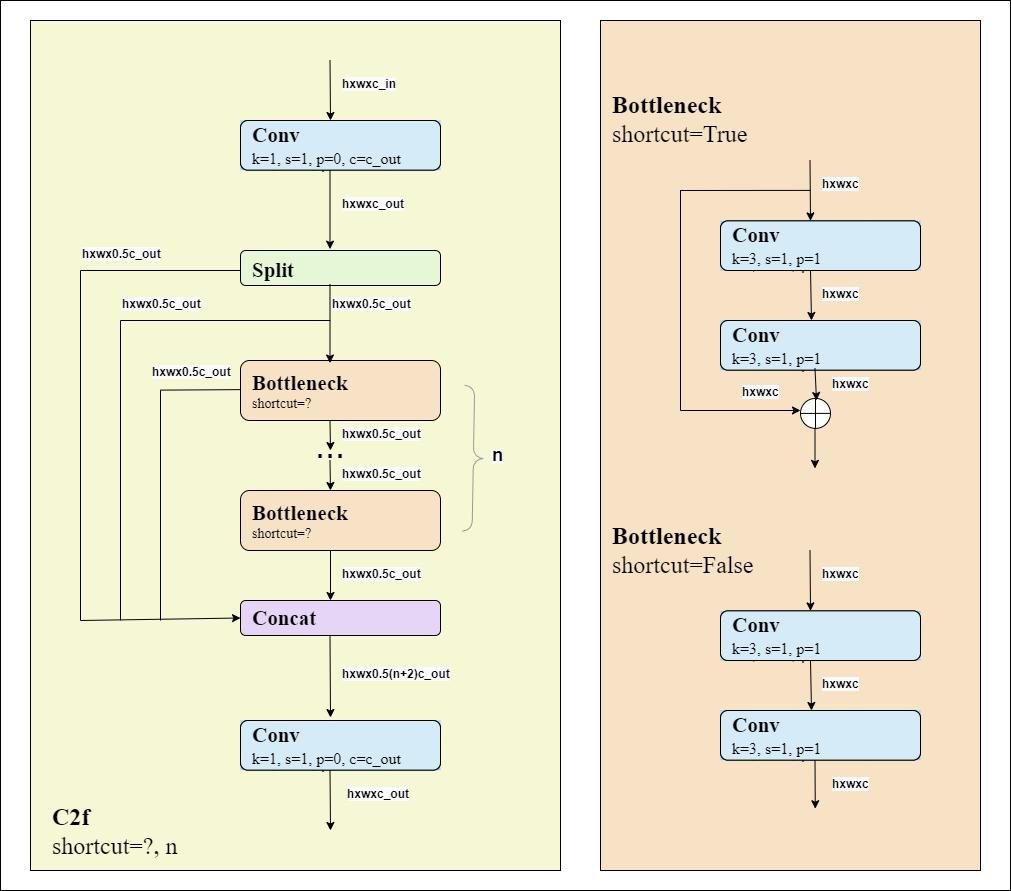}
    \caption{C2f Block}
    \label{fig:c2f_block}
\end{figure}

\textit{Cross-Stage} means splitting feature maps into different paths, two in this case, and processing them independently in the different stages to maintain feature diversity.

 \textit{Partial} indicates that only a segment of the input feature map undergoes processing within the complex dense block. This approach reduces computation costs, rendering the model to be fast and efficient. Combining the low-level information preserved by the simpler transitional layer with the high-level features extracted by the dense block results in a richer and more comprehensive feature representation.\\

\textit{Bottleneck} is a stack of convolutional blocks with residual connections, a shortcut path to allow the free flow of information, combining input and output using element-wise addition to learn more efficient representation. This helps preserve gradients and features, enabling the network to learn more complex patterns and promoting efficient learning while doing so.\\

The input, and output dimensions details along with the visualization of the feature maps for the same image for the C2f block at each layer are given below.

\paragraph{C2f Block at Layer 2} \mbox{}\\
Input dimension:      160 x 160 x 64 \\
Output dimension:     160 x 160 x 64 \\
Output Channels/Feature Maps: 64

\begin{figure}[H]
    \centering
    \includegraphics[scale=0.29]{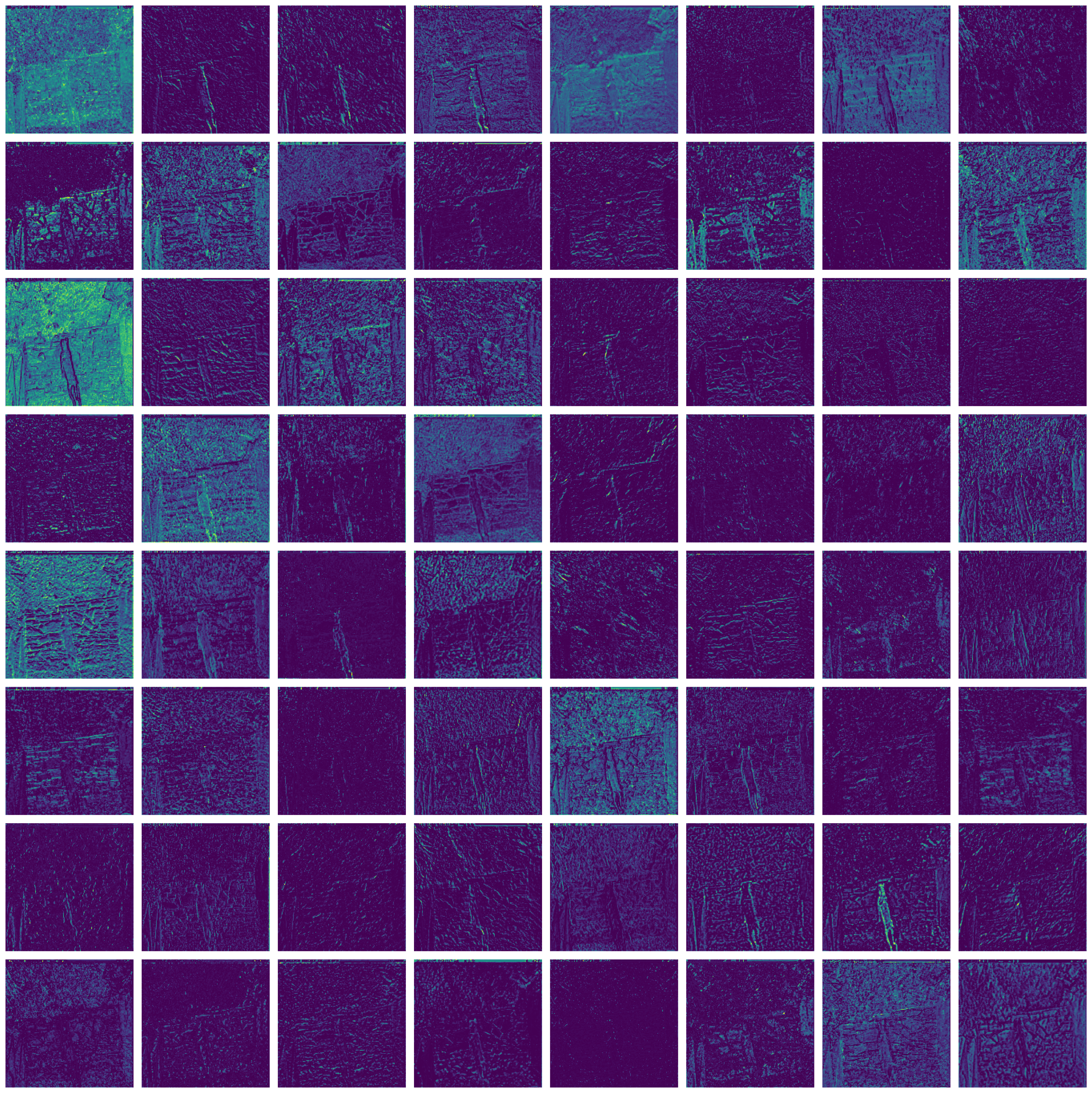}
    \caption{Visualization of feature maps produced by C2f Block at Layer 2}
    \label{fig:stage2_C2f_features}
\end{figure}

\paragraph{C2f Block at Layer 4} \mbox{}\\
Input dimension:      80 x 80 x 128 \\
Output dimension:     80 x 80 x 128 \\
Output Channels/Feature Maps: 128

\begin{figure}[H]
    \centering
    \includegraphics[scale=0.34]{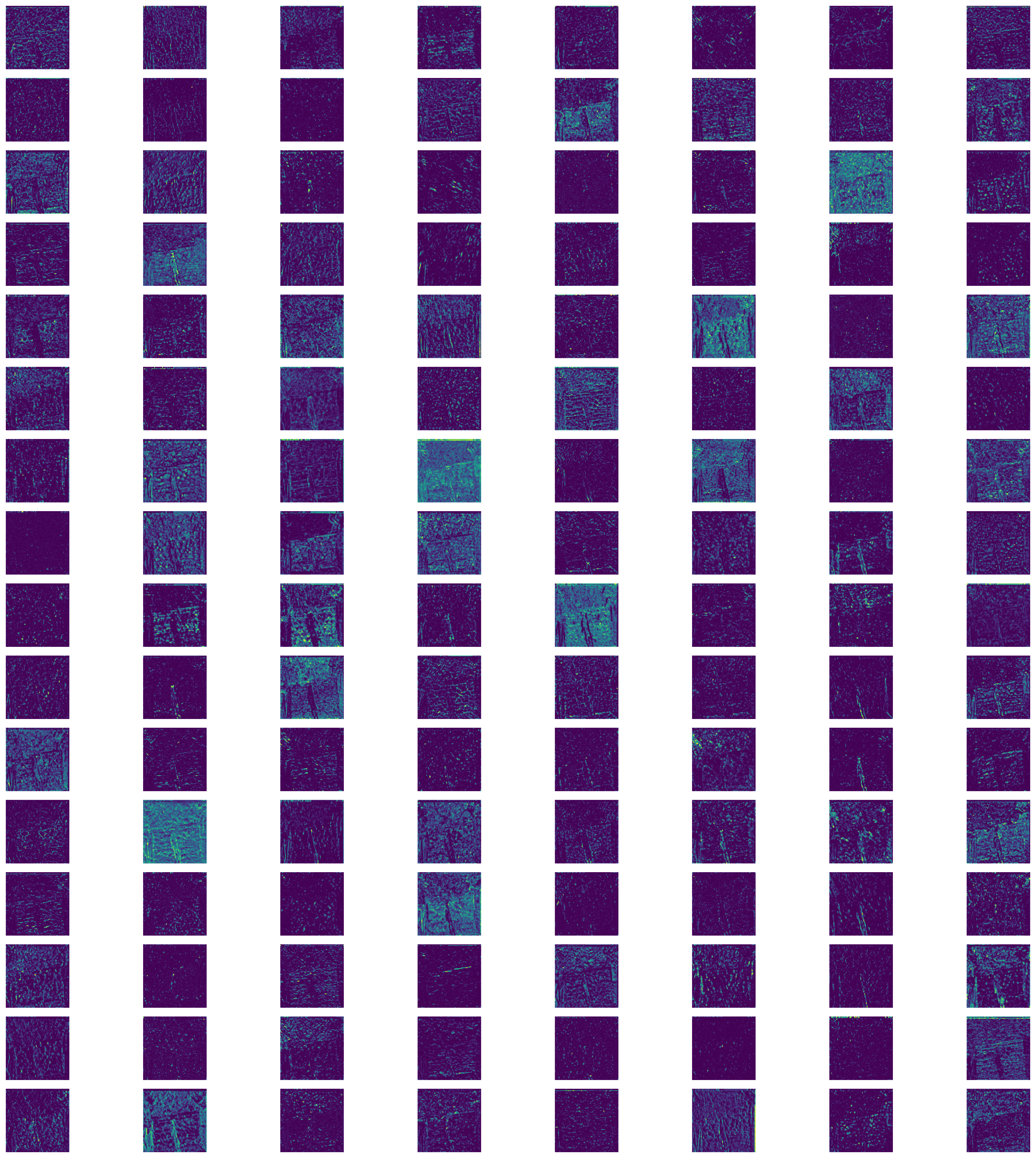}
    \caption{Visualization of feature maps produced by C2f Block at Layer 4}
    \label{fig:stage4_Conv_features}
\end{figure}

\paragraph{C2f Block at Layer 6} \mbox{}\\
Input dimension:      40 x 40 x 256 \\
Output dimension:     40 x 40 x 256 \\
Output Channels/Feature Maps: 256

\begin{figure}[H]
    \centering
    \includegraphics[scale=0.34]{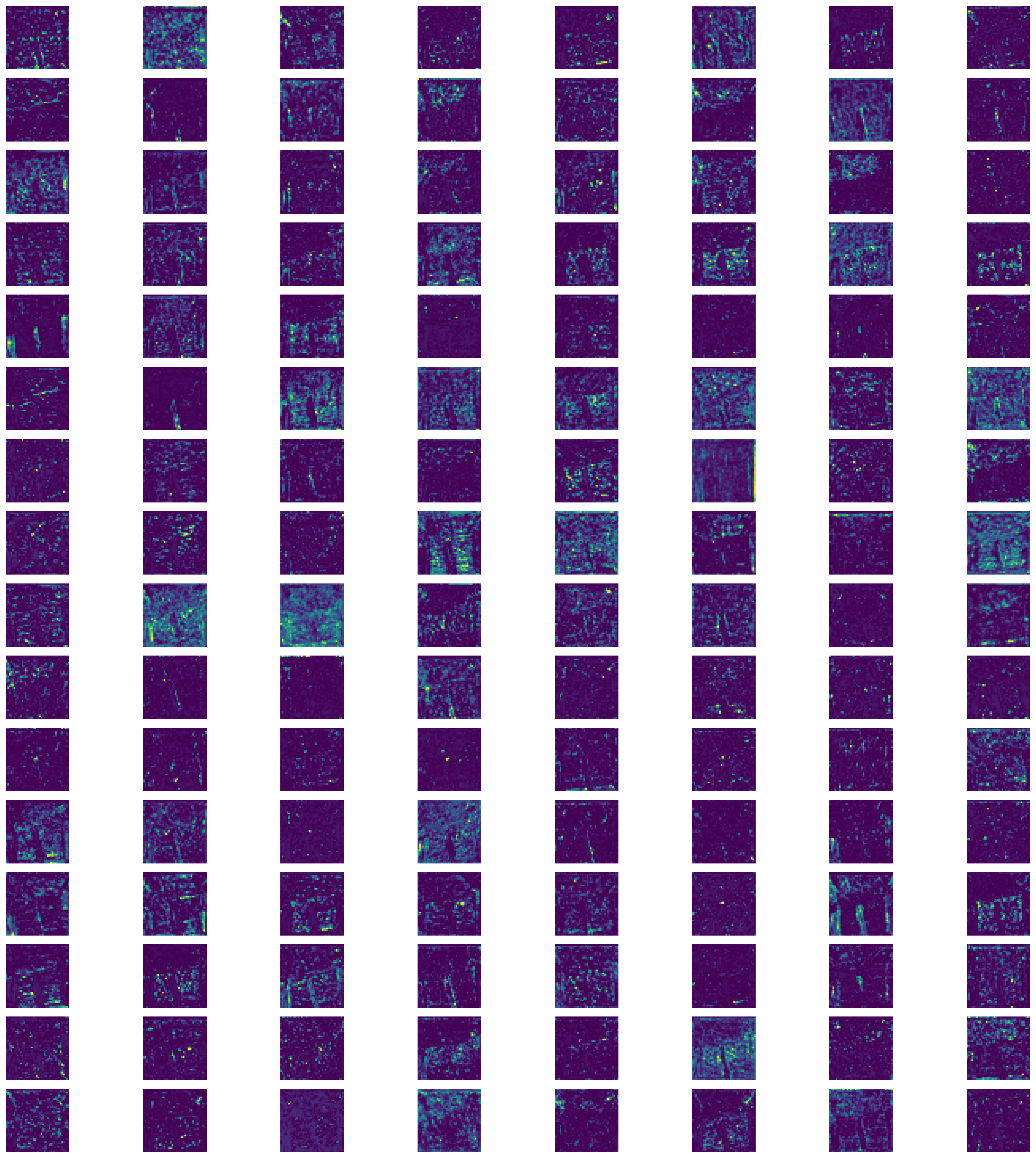}
    \caption{Visualization of feature maps produced by C2f Block at Layer 6}
    \label{fig:stage6_C2f_features}
\end{figure}

\paragraph{C2f Block at Layer 8} \mbox{}\\
Input dimension:      20 x 20 x 512 \\
Output dimension:     20 x 20 x 512 \\
Output Channels/Feature Maps: 512

\begin{figure}[H]
    \centering
    \includegraphics[scale=0.34]{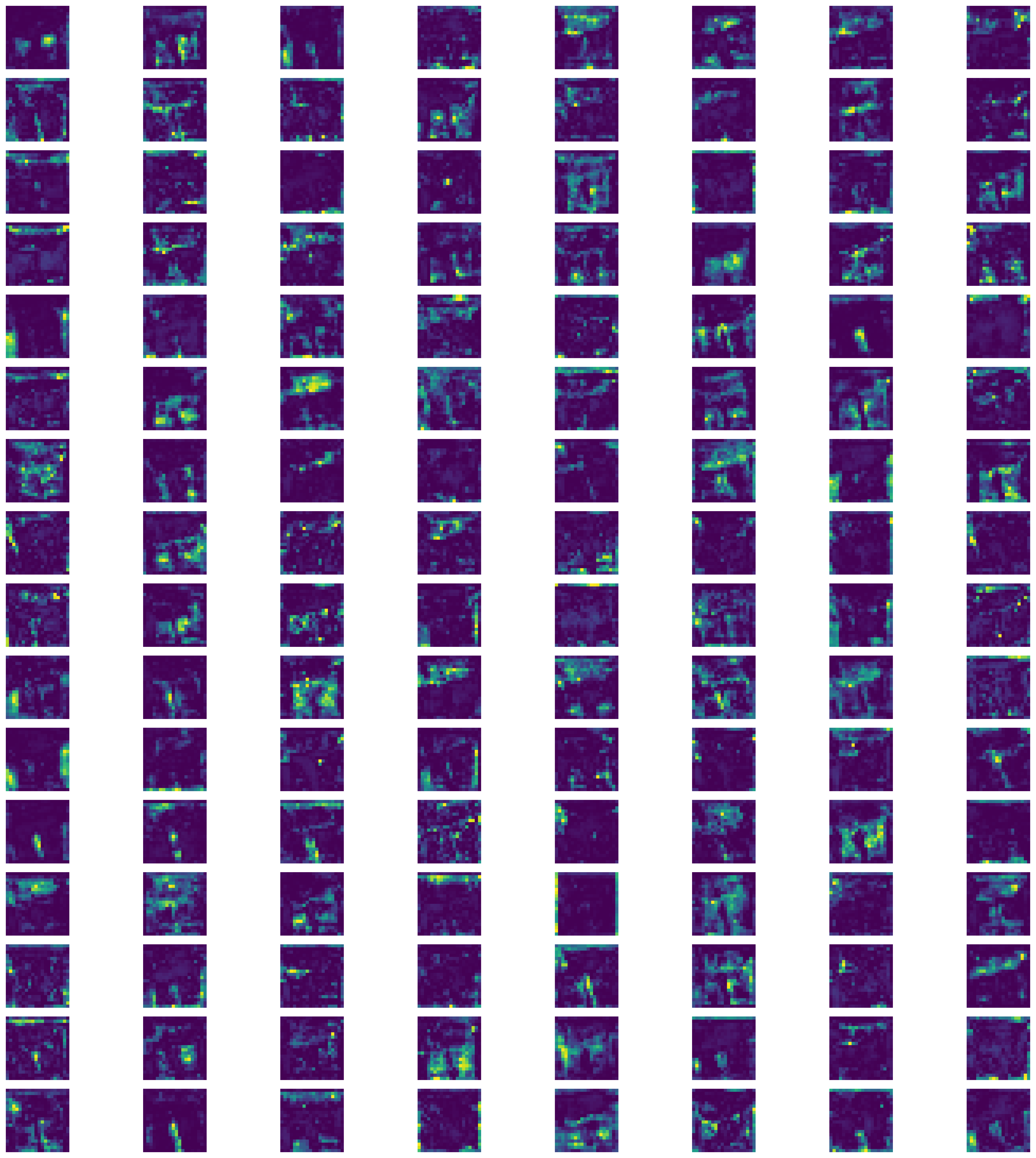}
    \caption{Visualization of feature maps produced by C2f Block at Layer 8}
    \label{fig:stage8_C2f_features}
\end{figure}

\subsubsection{SPPF Block}
The Spatial Pyramid Pooling Fast (SPPF) processes the input feature maps at a variety of scales capturing distinct levels of details. The objective of this block is to enhance the model’s ability to detect objects by performing multi-scale feature extraction and fusion to have a richer representation of feature maps. The SPPF block applies the MaxPooling operation (a downsampling operation where maximum intensity (high pixel) values are picked effectively highlighting prominent features) with kernels (k) of size 5x5 and padding (p) of size k//2, a total of three times to generate feature maps. The features are captured at different scales: The first MaxPool operation captures fine-grained details, the second MaxPool operation focuses on the mid-level features, and the third MaxPool operation effectively captures the global structures/patterns. Each MaxPool operation working in tandem zooms out, capturing a large receptive field and helping in summarizing key information. The output feature maps are then concatenated channel-wise and undergoes convolution operation which refines the concatenated features to generate a comprehensive and refined feature map.

\begin{figure}[H]
    \centering
    \includegraphics[scale=0.48]{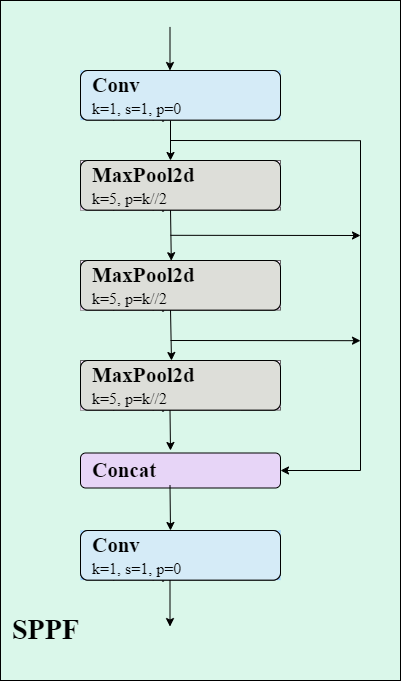}
    \caption{SPPF Block}
    \label{fig:sppf_block}
\end{figure}

During MaxPooling, Kernels (k) of size 5x5 and padding of size k//2 is used.\\
From equation \ref{eq:output-dimension}, \\
Input Dimension = 20x20x512 \\ \\
Output Dimension from 1st MaxPool = 20x20x512 \\
Output Dimension from 2nd MaxPool = 20x20x512 \\
Output Dimension from 3rd MaxPool = 20x20x512 \\ \\
After concatenation, \\
= 20x20x(4x512) \\
= 20x20x2048\\ \\
After the convolutional block, the final output is:
= 20x20x512 \\ \\
Output Dimension from SPPF block = 20x20x512 \\

The visualization of the feature maps for the same input image as the feature maps flows down the network and passes through the SPPF block is:

\begin{figure}[H]
    \centering
    \includegraphics[scale=0.36]{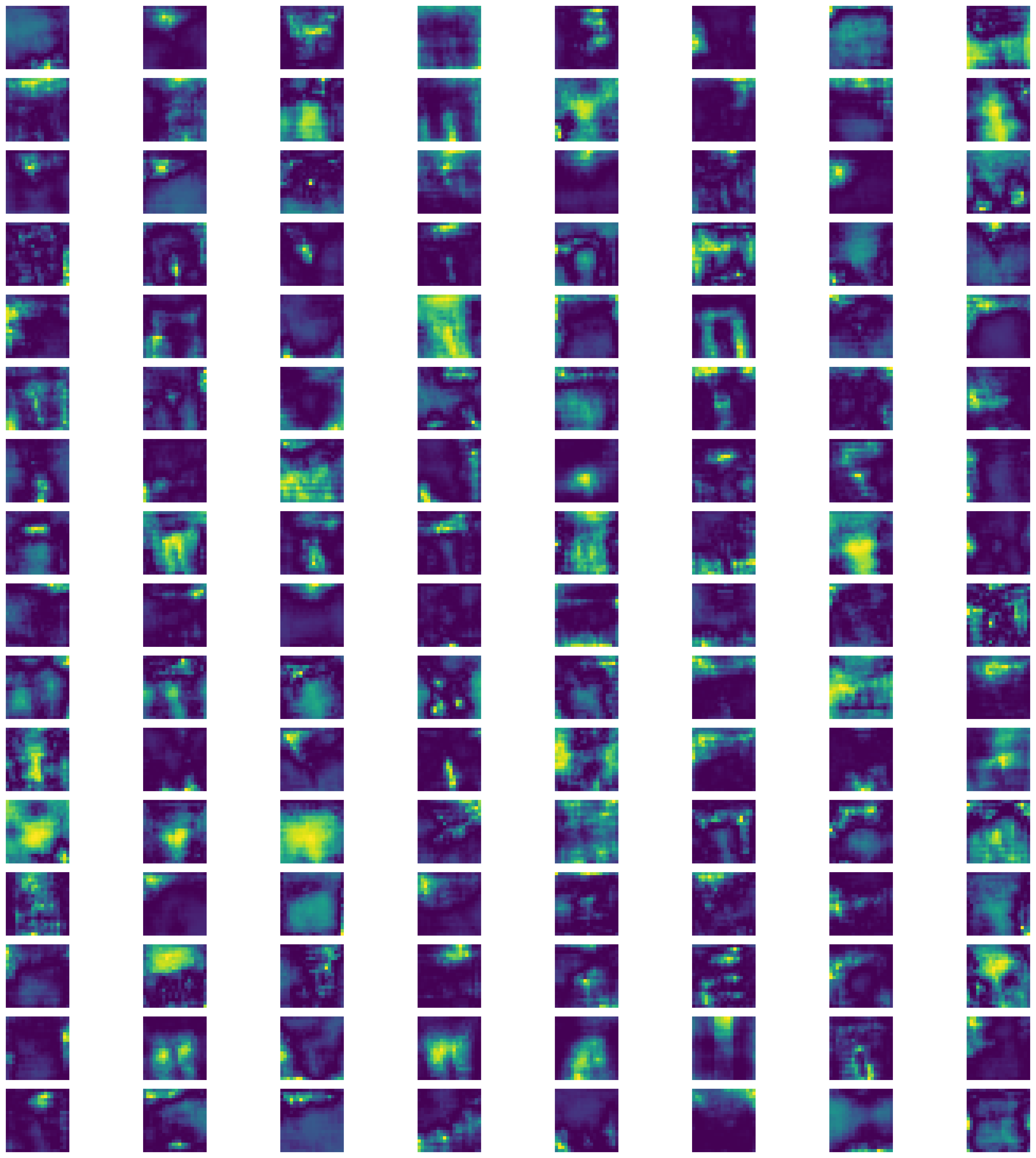}
    \caption{Visualization of feature maps produced by SPPF Block at Layer 9}
    \label{fig:stage9_sppf_features}
\end{figure}

\subsection{Neck}
The neck combines low-level features like edges and textures and high-level features like object shapes and patterns extracted from different layers of the backbone using upsampling and downsampling to facilitate the transfer of semantic and localization features. Additionally, it enhances the model’s feature fusion capability for the representation of diverse feature maps. Neck also allows the information to flow more freely across the network, leading to improved feature representation, enhanced robustness, and better context understanding.

\subsubsection {Upsample Block}
The upsample block increases the spatial resolution of the feature maps by utilizing the Nearest-neighbor algorithm with a scale factor of 2. \\
For example,
For a 2x2 matrix, \\
\[
\begin{bmatrix}
    1 & 2 \\
    3 & 4 \\
\end{bmatrix}
\]

Upsampling with the nearest Nearest-neighbor algorithm with a scale factor of 2 results in the 4x4 matrix given below:

\[
\begin{bmatrix}
    1 & 1 & 2 & 2 \\
    1 & 1 & 2 & 2 \\
    3 & 3 & 4 & 4 \\
    3 & 3 & 4 & 4 \\
\end{bmatrix}
\]

The upsampling operation is performed to recover the spatial details that may have been lost during the downsampling operation.

\subsubsection {Concat Block}
The Concat block in the neck of the YOLOv8 is responsible for integrating feature maps from different layers with different levels of abstraction to increase feature diversity and enhance the ability of the model to detect objects of various sizes. Concat block plays a crucial role in the multi-scale feature representation and also in the Feature Pyramid construction.

\subsection{Head}
The head of the YOLOv8 consists of the detect module responsible for predicting object bounding boxes and object classes. The YOLOv8 head is decoupled, signifying that the regression and classification tasks are executed independently of each other. The regression branch focuses on predicting the location and size of the detected object in the image in the form of bounding box coordinates while the classification branch focuses on identifying the class label or object category that the detected object belongs to.

\begin{figure}[H]
    \centering
    \includegraphics[scale=0.35]{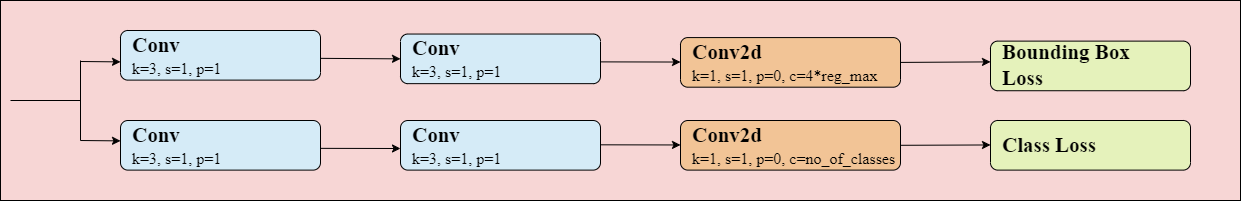}
    \caption{Detect Module}
    \label{fig:detect_module}
\end{figure}

Both the regression branch and the classification branch in the Detect module have a series of Convolution blocks to perform convolution operations for further processing and refining the feature maps received from the neck where the convolution block operates to reduce the spatial dimension, allowing the following layers to capture more abstract and task-specific information essential for bounding box prediction in case of regression branch and object class prediction in case of classification branch. \\

The Conv2d layer in the regression branch is configured to output the necessary information for generating the precise bounding box predictions for detected objects while the Conv2d layer in the classification branch is configured to output the class/object category that the object might belong to. There is a loss function associated with both the branches that guides the model towards accurate bounding box and class label predictions. \\

The bounding box loss determines the localization performance of the model while the class loss determines the classification performance of the model. In YOLOv8, Complete-Intersection Over Union (CIoU) \cite{zheng2019distanceiou} and Distribution Focal Loss (DFL) is used for the bounding box loss while Binary Cross Entropy (BCE) loss is used for the class loss. During backpropagation, the gradients from the bounding box and the class are calculated and used to update the weights of the convolutional block, leading to gradual improvement in the prediction accuracy.\\

\begin{description}
\item\textbf{Issues with the current model}\\
Firstly, During the downsampling by the Conv block in the backbone, several important features related to the detections of small object and the object localizations tend to get lost. Secondly, the model tends to focus on the features related to the background as well during object predictions. Lastly, the optimization process is influenced by numerous bounding boxes, both with small and significant overlaps, leading to the suppression of gradients generated by ordinary/regular bounding boxes. This interference adversely affects the overall performance of the model.
\end{description}
\chapter{Related Works}

Since the initial release of YOLOv8 in January 2023, various adaptations to the model structure have been proposed to enhance its performance and speed up training for specific tasks. The YOLOv8 baseline model has demonstrated superior performance, achieving a mAP50-95 value of 53.9 on the MS COCO dataset \cite{lin2015microsoft} and 36.3 on the Open Image V7 dataset for the YOLOv8x model. \\

Researchers have made significant improvements to the YOLOv8 baseline structure from different perspectives, leading to notable advancements. One such perspective is the use of alternative convolutional designs in the backbone layer to facilitate the extraction of more richer and discriminant features from the input image. Liu et al. \cite{liu_dsw_2023} focused on underwater object detection for search and rescue, replacing C2f modules in the backbone with Deformable Convnets v2 modules to adapt better to object deformations. This modification resulted in an impressive mAP50 and mAP50-95 of 91.8\% and 55.9\%, outperforming the baseline YOLOv8n model which had mAP50 of 88.6\% and mAP50-95 of 51.8\%. In a distinct domain, Yang et al. \cite{yang_tomato} addressed low automation in tomato harvesting by incorporating DepthWise Separable Convolution (DSConv), Dual-Path Gate Module (DPAG), and Feature Enhancement Module (FEM), achieving a 1.5\% improvement in mAP and reached 93.4\% mAP. Additionally, Ma et al. \cite{ma_remote_sensing} proposed using SPD-Conv modules and SPANet path aggregation networks to enhance feature extraction and fusion for tiny object recognition, resulting in a substantial improvement of mAP50 by 4.9\% and 9.1\% and mAP50-95 by 3.4\% and 3.2\% respectively for AI-TOD and TinyPerson datasets. Lou et al. \cite{lou2023dcyolov8} modified the baseline YOLOv8s model, achieving improvements of 2.5\%, 1.9\%, and 2.1\% in mAP, precision, and recall, respectively, in the VisDrone dataset by replacing C2f module with a stack of Depthwise separable convolution and ordinary convolutions. Furthermore, Ling et al. \cite{ling_pcb} introduced the C2Focal module, lightweight ghost convolution in the place of ordinary convolutions, and a new bounding box regression loss function in the form of Sig-IoU loss to replace the CIoU loss function and achieved the highest mAP50 score of 87.7\% compared to the 85.8\% acquired in the baseline YOLOv8n model. \\

An inherent challenge in object detection is the variability in the appearance of the object of interest across different scales. Recognizing smaller objects demands detailed information, whereas accurately localizing larger objects necessitates a broader contextual understanding. Multi-scale feature fusion seeks to tackle this challenge by integrating both localization details and semantic understanding. This approach holds considerable potential in enhancing the object detection capabilities of the YOLOv8 model. Notably, Huang et al. \cite{huang2023research} employed an asymptotic feature pyramid network (AFPN) to modify the feature fusion stage in the neck of the YOLOv8 model, resulting in improvements of 3.31\% in mAP and 3.59\% in recall. Similarly, Han et al. \cite{han_camouflaged} introduced an enhancement module with multiple asymmetric convolution branches to amplify multi-scale feature fusion, achieving detection performance enhancements of 8.3\% and 9.1\% on COD10K and CAMO datasets using YOLOv8 algorithm. Li et al. \cite{li_aerial} proposed the Bi-PAN-FPN architecture in the neck of the YOLOv8 model to fuse shallow and deep features, optimizing the backbone network with Ghostblock units replacing Convolutional layer and WIoUv3 as bounding box regression loss function, resulting in a mAP improvement of 9.06\%.  In a different context, Wang et al. \cite{wang_aerial} incorporated a FasterNet block, dynamic sparse attention mechanism BiFormer, and WIoUv3 as the bounding box regression loss function into the YOLOv8's model structure, achieving a mean detection accuracy improvement of 7.7\%.\\

Incorporating attention modules presents a promising strategy for addressing limitations in object detection. These modules enable the model to selectively concentrate on pertinent regions likely to contain objects, as well as to differentiate between objects and backgrounds, thereby enhancing performance. Lu et al. \cite{lu2023improved} added a Convolutional Block Attention Module (CBAM) in the backbone to facilitate the effective extraction of feature information and added a swin transformer to the YOLOv8s backbone network to extract global features from the training image and improve the detection accuracy of smaller targets. Addressing automated brain tumor detection, Kang et al. \cite{kang2023bgf} improved the baseline YOLOv8 model by incorporating Bi-level Routing Attention (BRA) and Generalized Feature Pyramid Networks (GFPN) to achieve a 4.7\% increase in mAP50 compared to YOLOv8x on the brain tumor detection dataset Br35H. Furthermore, Xia et al. \cite{xia2023global} proposed a global contextual attention augmented YOLOv8 model combining a Global Context module with a C2 module and a BiFPN feature fusion path. Wang et al. \cite{wang_road} optimized the feature pyramid layer in the YOLOv8 model with a SimSPPF module and introduced LSK-Attention to improve the recognition of road defects by 3.3\% in average precision mAP50, 29.92\% reduction in parameter volume, and 11.45\% reduction in computational load.
 \\

In summary, these changes and improvements demonstrate ongoing efforts to adapt YOLOv8 for different uses, leading to significant boosts in detection accuracy and efficiency across various fields.

\chapter{Improved YOLOv8 Model Structure}
To address the issues outlined in the Problem Statement \ref{sec:problem-statement}, this thesis study makes an effort to produce improvements to the baseline model to increase its accuracy of prediction and robustness in camera trap images. These enhancements include Integration of Attention mechanism, Modified Feature Fusion process, and a new loss function for Bounding Box Regression.

\section{Integration of Attention Mechanism}
The baseline model of the YOLOv8 has a poor ability to suppress the background information, hindering its capability to concentrate on the prominent features of the target object during the detection process. Therefore, the parts of the feature maps including the background are also activated during the object predictions. To suppress the background noisy information and further enhance the feature extraction capabilities of the model, an attention module has been introduced into the model.\\

The attention module enables the model to prioritize relevant information regarding the object of interest in an image. It also helps in suppressing background information to distinguish objects from surrounding clutter and identifying key features of the object from the image for precise object detection. The attention module serves to assign weight to specific features of the object in comparison to others, determining their relative importance. \\

\begin{figure}[H]
    \centering
    \includegraphics[scale=0.40]{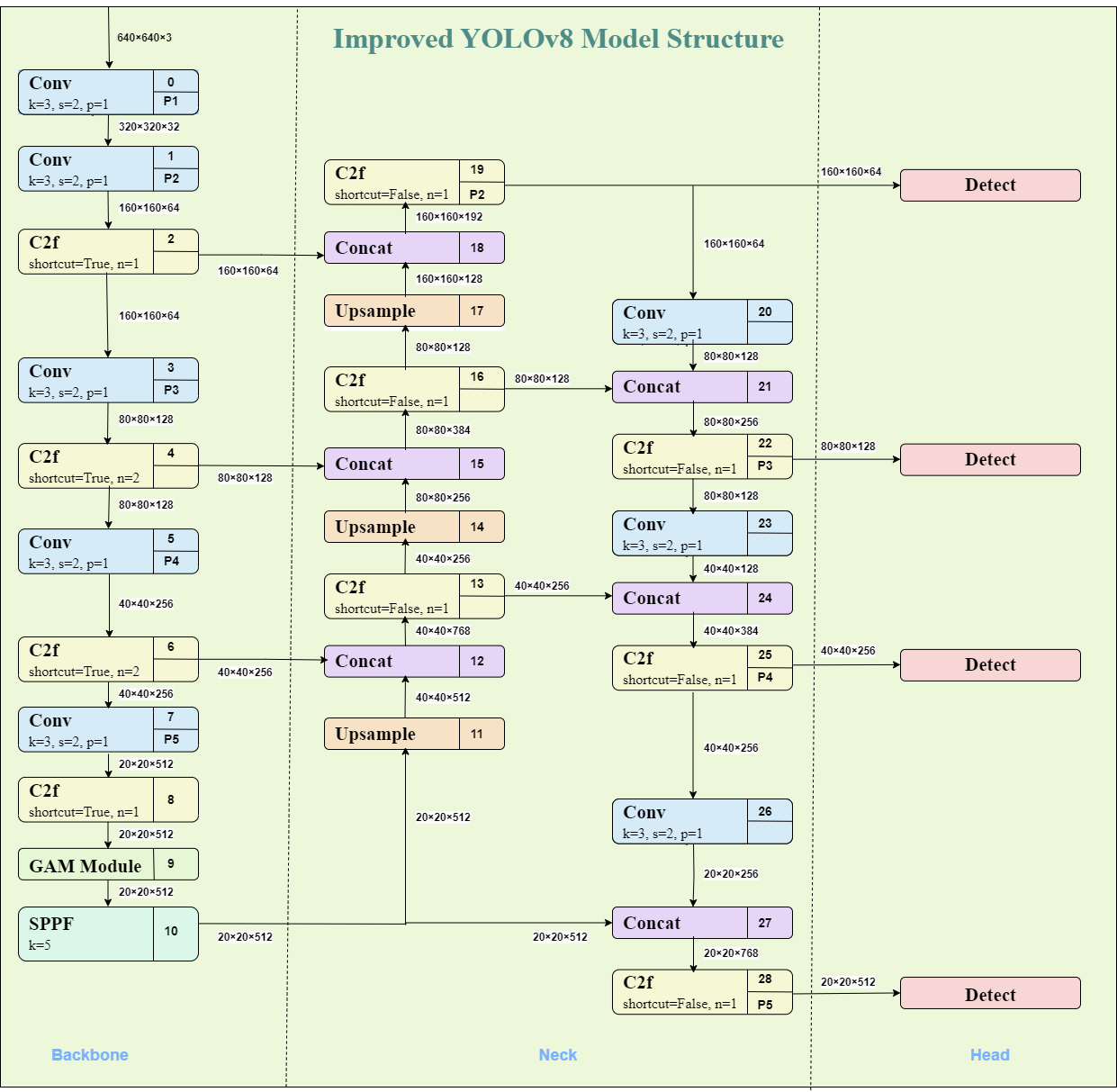}
    \caption{Improved YOLOv8s Model Structure}
    \label{fig:improved-yolov8-model}
\end{figure}

Out of all the attention modules available, the Global Attention Mechanism (GAM) \cite{liu2021global} has been incorporated in layer 9 of the improved model. The layer 9 is the last layer before the SPPF block in the backbone and is the most crucial point of the information flow. The extracted features after this layer are processed at a variety of scales, upsampled, downsampled, concatenated, and used by the detection head to make final predictions. Thus, refining the feature map at this layer contributes to the overall quality of the feature representations used for object detection. \\

The Global Attention Mechanism (GAM) enhances its focus on the essential features of the object of interest by sequentially incorporating Channel Attention and Spatial Attention modules. This approach allows the model to capture significant features across all three dimensions (Channel, Width, and Height) and produces an attention map, highlighting important regions. \\

\begin{figure}[H]
    \centering
    \includegraphics{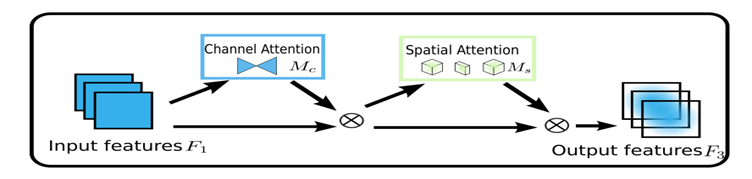}
    \caption{Global Attention Mechanism. Source: \cite{liu2021global}}
    \label{fig:gam}
\end{figure}

\begin{enumerate}
  \item\textbf{Channel Attention} \\
It focuses on “WHAT” information is most important across different channels/feature maps. It aims to capture the interconnections among different channels or feature maps, assigning distinct weights based on their significance for a specific task. This process enables the model to emphasize pertinent information in relevant channels while suppressing less relevant information. \\

\item\textbf{Spatial Attention} \\
It focuses on “WHERE” within an image or feature map important information lies. It works to capture the relationships across diverse spatial locations in the input image or feature maps. By assigning varying weights to different spatial locations, the model can focus on the important regions, effectively filtering out irrelevant or background areas. \\
\end{enumerate}

Some of the attention mechanisms/modules considered during the thesis focused only on either channel information or spatial information, overlooking the crucial interactions between the two. This oversight leads to information loss or suboptimal performance. The Global Attention Mechanism, employed as an attention module in this project, addresses this issue by preserving information across both channels and spatial dimensions. This approach mitigates information loss and consequently improves feature representation. \\

\subsection{Channel Attention Submodule in GAM}
The channel attention submodule in Global Attention Mechanism (GAM) consists of a 3D permutation to retain information across three dimensions (Channel, Width, and Height) and a two-layer Multi-Layer Perceptron to magnify cross-dimensional channel-spatial dependencies.

\begin{figure}[H]
    \centering
    \includegraphics{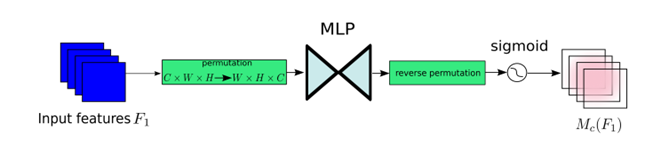}
    \caption{Channel Attention Submodule in GAM. Source: \cite{liu2021global}}
    \label{fig:channel-attention-submodule}
\end{figure}

\subsubsection{3D Permutation of Channel Attention}
This part promotes cross-channel interactions to enrich feature representation by distinctively reorganizing feature maps, encouraging the exchange of information among different channels. \\

Given the Input Features ($F_{1}$) of dimensions CxWxH, where C=Channels, W=Width, and H=Height, the 3D permutation step aligns the channels along the last dimension. \\

Before Permutation (CxWxH): \\
\[
\begin{array}{ccc}
 \begin{subarray}{c}
    \text{Channel 1} \\
    \begin{bmatrix}
        \begin{aligned}
        [a, b] \\
        [c, d] \\
    \end{aligned}
    \end{bmatrix}
     \end{subarray}
    &
     \begin{subarray}{c}
      \text{Channel 2} \\
        \begin{bmatrix}
            \begin{aligned}
            [e, f] \\
            [g, h] \\
        \end{aligned}
        \end{bmatrix}
     \end{subarray}
     &
     \begin{subarray}{c}
      \text{Channel 3} \\
        \begin{bmatrix}
            \begin{aligned}
            [i, j] \\
            [k, l] \\
        \end{aligned}
        \end{bmatrix}
     \end{subarray}
     
\end{array}
\]
\\

After Permutation: \\
\[
\begin{array}{cccc}
    \begin{subarray}{c}
        \text{Position 1} \\
        [a, e, i]
    \end{subarray}
    &
    \begin{subarray}{c}
        \text{Position 2} \\
        [b, f, j]
    \end{subarray}
    &
    \begin{subarray}{c}
        \text{Position 3} \\
        [c, g, k]
    \end{subarray}
     &
    \begin{subarray}{c}
        \text{Position 4} \\
        [d, h, l]
    \end{subarray}
\end{array}
\]

After permutation, the spatial dimensions are flattened, and the values from different channels are stacked together at each position. \\

After flattening ((W*H)xC),\\
\[
\begin{array}{c}
 \begin{subarray}{c}
    \begin{bmatrix}
        \begin{aligned}
        [a, e, i] \\
        [b, f, j] \\
        [c, g, k] \\
        [d, h, l] \\
    \end{aligned}
    \end{bmatrix}
     \end{subarray}
\end{array}
\]
\\

The 3D permutation operation plays a crucial role in shaping the input features. As indicated in the ``Shape" section of the PyTorch linear layer documentation \cite{pytorch_linear}, the linear layers within the Multi-Layer Perceptron (MLP) expect the input features to be situated in the last dimension. The 3D permutation operation achieves this adjustment, ensuring that the input features align with the expected format for processing by the linear layers of the Multi-Layer Perceptron (MLP).

\subsubsection{Multi-Layer Perceptron (MLP)}
After the channels are aligned by the 3D permutation operation, the 2-layer MLP analyses and models cross-channel dependencies to effectively amplify interactions. The 2-layer MLP consists of an input, hidden, and output layer and captures pair-wise interactions, accounting for the influence between two channels, as well as higher-order dependencies. The size of the hidden layer in the MLP used in the channel attention is (number of input channels (C)/4) while the size of the output layer is equal to the number of input channels (C) and helps us understand how multiple channels collectively contribute to feature representation. \\ \\
Suppose we have 16 channels (C), each having a height (H) and width (W) of 2. This results in a dimension of (CxWxH) i.e. 16x2x2. After applying a 3D permutation, it is reshaped into (W*H)xC i.e. 4x16. 
Following this permutation, the reshaped dimensions 4x16 is then fed into a Multi-Layer Perceptron (MLP) for further transformation.\\ \\
\textbf{Following our above example, The respective input and output dimensions are:} \\
Input to the 1st Linear Layer = (W*H)xC = 4x16\\
Output from the 1st Linear layer = (W*H)x(C/4) = 4x4\\\\
Input to the 2nd Linear Layer = (W*H)x(C/4) = 4x4\\
Output from the 2nd Linear layer = (W*H)xC = 4x16\\

The output of the MLP undergoes reverse permutation to restore the original 3D tensor shape (CxWxH), preserving the spatial structure of the feature and integrating the acquired cross-dimensional interactions. Subsequently, the reshaped tensor passes through the sigmoid function, which squashes the values within the range of 0 to 1. Each element in this tensor represents the attention weight for the corresponding channel at a specific spatial location. These weights are then employed to rescale the input features, selectively accentuating, or dampening different channels according to their learned importance. \\

\subsection{Spatial Attention Submodule in GAM}
The spatial attention submodule in the Global Attention Mechanism (GAM) consists of two convolutional layers. These layers analyze the input feature maps at various locations, facilitating the assignment of higher weights to regions that may potentially contain objects.

\begin{figure}[H]
    \centering
    \includegraphics{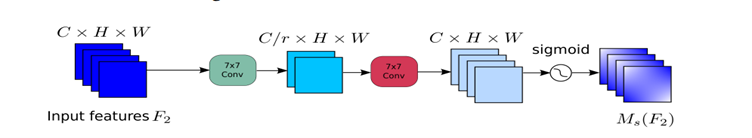}
    \caption{Spatial Attention Submodule in GAM. Source: \cite{liu2021global}}
    \label{fig:spatial-attention-submodule}
\end{figure}

The initial convolutional layer utilizes a 7x7 kernel to capture long-range dependencies by considering a broad receptive field and extracts global patterns and relationships. To enhance computational efficiency and compress features, the number of channels is reduced by a factor of ‘r’. The second convolutional layer also employs a 7x7 kernel to further process the condensed spatial information, potentially refining the extracted patterns. This layer restores the number of channels back to the original configuration. The sigmoid function maps the output values of the second convolutional layer to a range between 0 and 1, creating spatial attention weights that indicate the relative importance of different spatial regions for the object detection task.

\section{Modified Feature Fusion Process}
In the baseline YOLOv8 model, the feature fusion process occurs in the neck after the SPPF block. The neck of the YOLOv8 model integrates both the Feature Pyramid Network (FPN) and Path Aggregation Network (PAN) to enhance the object detection capabilities. The Feature Pyramid Network (FPN) constructs a feature map pyramid by combining a high-level feature map with a low-level feature map through the process of upsampling and concatenation. This process helps preserve spatial details for small objects and maintains semantic information for accurate class predictions. Simultaneously, The Path Aggregation Network (PAN) establishes additional connections between different levels in the FPN pyramid to improve the feature fusion process, facilitating the free flow of information across the network. This collaborative approach between FPN and PAN creates a comprehensive representation of feature maps, leading to improved feature representation and better context understanding, enabling the model to excel on diverse datasets with objects of varying sizes and complexities. \\

The Neck of the current baseline model doesn’t incorporate the processed feature maps from the first C2f block at layer 2. Generally, the upper layers in the model extract low-level features like edges and textures, and the features picked up by the upper layers also contain background clutters, noise, and other irrelevant details, potentially hindering detection accuracy. However, the fine-grained spatial details in the upper layers are essential for detecting small objects and for describing precise object boundaries. The effectiveness of including the upper layers in the feature fusion process is a debated topic and it depends on the specific dataset, tasks, and desired objectives due to the drawbacks considering the huge computation resource it requires and the inference time it takes. \\

As the information flows deeper into the network, the feature maps are downsampled to make the computation process cheaper and to focus on capturing patterns and shape. In this process, crucial features related to small objects tend to get lost or suppressed and cannot be retained even after multiple upsampling and concatenation operation performed in the Neck. Therefore, to create a more balanced representation of semantic information captured by the lower layers and the fine-grained details extracted by the upper convolutional layer, the improved model presented in the figure \ref{fig:improved-yolov8-model} incorporated the output of the C2f module present in the Layer 2 into the neck which then contributes to the final detection process. \\

\section{Bounding Box Regression Loss Function}

The bounding box regression is a technique used in object detection and localization tasks to refine the coordinates of the predicted bounding boxes, thereby improving the model’s localization performance. Initial predictions for the location of an object are typically imprecise, necessitating regression to improve the accuracy of the bounding boxes and achieve more precise object localization. The bounding box regression involves optimizing a loss function that quantifies the disparity between the predicted bounding box and the ground truth box. Through backpropagation, this optimization process adjusts the model’s parameters i.e. the neural network weights to minimize the loss. The fine-tuning process ensures the predicted bounding boxes closely align with the ground-truth box. The choice of the loss function for bounding box regression is important as it serves as a penalty measure that needs to be minimized during the training to lead to maximum overlap between the predicted and the ground truth boxes.

\subsection{Intersection Over Union (IoU) loss}
All the performance measure in the object detection task specifically relies upon the Intersection over Union (IoU) and the metric itself can be used for the loss function. The IoU metric is invariant to the scale of the problem under consideration so, therefore doesn’t have a bias towards large bounding boxes.

\begin{figure}[H]
    \centering
    \includegraphics[scale=0.40]{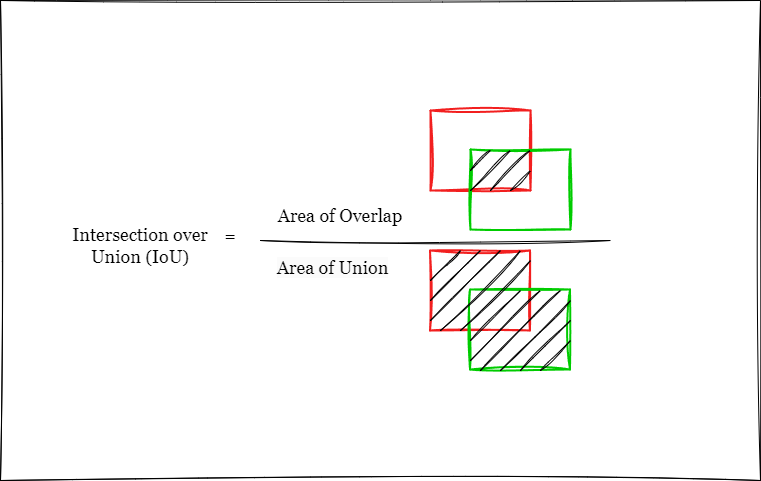}
    \caption{Intersection over Union (IoU)}
    \label{fig:iou}
\end{figure}

The IoU loss is given by,
\[
\mathcal{L}_{\text{IoU}} = 1 - \text{IoU}
\]

\begin{description}
\item\textbf{Issues with IoU as a loss function:} \\
For the instances where the predicted bounding box overlaps with the ground truth bounding box, the Intersection Over Union (IoU) can be directly employed as the objective function to minimize the loss. However, when dealing with non-overlapping bounding boxes, the gradient is zero, failing to indicate the proximity or distance between the predicted box and the ground truth box. With no moving gradient in non-overlapping cases, the loss function is rendered unoptimizable, leading to reduced accuracy, and causing slow convergence during the training.
\end{description}

Several loss functions have been introduced to tackle this problem,
\[
\mathcal{L}_{\text{i}} = \mathcal{L}_{\text{IoU}} + \mathcal{R}_{\text{i}}
\]

All the loss functions adds a penalty term (\( \mathcal{R}_{i} \)) to the IoU loss to address the issue.

\begin{figure}[H]
    \centering
    \includegraphics[scale=0.40]{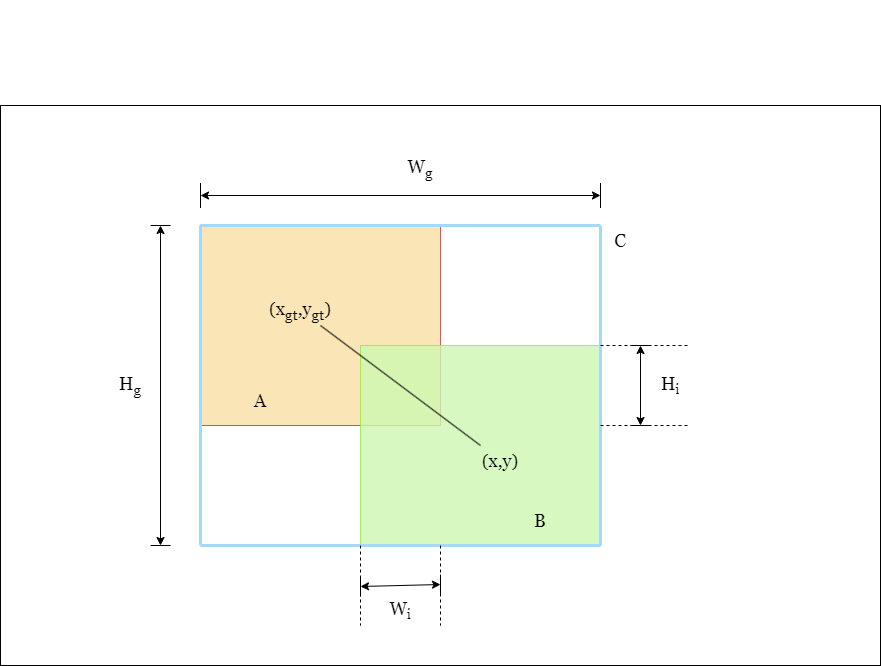}
    \caption{Ground Truth and Predicted Bounding box}
    \label{fig:iou-penalty}
\end{figure}

\subsection{Generalized Intersection Over Union (GIoU) loss}
The GIoU loss \cite{rezatofighi2019generalized} is strongly dependent on the Intersection Over Union (IoU) but tries to address non-overlapping cases by specifically considering the empty volume (area) between the predicted and ground truth box. The GIoU loss takes into account the difference between the area of the convex hull (C) and the area of the union (U) to penalize the predictions that do not entirely encompass the ground truth bounding box. This approach aids in aligning the predicted box more accurately with the objects it represents.

\[
\mathcal{L}_{\text{GIoU}} = \mathcal{L}_{\text{IoU}} + \mathcal{R}_{\text{GIoU}}
\]

\[
\mathcal{R}_{\text{GIoU}} =\frac{|C - (A \cup B)|}{|C|}
\]

The penalty term added to the IoU loss in GIoU loss guides the predicted bounding box to adjust its position towards the target box in the cases where there is no overlap between the predicted and target bounding boxes. 
\begin{description}
\item\textbf{Issues with GIoU as a loss function:}\\
a.	If $|A \cap B| = 0$, GIoU loss results in an expansion of the predicted bounding box to establish an overlap with the ground truth box. \\
b.	If $|A \cap B| > 0$, the area of $|C - (A \cup B)|$ is consistently small or equal to zero value. Consequently, GIoU loss reduces to IoU loss in such cases, giving rise to the problem of slow convergence and inaccurate regression
\end{description}

\subsection{Distance Intersection Over Union (DIoU) loss}
In DIoU \cite{zheng2019distanceiou}, the penalty term is introduced to directly minimize the normalized distance between the central points of predicted and ground truth bounding boxes, resulting in significantly quicker convergence. Unlike GIoU, which concentrates on reducing the difference between the area of the convex hull and the union area, the DIoU loss focuses on directly minimizing the distance between the central points. The DIoU loss concurrently considers the overlap area and central point distance of bounding boxes. While GIoU enlarges predicted bounding boxes to enhance the degree of overlap, the DIoU aims to decrease the central point distance between the two boxes. 
\[
\mathcal{L}_{\text{DIoU}} = \mathcal{L}_{\text{IoU}} + \mathcal{R}_{\text{DIoU}}
\]

\[
\mathcal{R}_{\text{DIoU}} = \frac{{(x - x_{\text{gt}})^2 + (y - y_{\text{gt}})^2}}{{W_{g}^2 + H_{g}^2}}
\]

The normalized distance is square of the Euclidean distance between the central points of two bounding boxes. \\

\begin{description}
\item\textbf{Issues with DIoU as a loss function:}\\
The distance term in the DIoU may exhibit heightened sensitivity to extreme outliers, potentially causing unstable gradients and impeding the convergence process during training.
\end{description}

\subsection{Complete Intersection Over Union (CIoU) loss}
In CIoU \cite{zheng2019distanceiou} loss, a penalty term is introduced by integrating three geometric factors into the bounding box regression: overlap area (IoU), central point distance, and aspect ratio. The CIoU loss function is presently employed for bounding box regression within the YOLOv8 model, facilitating object localization in images. \\

The aspect ratio term incorporated into the loss function encourages the predicted bounding box to closely match the aspect ratio of the ground truth bounding box. This ensures that the predicted bounding box maintains a similar shape and orientation as the ground truth bounding box. The loss function penalizes the predicted bounding box with aspect ratios significantly deviating from the aspect ratio of the ground truth bounding box.

\[
\mathcal{L}_{\text{CIoU}} = \mathcal{L}_{\text{IoU}} + \mathcal{R}_{\text{CIoU}}
\]

\[
\mathcal{R}_{\text{CIoU}} = \mathcal{R}_{\text{DIoU}} + \alpha \cdot v, \alpha = \frac{v}{{\mathcal{L}_{\text{IoU}} + v}}
\]
Here, $v$ describes the consistency of the aspect ratio and is defined by:
\[
v = \left(\frac{4}{\pi^2}\right)\left(\tan^{-1}\frac{w}{h} - \tan^{-1}\frac{w_{gt}}{h_{gt}}\right)^2
\]

\begin{description}
\item\textbf{Issues with CIoU as a loss function:}\\
Efficiently measuring the disparities between the target box and the predicted box is challenging, resulting in slower convergence and inaccurate localization during model optimizations.
\end{description}

\subsection{Efficient Intersection Over Union (EIoU) loss}
The EIoU loss \cite{zhang2022focal} preserves the beneficial characteristics of the CIoU loss and directly minimizes the difference between the predicted bounding box’s width and height and that of the ground truth bounding box. It assesses the discrepancies in three key geometric factors i.e. the overlap area, the central point, and the side length. \\

An inherent challenge in bounding box regression, known as the imbalance problem, arises when a large number of predicted bounding boxes having small overlaps with ground truth boxes disproportionately contribute to the optimization step. The EIoU loss addresses this problem by introducing the regression version of focal loss, directing the regression process to focus on high-quality boxes with large Intersection over Unions (IoUs) and down-weighting the contributions of the low-quality boxes, resulting in the Focal-EIoU loss. This is achieved by assigning the highest gradient gains to boxes with higher IoUs.
\[
\mathcal{L}_{\text{EIoU}} = \mathcal{L}_{\text{IoU}} + \mathcal{R}_{\text{EIoU}}
\]
\[
\mathcal{R}_{\text{EIoU}} = \mathcal{L}_{\text{dis}} + \mathcal{L}_{\text{asp}}
\]
\hfill{where dis = central point distance and asp = side length}
\[
\mathcal{R}_{\text{EIoU}} = \frac{(x - x_{\text{gt}})^2 + (y - y_{\text{gt}})^2}{W_g^2 + H_g^2} + \frac{\rho^2(w, w_{\text{gt}})}{W_g^2} + \frac{\rho^2(h, h_{\text{gt}})}{H_g^2}
\]

The Focal-EIoU is given by
\[
\mathcal{L}_{\text{Focal-EIoU}} = \text{IOU}^{\gamma} \cdot \mathcal{L}_{\text{EIoU}}
\]

$\gamma$ = \text{parameter to control the degree of inhibition of outliers and works well when } $\gamma$ = 0.5\\

\begin{description}
\item\textbf{Issues with EIoU as a loss function:}\\
In specific situations, such as cases involving highly elongated boxes or significant overlap between predicted and ground truth boxes, the EIoU loss may display unstable behavior or become less effective.
\end{description}

\subsection{Wise Intersection Over Union (WIoU) loss}
Traditional IoU-based losses assign uniform weights to all predicted bounding boxes, leading to challenges when making actual predictions. Training data inevitably comprises low-quality examples and the usage of geometric factors like distance and aspect ratio in the penalty term exacerbates the penalty during training potentially impacting the model’s ability to learn from such instances and may compromise the overall model’s performance. Simply reinforcing the fitting ability of the Bounding Box Regression loss on low-quality examples poses a risk to the overall localization performance of the model. The WIoU \cite{tong2023wiseiou} loss addresses this issue by mitigating the harmful gradients generated by the low-quality examples. It also reduces the competitiveness of the high-quality bounding boxes, allowing for a greater influence of the ordinary quality boxes during the training process. \\

To diminish the competitiveness of high-quality boxes, the WIoU loss employs a dynamic non-monotonic focusing mechanism. A Focusing Mechanism (FM) is a strategy designed to adapt the gradient update of the model based on the quality of the anchor box. Unlike monotonic strategies that assign higher gradient gains to higher-quality boxes with lower IoU losses, the non-monotonic and dynamic nature of FM considers factors such as the consistency of IoU losses over time and the degree of outliers to enhance the overall localization performance of the model and doesn’t simply assign higher weights to higher IoU values all the time. \\

\[
\mathcal{L}_{\text{WIoUv1}} = \mathcal{R}_{\text{WIoU}} \cdot \mathcal{L}_{\text{IoU}}
\]

\[
\mathcal{R}_{\text{WIoU}} = \exp\left(\frac{{(x - x_{\text{gt}})^2 + (y - y_{\text{gt}})^2}}{{W_{g}^2 + H_{g}^2}}\right)
\]

The dynamic non-monotonic FM mitigates the impact of low-quality examples by assigning small gradient grains to those instances with a large value of $\beta$. A small $\beta$ indicates that the box is of high quality, and a small gradient gain is assigned to it, allowing the model to focus on the other examples. \\

The outlier degree $\beta$ is given by
\begin{align*}
\beta = \frac{\mathcal{L}^*_{\text{IoU}}}{\overline{\mathcal{L}_{\text{IoU}}}} \quad \text{where} \quad \beta \in [0, +\infty)
\end{align*}

And, the Wise Intersection Over Union (WIoU) is given by
\begin{align*}
\mathcal{L}_{\text{WIoUv3}} &= r \cdot \mathcal{L}_{\text{WIoUv1}} \\
\text{where} \quad r &= \frac{\beta}{\delta \alpha^{(\beta-\delta)}}
\end{align*}

$\alpha$ and $\delta$ are hyper-parameters and the model performs superiorly when the value of $\alpha$ = 1.9 and $\delta$=3. 

\chapter{Visualization and Evaluation Criteria}
\section{Visualization}
Heatmaps serve as a visual representation of spatial information by applying a color-coded grid to indicate the presence and location of objects within an image. Primarily used for the purpose of AI Explainability, each pixel in the heatmap corresponds to a specific location in the input image. The generated heatmaps function as a visualization tool, facilitating the interpretation of a model’s predictions. Overlaying the heatmap on the input image reveals the areas where the model concentrates its attention during object predictions. \\

Typically, warmer colors such as red and yellow signify higher probabilities of object presence, while cooler colors like blue and green indicate lower probabilities. By emphasizing the prominent features contributing positively to the end class predictions, heatmaps offer insights into the decision-making process of the model. Subsequently, heatmaps enhance the transparency and explainability of workings of the deep learning models. \\

In the context of YOLOv8, Heatmaps are particularly valuable for understanding which parts of the feature maps or the image contribute significantly to the detection process. This helps explain whether the model focuses on the object of interest or on the background clutters when making predictions. Overall, heatmaps play a crucial role in providing interpretable insights into the inner workings of the YOLOv8 model, contributing to better understanding and trust in the Artificial Intelligence system in the broad sense. \\

\begin{figure}[h]
    \centering
    \includegraphics[scale=0.45]{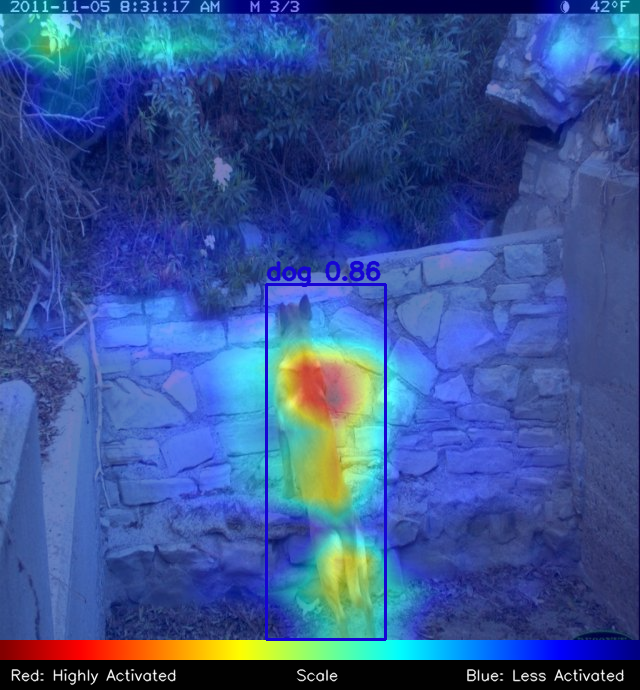}
    \caption{Visualization of heatmap for one of the input image}
    \label{fig:dog-heatmap}
\end{figure}

There are different ways heatmaps can be generated for AI explainability and one of the most important and popular such techniques is Gradient Weighted Class Activation Mapping (Grad-CAM) \cite{Selvaraju_2019}. Grad-CAM is a technique within deep learning and computer vision, employed to generate heatmaps that visually highlight the regions in an input image where a model focuses during predictions. It aids in comprehending which parts of an image are pivotal in the decision-making process of a deep neural network. The primary purpose of Grad-CAM is to understand the reasoning behind a deep neural network’s decision for a specific image and class. \\

In the Grad-CAM, the gradients of the predicted class’s confidence score with respect to the feature maps generated by the selected convolutional layer are analyzed to identify the important regions in the input image that influenced/contributed to the model’s predictions. Generally, the visualization focuses on the end layers of the deep neural network, as these layers provide comprehensive feature representations. \\

Here is a step-by-step breakdown of how Grad-CAM generates the heatmaps in the case of YOLOv8 object detection model:
\begin{enumerate}
  \item \textbf{Calculate gradients for all the feature maps of the selected Conv layer:} \\ 
  The gradients measure how much the model’s output (confidence score) changes with respect to the small changes in the input (pixels in the selected feature map). These gradients in turn reveal how much each pixel in the feature maps contributes to the model’s confidence in predicting that class.
  
  \item \textbf{Global Average Pooling:} \\ 
  Take the global average of these gradients for each feature map in the selected convolutional layer to obtain a scalar weight for each feature map.

  \item \textbf{Generate Activation map:} \\ 
  The activation map is generated by multiplying the feature map with the assigned weight and calculating the weighted sum to emphasize the most important features.

  \item \textbf{Identification of samples that contribute positively:} \\ 
  The Rectified Linear Unit (ReLU) activation function is applied to consider only the samples that contribute positively toward the model’s prediction of a target class.

  \item \textbf{Upsample and normalize:} \\ 
  The reweighted activation map is upsampled to match the size of the original image and then normalized to a scale of 0 to 1.

  \item \textbf{Overlay on top of input image:} \\ 
  The output from step 5 is then overlaid onto the original image, creating a visual representation of the crucial regions where the model focused on while making predictions.
\end{enumerate}

\section{Evaluation Criteria}

To assess the effectiveness of the changes made to the structure of the baseline model, the following evaluation criteria, Precision (P), Recall (R), Average Precision (AP), and Mean Average Precision (mAP) are employed. These evaluation criteria help us objectively know how well our model is performing, and whether the performance of the model meets our expected standards and helps evaluate the overall accuracy of the model predictions.\\

\subsection{Confusion Matrix}
Confusion Matrix provides a granular assessment of the performance of an object detection model, enabling us to make informed decisions about model enhancements and optimizations. Its significance lies in assessing the model's effectiveness across various classes and its precision in localizing objects within images.  It helps in the examination of the accuracy of both class predictions and bounding box localizations to understand the strengths and weaknesses inherent in the trained model. The confusion matrix breaks down the model’s predictions for each class, providing information on true positives, false positives, true negatives, and false negatives. \\

\begin{figure}[H]
    \centering
    \includegraphics[scale=0.40]{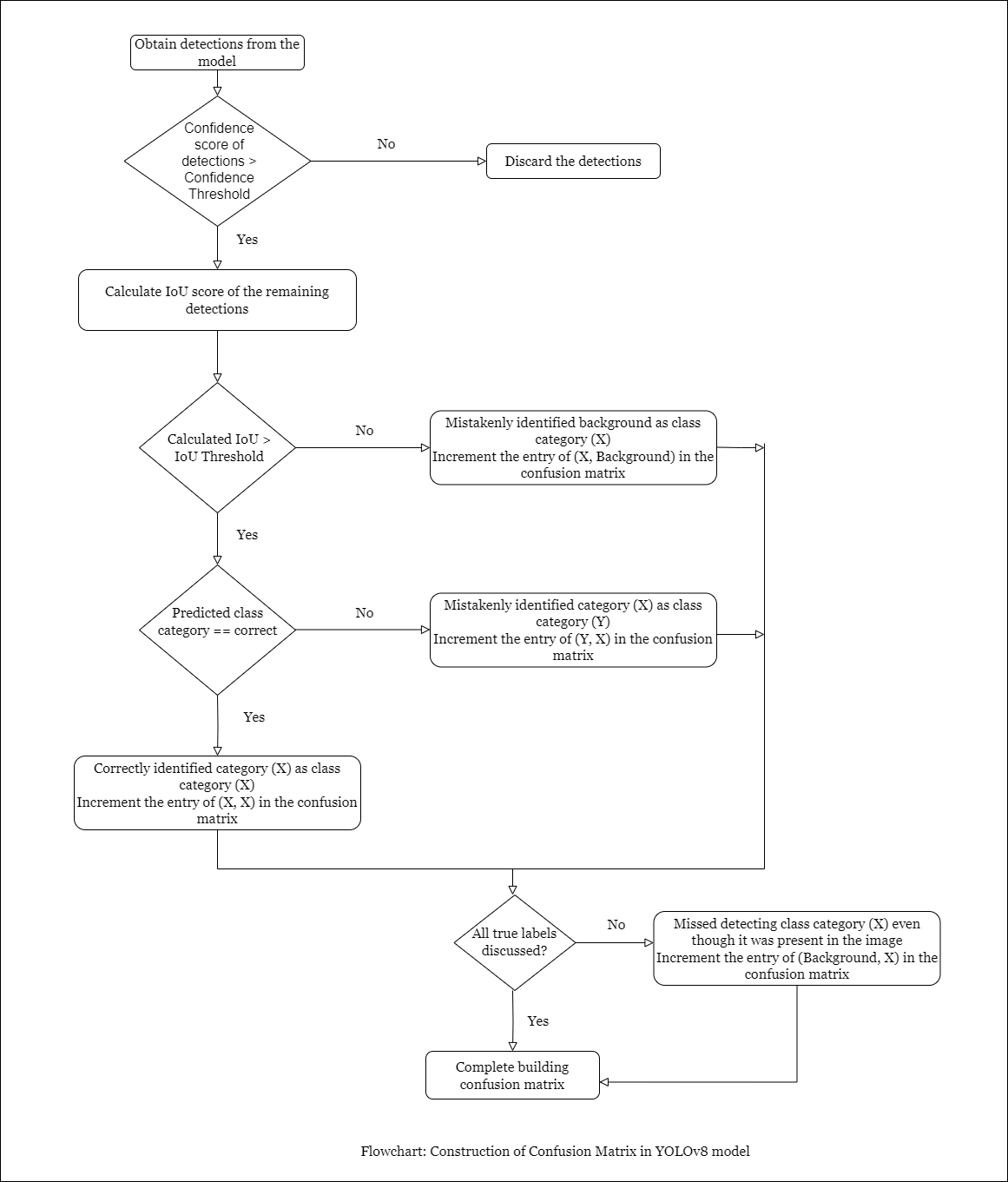}
    \caption{Process of building Confusion Matrix in YOLOv8 model}
    \label{fig:confusion-matrix}
\end{figure}

\subsubsection{True Positive (TP)}
A True Positive (TP) is an instance where the trained model is able to localize and categorize the object of interest correctly. In other words, the model identifies an object, assigns it to the correct class, and accurately determines the bounding box co-ordinates to locate the object.

\subsubsection{False Positive (FP)}
A False Positive (FP) is an instance where the trained model is either not able to localize the object of interest at all or is able to localize and not able to categorize the object. In other words, in one case, the model determines that there is an object but assigns incorrect bounding box coordinates and is unable to locate the object, or in the second case, the model accurately determines the bounding box co-ordinates to locate the object but is not able to identify it and assigns the incorrect class.

\subsubsection{False Negative (FN)}
A False Negative (FN) is an instance where the trained model fails to detect the object present in the image. In other words, the model misses or fails to recognize an object that should have been detected.

\subsubsection{True Negative (TN)}
There is no concept of true negatives in object detection as there isn’t a well-defined negative class in the context of identifying and localizing objects.

\subsection{Precision}
Precision in the context of object detection is the fraction of accurate predictions out of all the predictions that the model has made. It is known as Positive Predictive Value and assesses the accuracy of the positive predictions made by the model. \\

The precision is calculated using the following formula:

\begin{equation}
\label{eq:precision}
Precision (P) = \frac{\text{True Positives}}{\text{True Positives} + \text{False Positives}}
\end{equation}

\subsection{Recall}
Recall in the context of object detection is the fraction of accurate predictions out of all the objects present in the ground truth. It is known as the Sensitivity or True Positive Rate and assesses how well the model can find all the relevant objects in an image.\\

The recall is calculated using the following formula:
\begin{equation}
\label{eq:recall}
Recall (R) = \frac{\text{True Positives}}{\text{True Positives} + \text{False Negatives}}
\end{equation}

\subsection{Trade-off between Precision and Recall}
There is a trade-off between the above two metrics, Precision and Recall. Adjusting the confidence threshold affects precision and recall.
\begin{enumerate}
  \item \textbf{Higher confidence threshold:} \\ 
  Increases precision as the model will be more cautious in making correct predictions. This in turn decreases recall as the model while being cautious misses most objects that it should have detected.
  
  \item \textbf{Lower confidence threshold:} \\ 
  Increases recall as the number of detections made by the model increases. This in turn decreases precision as more detections might result in more false positives.
\end{enumerate}

\subsection{Precision – Recall curve (P-R curve)}
A P-R curve serves as a graphical representation that depicts the trade-off between precision and recall at varying confidence scores in the context of an object detection. The P-R curve is plotted as a function of confidence thresholds, and it is plotted for each category/class label. The X-axis represents recall (R) and the Y-axis represents precision (P). Each point along the curve corresponds to a specific confidence score, offering insights into how adjustments in confidence scores impact the trade-off between precision and recall in object detection. \\
To plot the P-R curve for a certain class label (category), we first consider all the detections made by the model for that class label along with the ground truth labels. All the detections have confidence scores signifying the confidence of the model in localizing the object and predicting the class label. \\

\begin{figure}[H]
    \centering
    \includegraphics[scale=0.35]{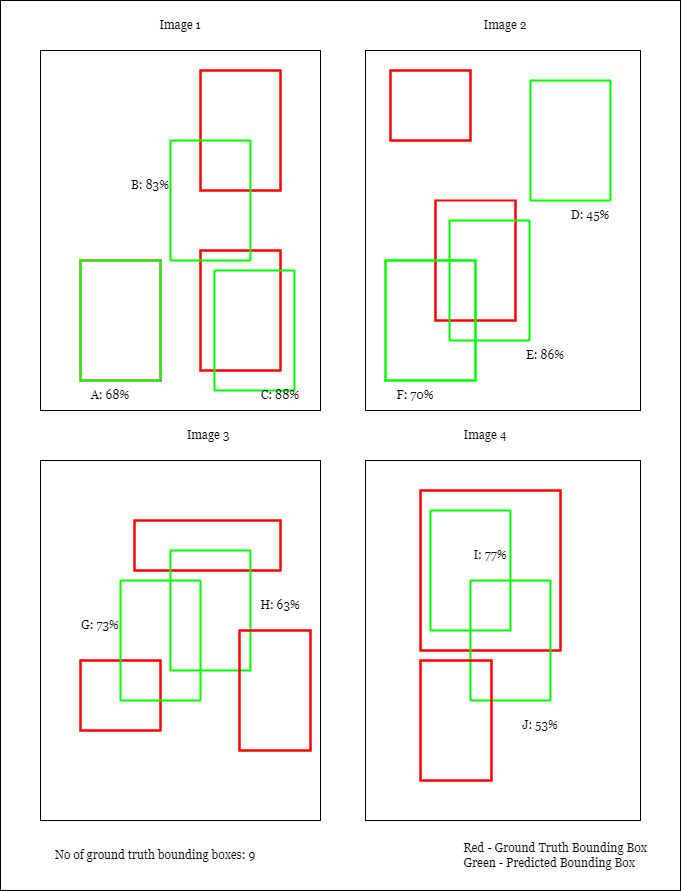}
    \caption{Ground Truth and Predicted Bounding Boxes in sample images}
    \label{fig:pr-calc-from-detections}
\end{figure}

We then create a table for all the detections throughout our image dataset for that specific class category and then determine if the detection categorizes as True Positive (TP) or False Positive (FP) using the flowchart in the figure \ref{fig:confusion-matrix}

\begin{figure}[H]
    \centering
    \includegraphics[scale=0.65]{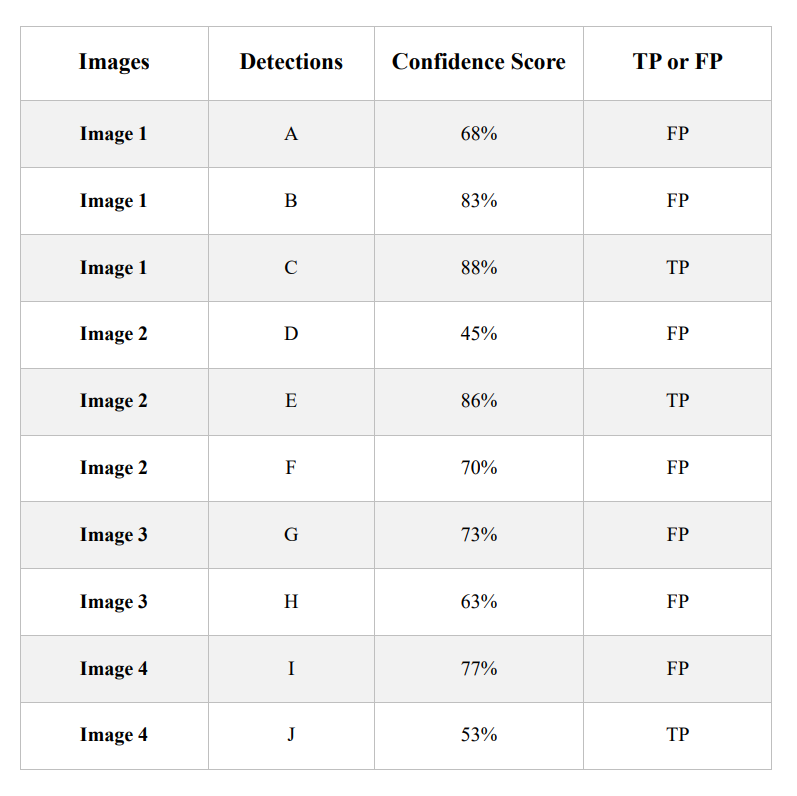}
    \caption{All the detections from all the images}
    \label{fig:detections-tp-fp}
\end{figure}

Then, the detections are sorted in the decreasing order of the confidence scores. After sorting, we compute Accumulated True Positive and Accumulated False Positive in a cumulative manner and calculate Precision (P) and Recall (R) using the equation \ref{eq:precision} and \ref{eq:recall} respectively.

\begin{figure}[H]
    \centering
    \includegraphics[scale=0.60]{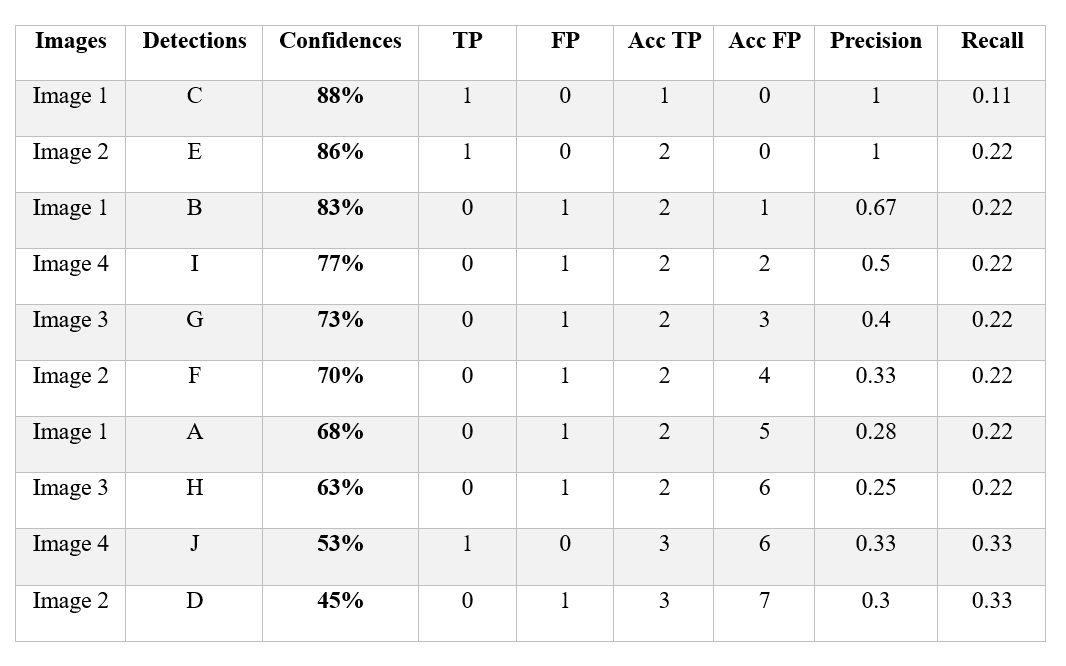}
    \caption{Detections sorted in decreasing order of confidence scores}
    \label{fig:detections-tp-fp-decreasing-conf}
\end{figure}

Using the above table, the Precision-Recall curve is then plotted:
\begin{figure}[H]
    \centering
    \includegraphics[scale=0.65]{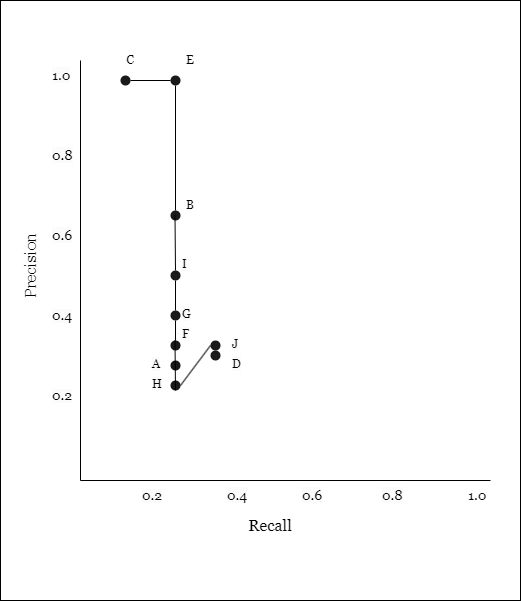}
    \caption{Precision-Recall Curve}
    \label{fig:pr-curve-custom}
\end{figure}
To obtain more comprehensive summary of the model’s performance for that class category and to obtain a smooth curve that is less sensitive to the fluctuations, the P-R curve is often interpolated at the end.
\begin{figure}[H]
    \centering
    \includegraphics[scale=0.65]{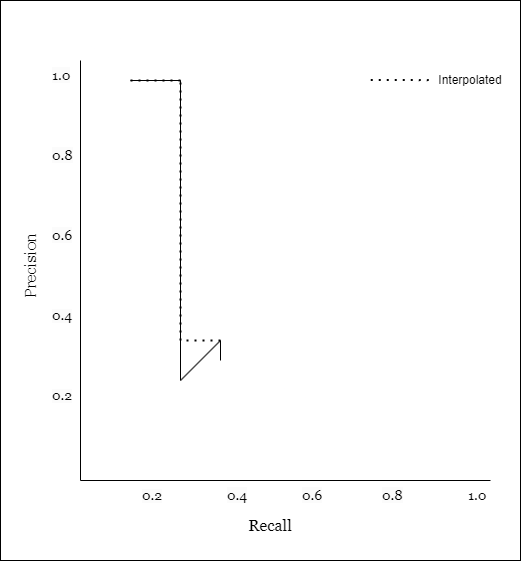}
    \caption{Interpolated Precision-Recall Curve}
    \label{fig:interpolated-pr-curve}
\end{figure}

\subsection{Average Precision (AP)}
The Precision–Recall (P-R) curve is plotted as a function of confidence thresholds for a certain class category. The Average Precision (AP) is computed from the area under the interpolated Precision-Recall curve. Mathematically, it is the average of precision values at different recall levels.

\begin{figure}[H]
    \centering
    \includegraphics[scale=0.65]{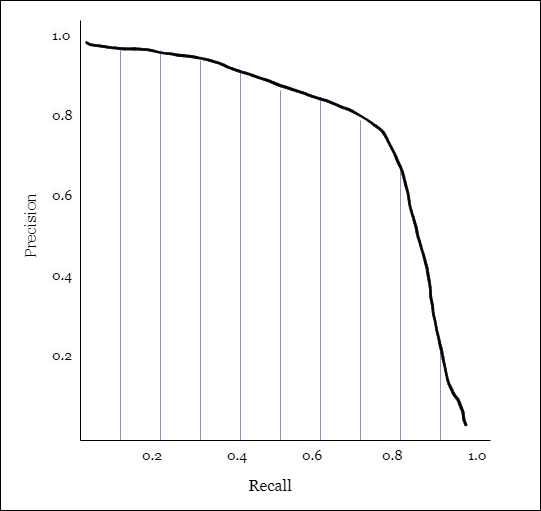}
    \caption{Precision-Recall Curve}
    \label{fig:pr-curve}
\end{figure}

The Average Precision (AP) is defined as:
\begin{equation}
\label{eq:average_precision}
AP = \frac{1}{101} \sum_{k=1}^{n}AP_{\text{r}}\ 
\end{equation}
where \(AP_{\text{r}}\) is the area under the interval and calculated using the Trapezoidal Rule.

\subsection{Mean Average Precision (mAP)}
Mean Average Precision (mAP) is the mean of the Average Precision (AP) calculated for all the class categories.

The Mean Average Precision (mAP) is defined as:
\begin{equation}
\label{eq:mean_average_precision}
mAP = \frac{1}{C} \sum_{i=1}^{C} AP_i
\end{equation}
where \(AP_{\text{i}}\) = the Average Precision (AP) for the class category $i$ computed using \ref{eq:average_precision}\\
C = Number of class categories/labels\\

mAP is a crucial metric offering a thorough assessment of a model’s performance across all classes. Unlike relying solely on precision and recall, mAP takes into consideration both the accuracy and completeness of the detections, accounting for the model’s performance across various confidence thresholds. mAP is widely adopted and serves as a standard metric for benchmarking and comparing different object detection models.\\

The metric used for evaluating the performance of the improved YOLOv8 model for this thesis is mAP50. mAP50 means that the Mean Average Precision (mAP) is computed at a specific IoU threshold of 0.5. It represents the average detection performance across all classes when considering a prediction as a true positive if it has at least 50\% overlap (0.5 IoU) with a ground truth bounding box. The IoU threshold referred to in the flowchart diagram in figure \ref{fig:confusion-matrix} is set to 0.5 in the case of mAP50.
\chapter{Results and Analysis}\label{chap:results-and-analysis}
\section{Dataset Selection}
Some of the popular camera trap datasets are Snapshot Serengeti, Caltech Camera Traps Dataset, Florida Wildlife Dataset, Snapshot USA 2019, Kaggle iWildCam 2019 etc. \\

To address the problem outlined in the \ref{sec:problem-statement}, the Caltech Camera Traps dataset is selected. The dataset’s objective is specifically targeted towards addressing the challenge of generalization in classification and detection. In the subset of the Caltech Camera Traps dataset, as explored by Beery et al. \cite{beery2018recognition}, the dataset is categorized based on the location IDs provided. The three categories include the Training set, Cis-location, and Trans-location. Within the Cis-location category, there are further subdivisions, namely the validation set and test set. Both sets contain images from the same pool of locations as the training set. On the other hand, the Trans-location category comprises a test set and a validation set with images from an entirely different pool of locations, not included in either the Training set or the Cis-location dataset. \\

Out of a total of 20 locations, a random selection includes 9 locations designated for use as trans-location test data and one location chosen at random for trans-location validation data. The remaining 10 locations contribute to the cis-location test data, with a specific focus on images captured on odd days. For cis-location validation data, 5\% of the images taken on even days are utilized, ensuring that the training and validation sets do not share identical image sequences. The remaining dataset captured on even days is allocated for training purposes. \\

The subset of the Caltech Camera Trap dataset discussed in \cite{beery2018recognition}, comprising 57,868 images, serves as an apt benchmark dataset for quantitatively assessing and measuring the recognition generalization ability of object detection models in unfamiliar or novel environments. This dataset encompasses 13,553 training images, 3,484 validation images, and 15,827 test images from cis-locations. Additionally, it includes 1,725 validation images and 23,275 test images from trans-locations. \\

To mitigate additional generalization challenges stemming from changing vegetation and animal species distributions across seasons, the cis and test data are interleaved day by day, as opposed to using a single date for data splitting. This interleaving approach builds a comprehensive dataset incrementally and helps reduce noise to facilitate a clean experimental comparison of results between cis and trans-locations. \\

The dataset comprises 16 animal categories, including an ``empty" category. These categories are opossum, raccoon, squirrel, bobcat, skunk, dog, coyote, rabbit, bird, cat, badger, empty, car, deer, fox, and rodent. \\

The image distribution in the dataset after analysis shows images with bounding box annotations and images with no bounding box annotations. \\
\begin{table}[H]
  \centering
  \resizebox{0.80\textwidth}{!}{%
  \begin{tabular}{|c|c|c|c|}
    \hline
     & with annotations & without annotations & Total \\
    \hline
    Train & 12099 & 1454 & 13553 \\
    Cis-Validation & 1665 & 1819 & 3484\\
    Cis-Test& 12691 & 3136 & 15827\\
    Trans-Test & 18033 & 5242 & 23275\\
    \hline
  \end{tabular}
  }
  \caption{Distribution of dataset with and without bounding box annotations}
\end{table}

For training, validation, and testing purposes, all the images with no bounding box annotations i.e. images with the label `empty' have been removed. Thus, the dataset after filtering has the following number of images in each set: \\
\begin{table}[H]
  \centering
  \resizebox{0.30\textwidth}{!}{%
  \begin{tabular}{|c|c|}
    \hline
     & Total \\
    \hline
    Train & 12099 \\
    Cis-Validation & 1665\\
    Cis-Test& 12691 \\
    Trans-Test & 18033\\
    \hline
  \end{tabular}
  }
  \caption{Number of images in each set after filtering}
\end{table}

Few of the data processing techniques are employed to prepare the dataset for the task of object detection.
\begin{enumerate}
    \item \textbf{Image Resizing:}\\
    All the images are resized to 640x640 pixels. The standard for the YOLOv8 object detection model training is 640x640 so all of the images from all 3 categories (Train, Cis, and Trans) are resized.
    
    \item \textbf{Data Augmentation:}\\
   Several data augmentation including rotation, scaling, and changes in brightness and contrast are applied to a select percentage of the images from the training set so that the model is able to generalize better to variations in the input data and to increase the number of input samples. The bounding box coordinates are updated accordingly after data transformation techniques to accurately represent the position of the object.
    
\end{enumerate}
The number of images after the application of the data processing technique to the training dataset:
\begin{table}[H]
  \centering
  \resizebox{0.30\textwidth}{!}{%
  \begin{tabular}{|c|c|}
    \hline
     & Total \\
    \hline
    Train & 19143 \\
    Cis-Validation & 1665\\
    Cis-Test& 12691 \\
    Trans-Test & 18033\\
    \hline
  \end{tabular}
  }
  \caption{Number of images in each set after the application of data processing technique}
  \label{tab:dataset-numbers}
\end{table}

The distribution of the animals in the Training set, Cis-Validation set, Cis-Test set, and Trans-Test set is given below

\begin{figure}[H]
    \centering
    \includegraphics[scale=0.65]{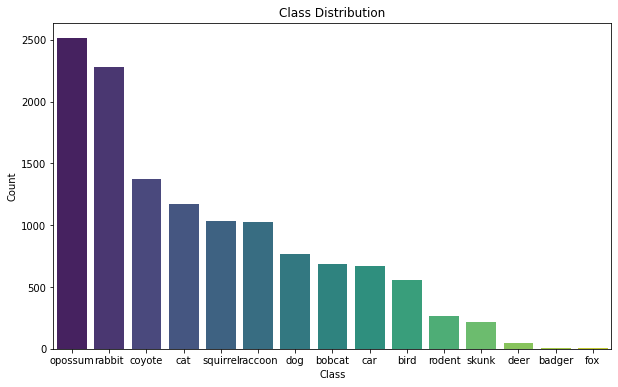}
    \caption{Class distribution in Training dataset}
    \label{fig:train-class-distribution}
\end{figure}

\begin{figure}[H]
    \centering
    \includegraphics[scale=0.65]{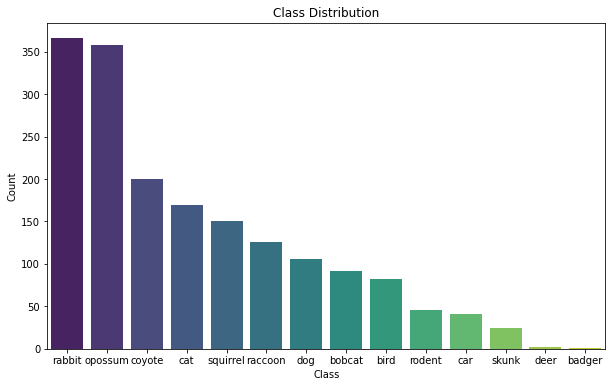}
    \caption{Class distribution in Cis-Validation dataset}
    \label{fig:cis-val-class-distribution}
\end{figure}

\begin{figure}[H]
    \centering
    \includegraphics[scale=0.65]{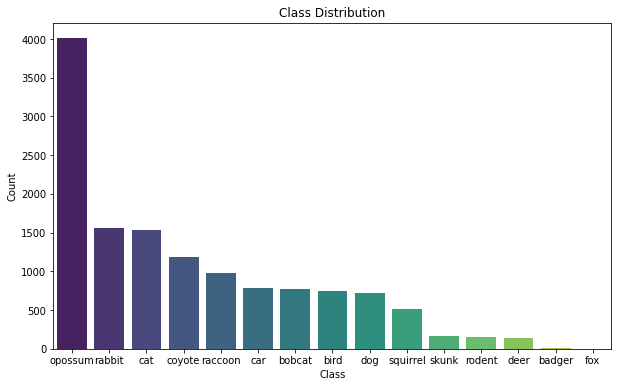}
    \caption{Class distribution in Cis-Test dataset}
    \label{fig:cis-test-class-distribution}
\end{figure}

\begin{figure}[H]
    \centering
    \includegraphics[scale=0.65]{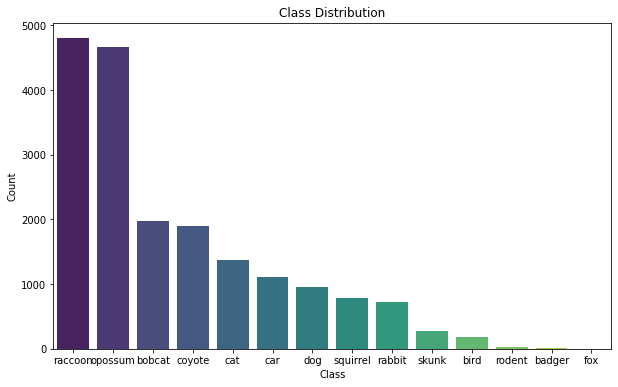}
    \caption{Class distribution in Trans-Test dataset}
    \label{fig:trans-test-class-distribution}
\end{figure}

\begin{figure}[H]
    \begin{minipage}{0.5\textwidth}
        \centering
        \includegraphics[width=\linewidth]{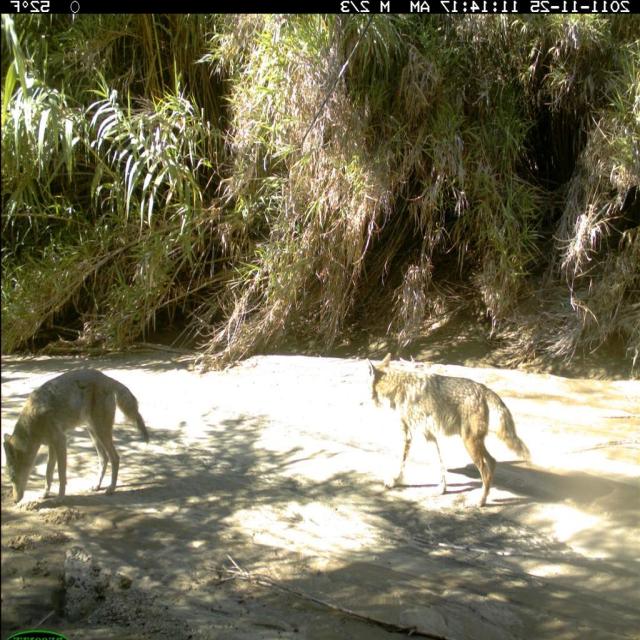}
    \end{minipage}%
    \begin{minipage}{0.5\textwidth}
        \centering
        \includegraphics[width=\linewidth]{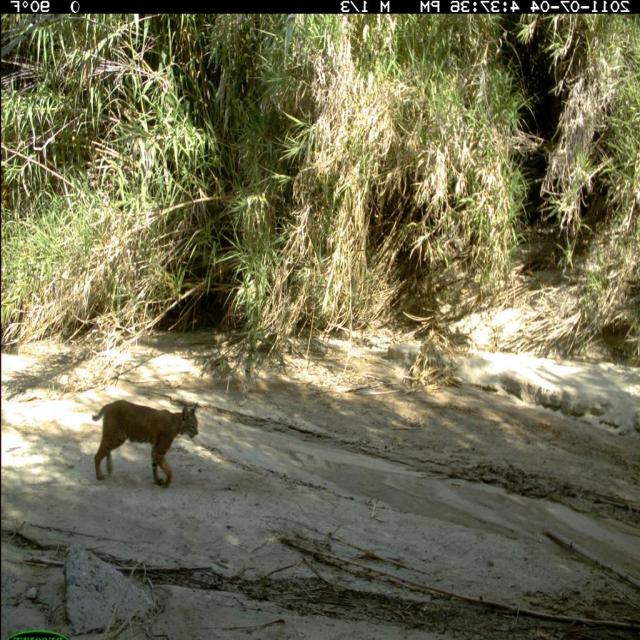}
    \end{minipage}
    
    \medskip
    
    \begin{minipage}{0.5\textwidth}
        \centering
        \includegraphics[width=\linewidth]{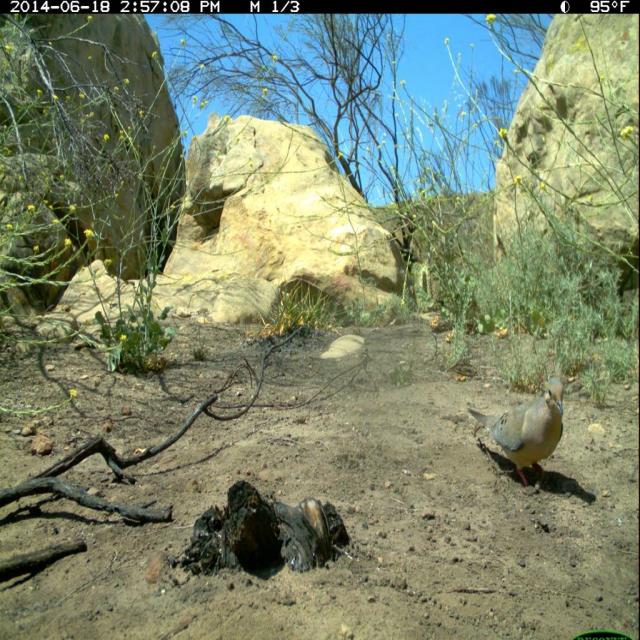}
    \end{minipage}%
    \begin{minipage}{0.5\textwidth}
        \centering
        \includegraphics[width=\linewidth]{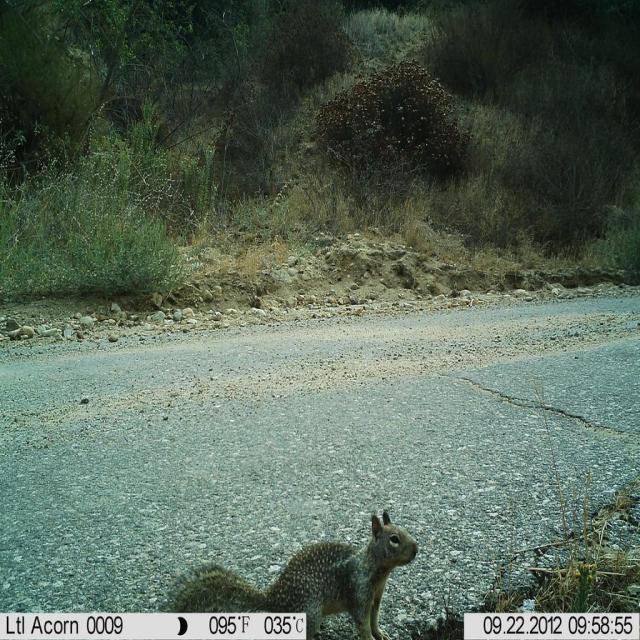}
    \end{minipage}
    
    \caption{Images from the training dataset}
\end{figure}

\begin{figure}[H]
    \begin{minipage}{0.5\textwidth}
        \centering
        \includegraphics[width=\linewidth]{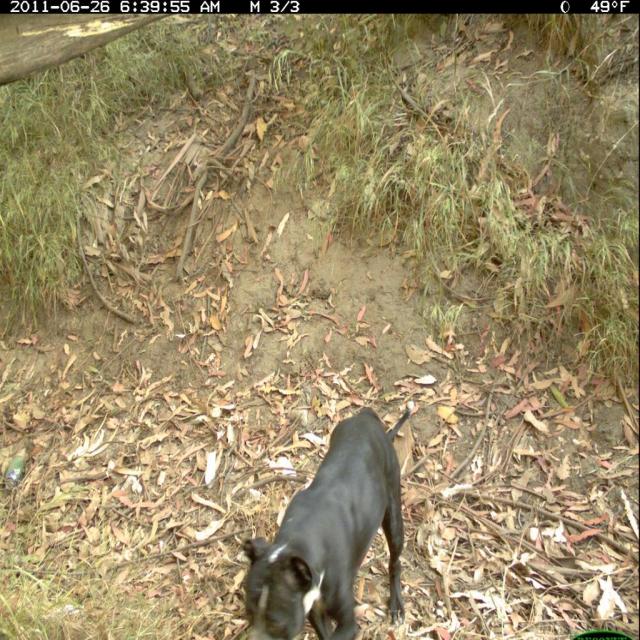}
    \end{minipage}%
    \begin{minipage}{0.5\textwidth}
        \centering
        \includegraphics[width=\linewidth]{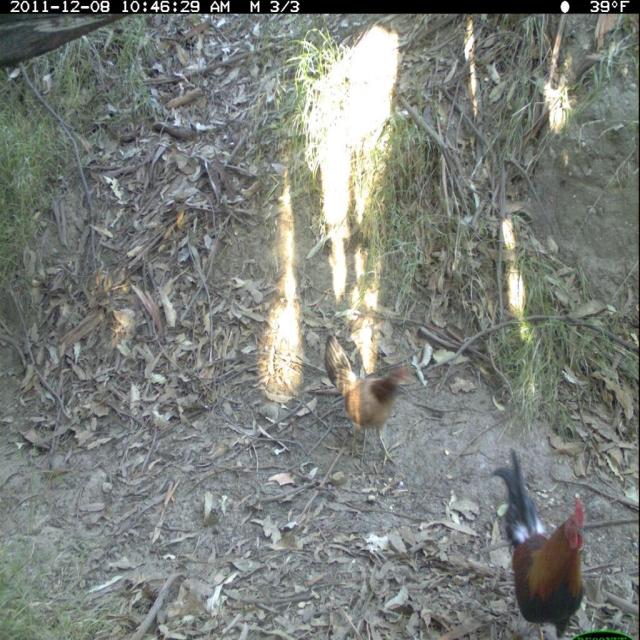}
    \end{minipage}
    
    \medskip
    
    \begin{minipage}{0.5\textwidth}
        \centering
        \includegraphics[width=\linewidth]{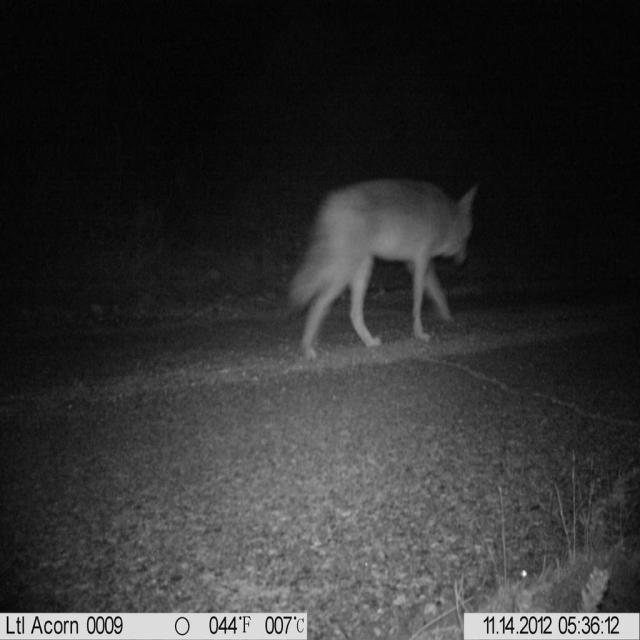}
    \end{minipage}%
    \begin{minipage}{0.5\textwidth}
        \centering
        \includegraphics[width=\linewidth]{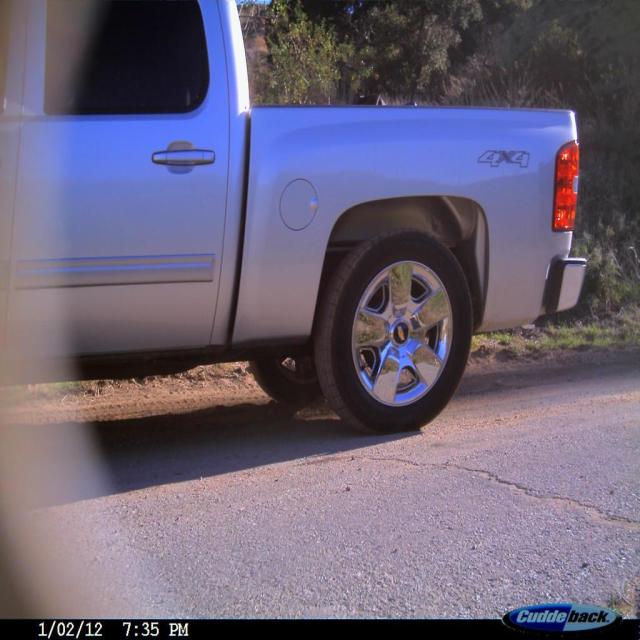}
    \end{minipage}
    
    \caption{Images from the cis-validation dataset}
\end{figure}

\begin{figure}[H]
    \begin{minipage}{0.5\textwidth}
        \centering
        \includegraphics[width=\linewidth]{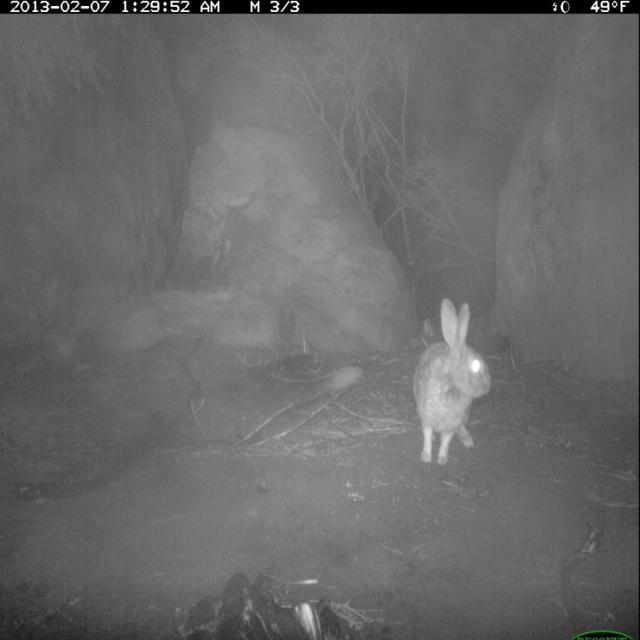}
    \end{minipage}%
    \begin{minipage}{0.5\textwidth}
        \centering
        \includegraphics[width=\linewidth]{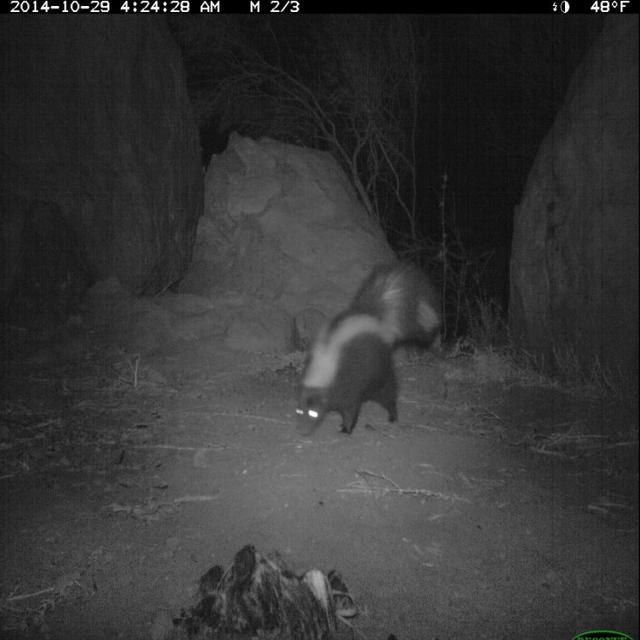}
    \end{minipage}
    
    \medskip
    
    \begin{minipage}{0.5\textwidth}
        \centering
        \includegraphics[width=\linewidth]{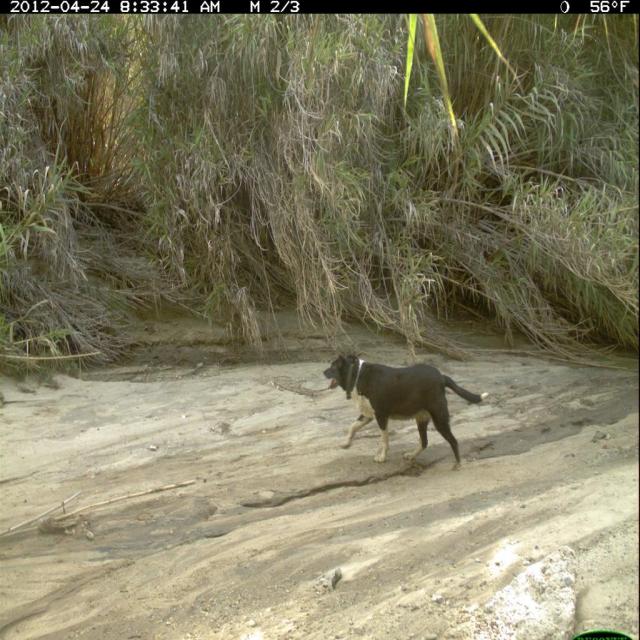}
    \end{minipage}%
    \begin{minipage}{0.5\textwidth}
        \centering
        \includegraphics[width=\linewidth]{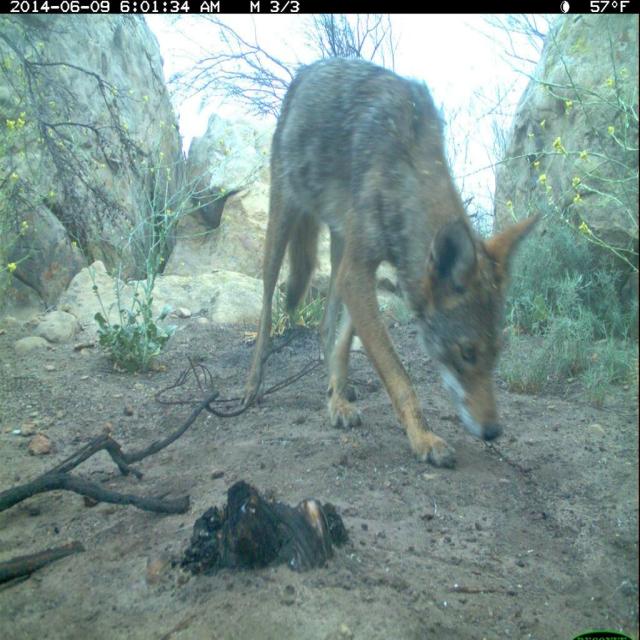}
    \end{minipage}
    
    \caption{Images from the cis-test dataset}
\end{figure}

\begin{figure}[H]
    \begin{minipage}{0.5\textwidth}
        \centering
        \includegraphics[width=\linewidth]{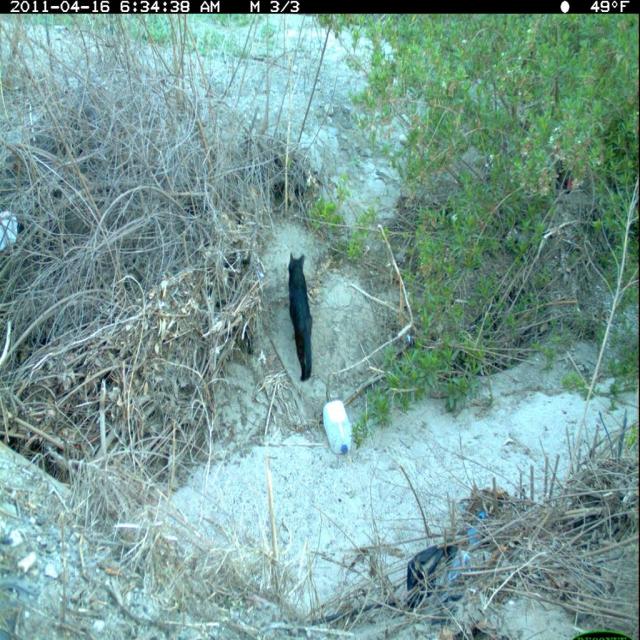}
    \end{minipage}%
    \begin{minipage}{0.5\textwidth}
        \centering
        \includegraphics[width=\linewidth]{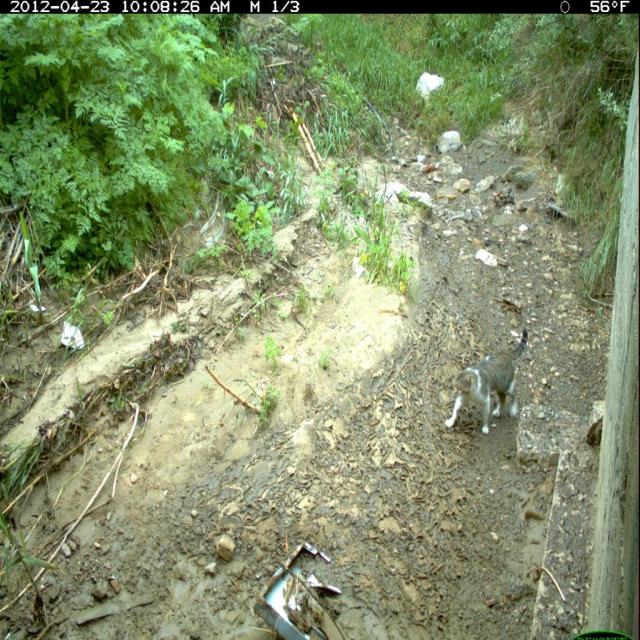}
    \end{minipage}
    
    \medskip
    
    \begin{minipage}{0.5\textwidth}
        \centering
        \includegraphics[width=\linewidth]{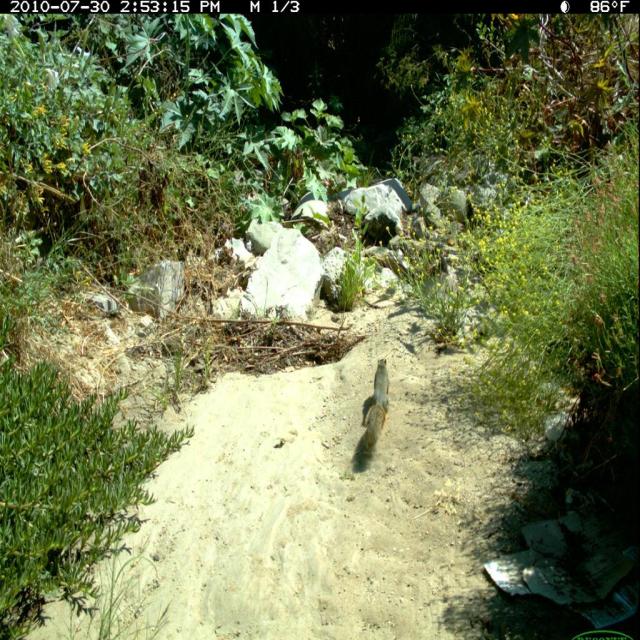}
    \end{minipage}%
    \begin{minipage}{0.5\textwidth}
        \centering
        \includegraphics[width=\linewidth]{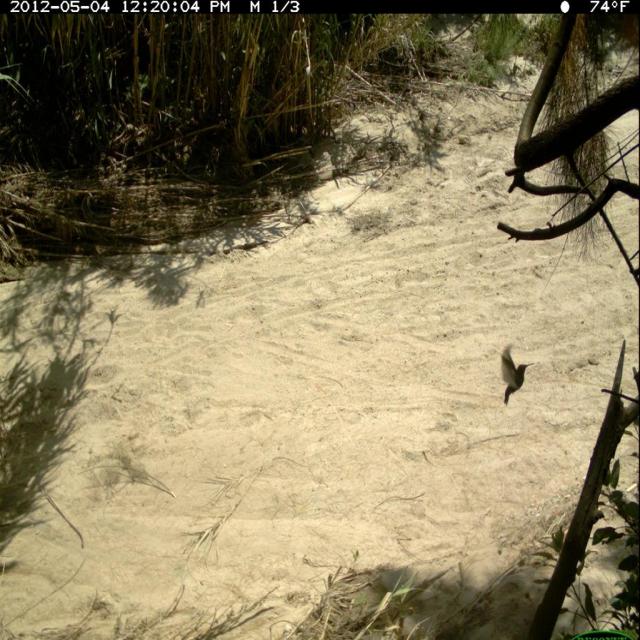}
    \end{minipage}
    
    \caption{Images from the trans-test dataset}
\end{figure}

\section{Experimental Setup}
\begin{table}[H]
  \centering
  \resizebox{0.55\textwidth}{!}{%
  \begin{tabular}{|c|c|}
    \hline
    Operating System & Linux \\
    Linux Kernel Version & 5.15.0-1030-nvidia \\
    Linux Distribution & Ubuntu \\
    Driver Version & 525.125.06\\
    GPU Memory & 81920 MiB \\
    CUDA Version & 12.0 \\
    \hline
  \end{tabular}
  }
  \caption{Experimental setup used for the training, validation, and testing.}
\end{table}

\begin{table}[H]
  \centering
  \resizebox{0.35\textwidth}{!}{%
  \begin{tabular}{|c|c|}
    \hline
    Image Size & 640x640 \\
    Learning Rate & 0.01 \\
    Weight Decay & 0.0005 \\
    Momentum & 0.937\\
    Batch Size & 16 \\
    Epoch & 180 \\
    \hline
  \end{tabular}
  }
  \caption{Hyperparameters used during training.}
\end{table}
All the evaluation experiments are performed at an IoU threshold of 0.45 and Confidence threshold of 0.25.

\section{Evaluation of baseline YOLOv8s model}

Upon conducting training and evaluation of the baseline YOLOv8s model using the above-discussed dataset, the model achieves the following evaluation metrics.

\begin{table}[H]
  \centering
  \resizebox{0.80\textwidth}{!}{%
  \begin{tabular}{|c|c|c|c|}
    \hline
     & Cis-Validation & Cis-Test & Trans-Test \\
    \hline
    Precision (P) & 0.881 & 0.798 & 0.599 \\
    Recall (R) & 0.86 & 0.791 & 0.484\\
    mAP50 & 0.889 & 0.813 & 0.52\\
    mAP50-95 & 0.672 & 0.595 & 0.371\\
    \hline
  \end{tabular}
  }
  \caption{Evaluation metrics for YOLOv8s baseline model}
\end{table}

A. Training Plots
\begin{figure}[H]
    \centering
    \includegraphics[scale=0.55]{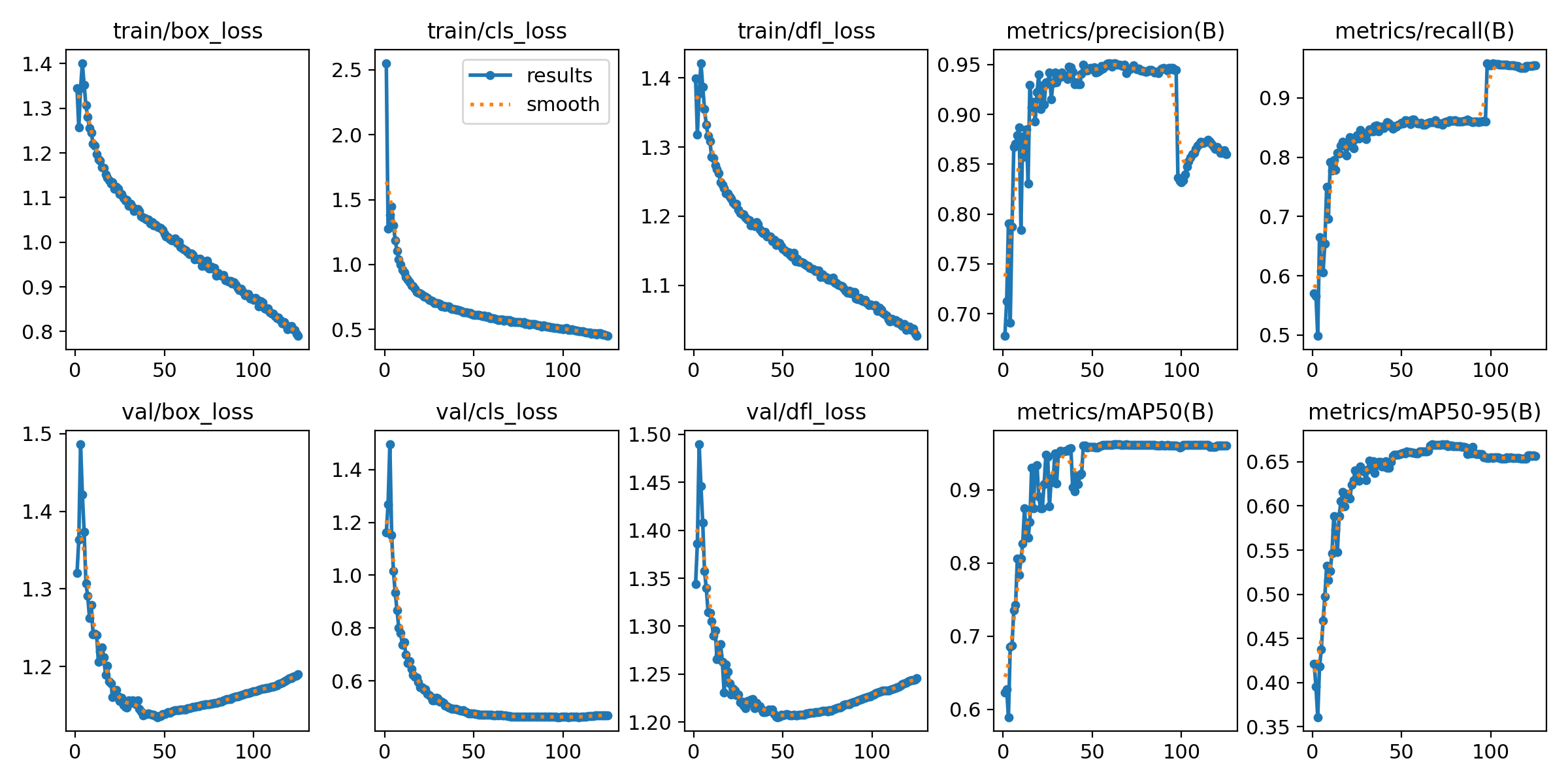}
    \caption{Visualization of different curves for YOLOv8s baseline model during training}
    \label{fig:baseline-different-curve}
\end{figure}

B. Cis-Test Evaluation Plots
\begin{figure}[H]
    \centering
    \includegraphics[scale=0.5]{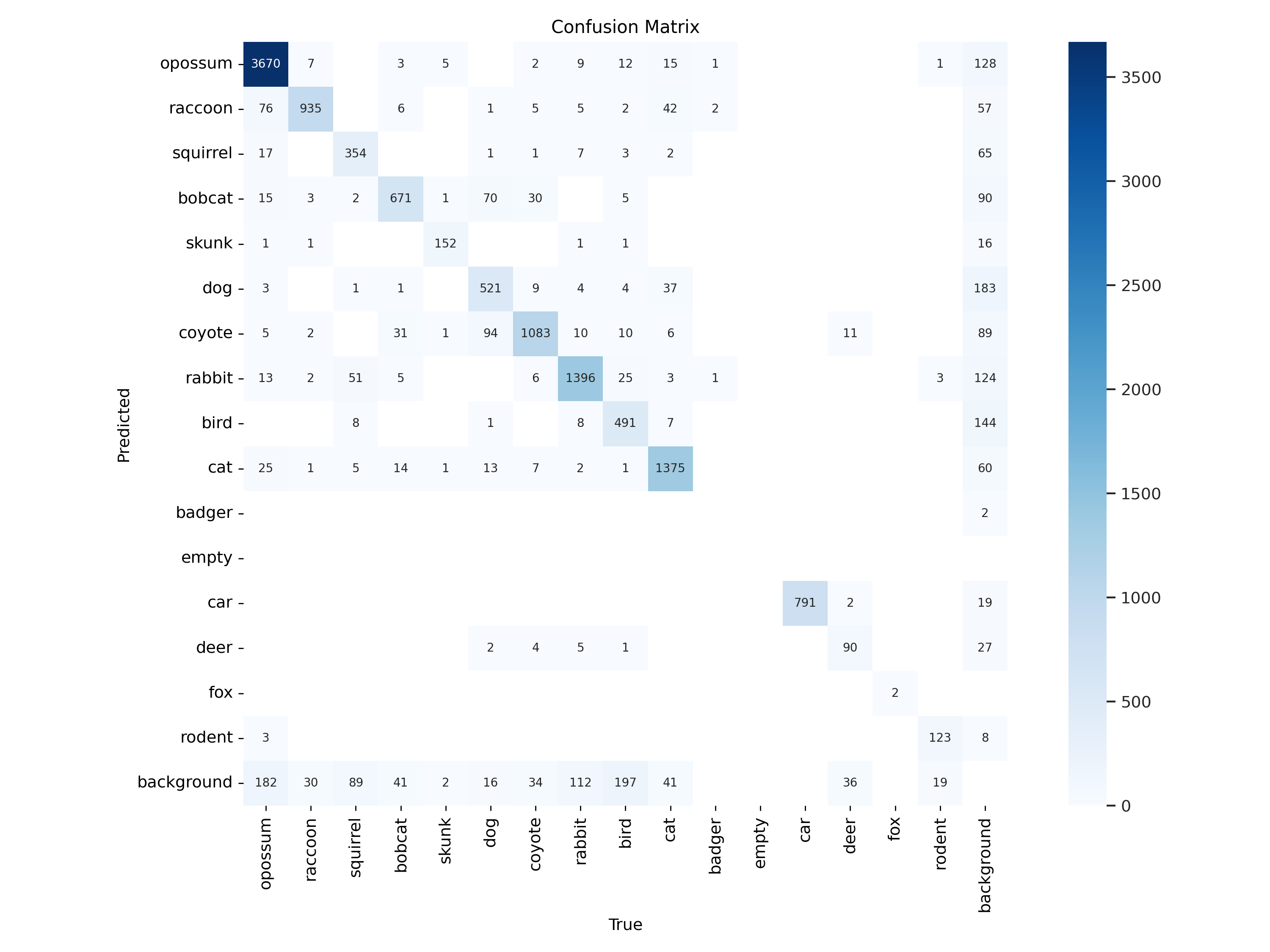}
    \caption{Confusion Matrix for YOLOv8s baseline model for Cis-Test set}
    \label{fig:baseline-cis-confusion-matrix}
\end{figure}

\begin{figure}[H]
    \centering
    \includegraphics[scale=0.5]{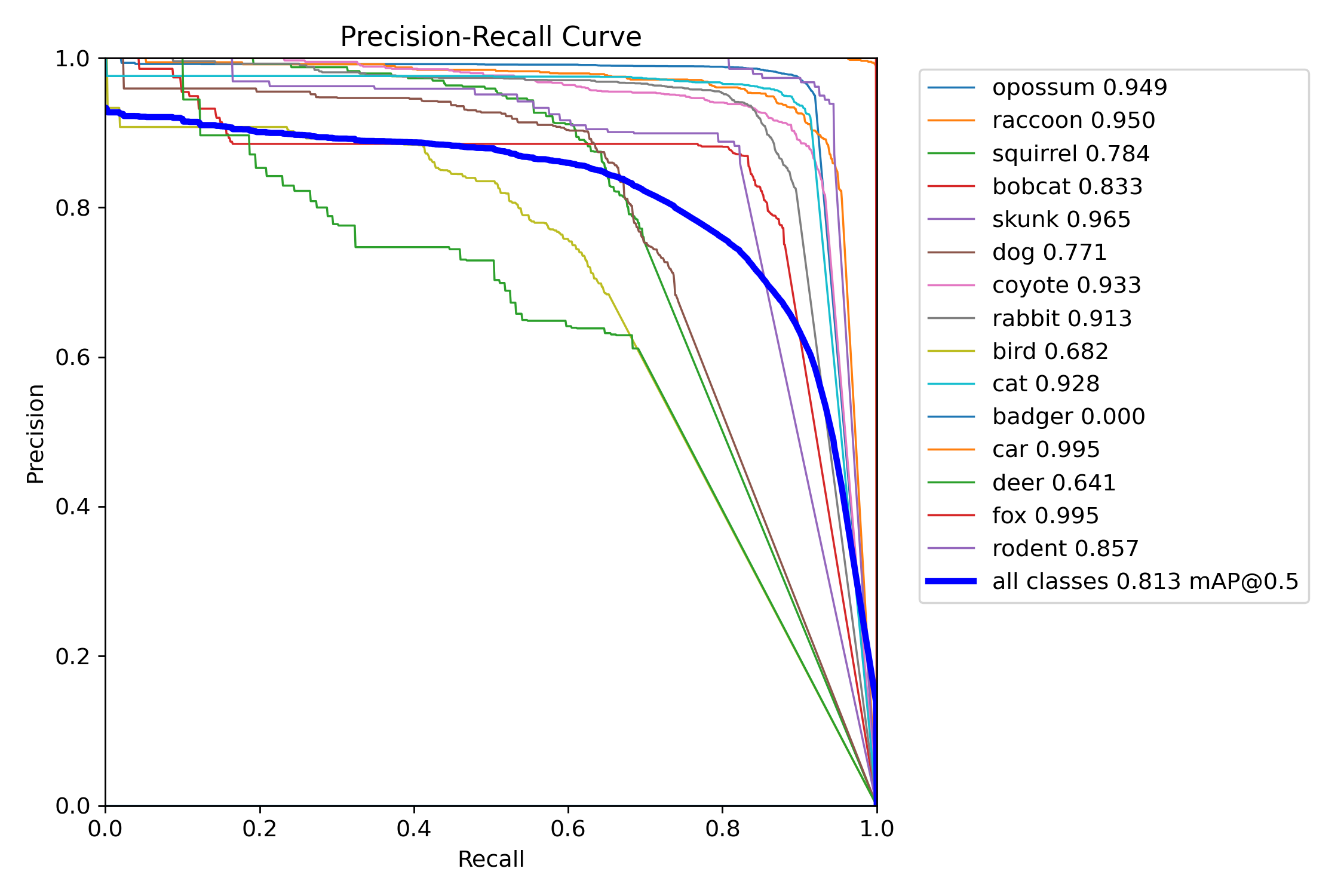}
    \caption{Precision-Recall curve for YOLOv8s baseline model for Cis-Test set}
    \label{fig:baseline-cis-pr-curve}
\end{figure}

C. Trans-Test Evaluation Plots
\begin{figure}[H]
    \centering
    \includegraphics[scale=0.5]{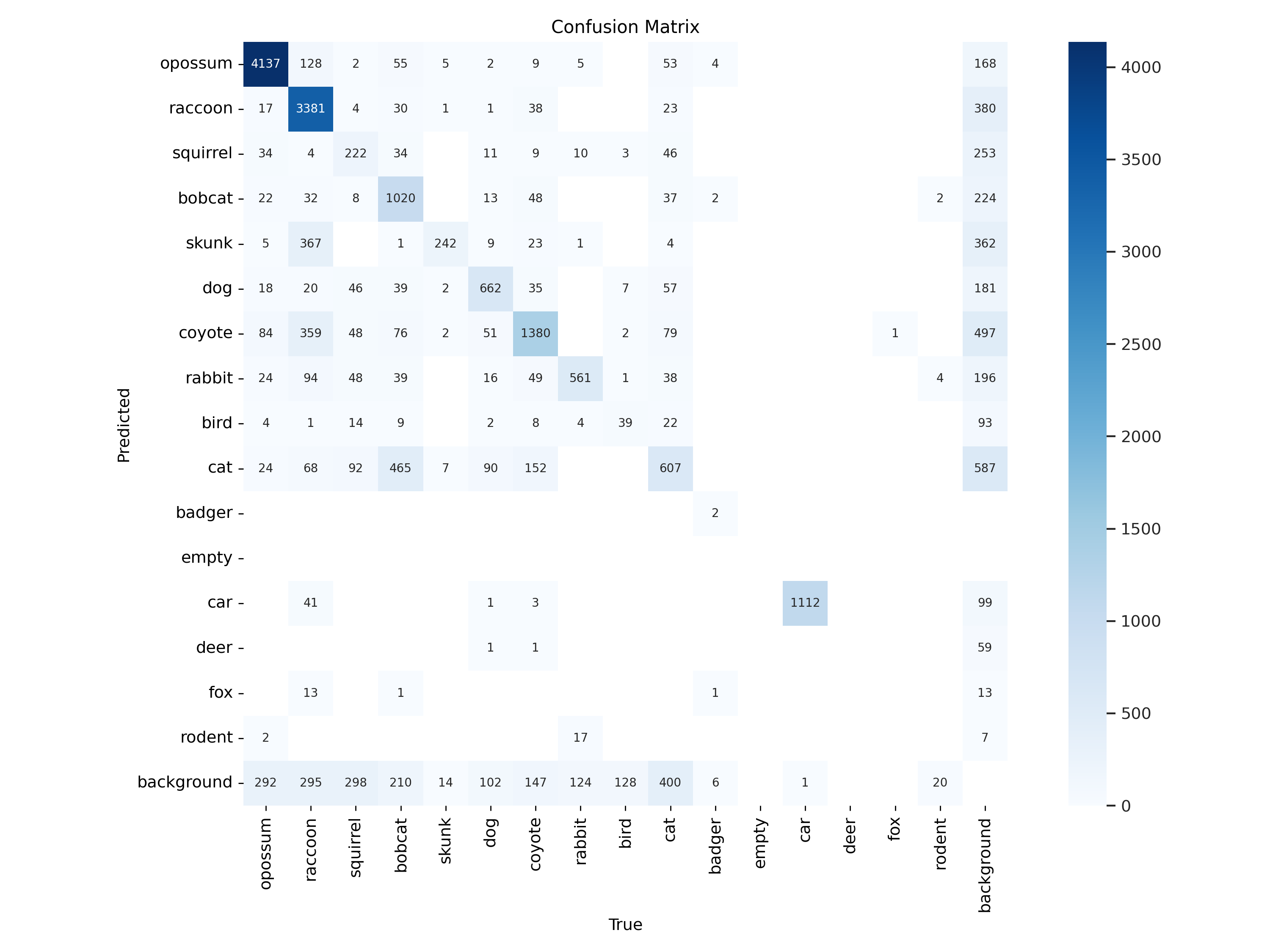}
    \caption{Confusion Matrix for YOLOv8s baseline model for Trans-Test set}
    \label{fig:baseline-trans-confusion-matrix}
\end{figure}

\begin{figure}[H]
    \centering
    \includegraphics[scale=0.47]{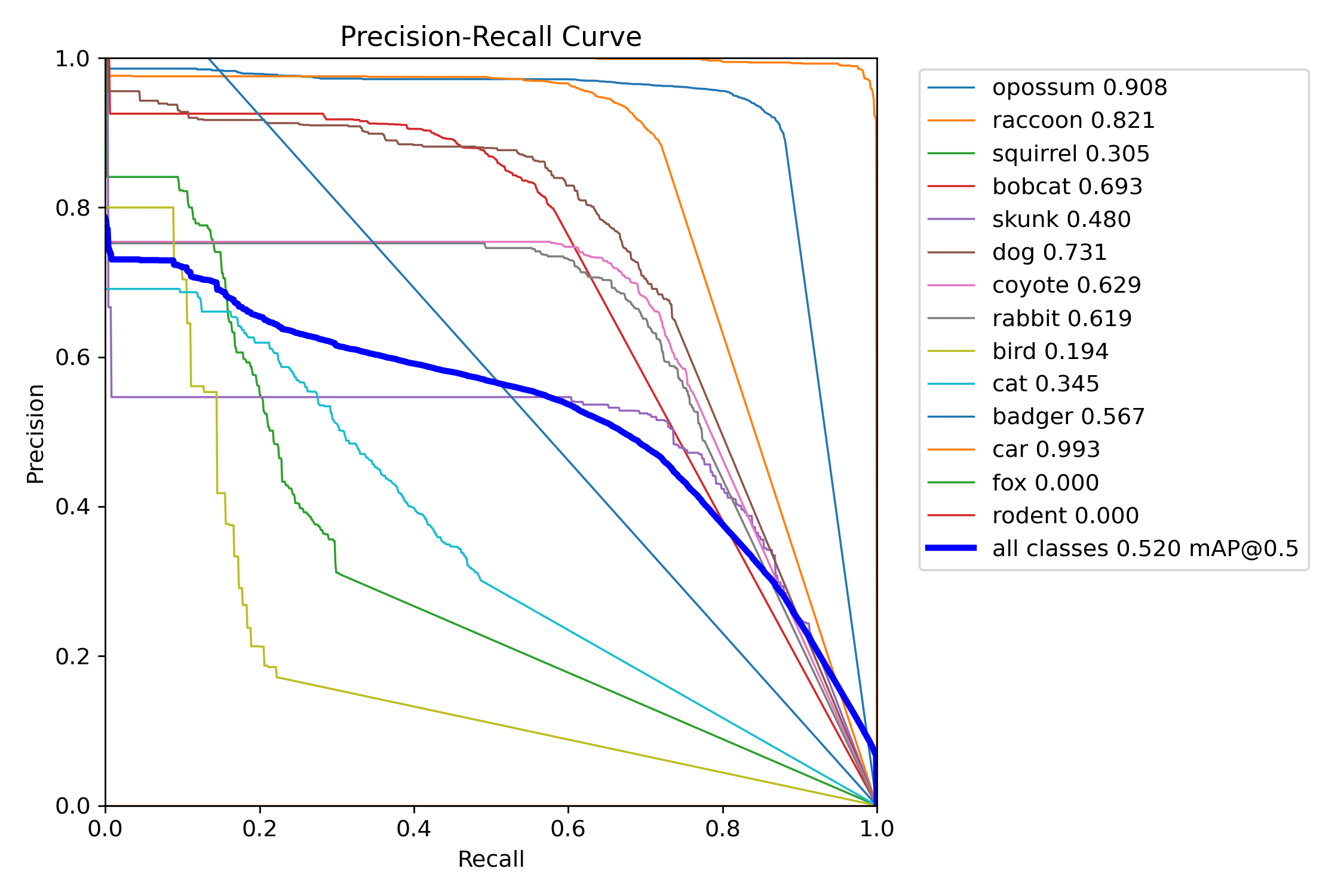}
    \caption{Precision-Recall curve for YOLOv8s baseline model for Trans-Test set}
    \label{fig:baseline-trans-pr-curve}
\end{figure}

\section{Evaluation of improved YOLOv8s model}
On training and evaluating the improved YOLOv8s model with the above-discussed dataset, the model demonstrates commendable performance as reflected in the following evaluation metrics. These metrics serve as a quantitative indicator, providing insights into the model’s capabilities and effectiveness in handling the given dataset and highlighting the efficacy of the improved YOLOv8s model in accurately detecting and localizing objects within the specified context.

\begin{table}[H]
  \centering
  \resizebox{0.70\textwidth}{!}{%
  \begin{tabular}{|c|c|c|c|}
    \hline
     & Cis-Validation & Cis-Test & Trans-Test \\
    \hline
    Precision (P) & 0.867 & 0.772 & 0.693 \\
    Recall (R) & 0.852 & 0.718 & 0.473\\
    mAP50 & 0.877 & 0.772 & 0.541\\
    mAP50-95 & 0.64 & 0.551 & 0.385\\
    \hline
  \end{tabular}
  }
  \caption{Evaluation metrics for improved YOLOv8s model}
\end{table}

A. Training Plots
\begin{figure}[H]
    \centering
    \includegraphics[scale=0.52]{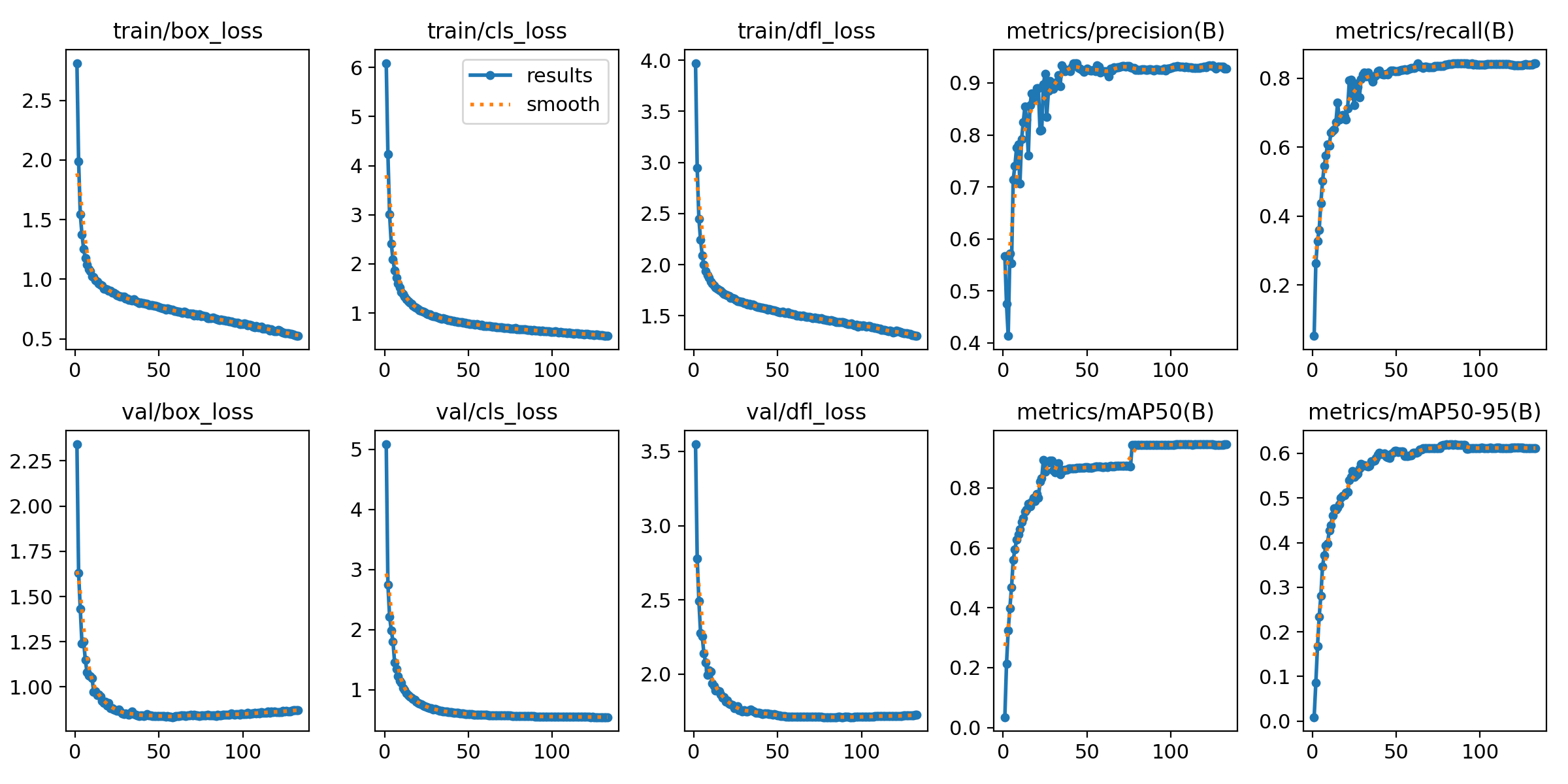}
    \caption{Visualization of different curves for improved YOLOv8s model during training}
    \label{fig:improved-different-curve}
\end{figure}

B. Cis-Test Evaluation Plots 

\begin{figure}[H]
    \centering
    \includegraphics[scale=0.5]{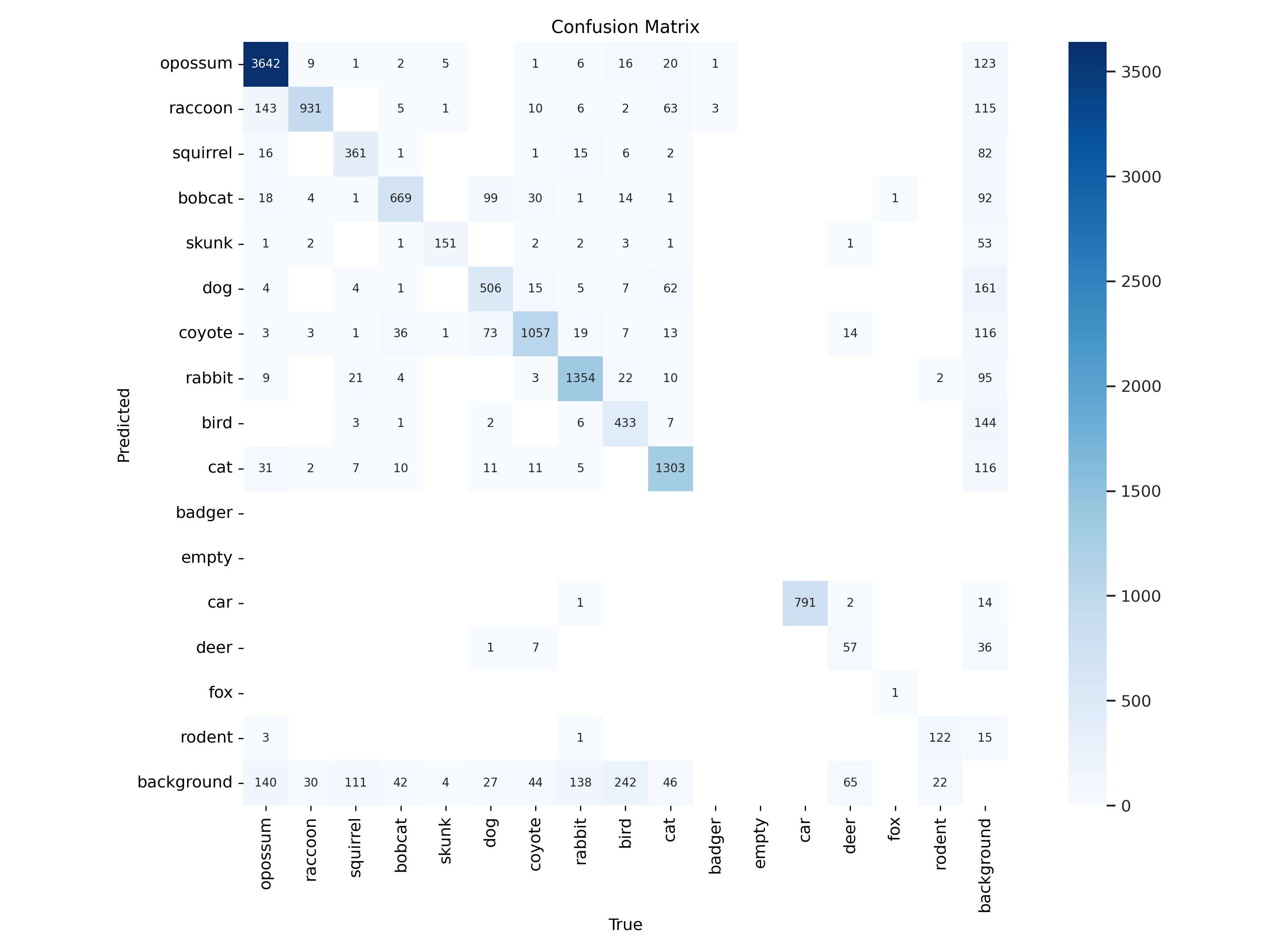}
    \caption{Confusion Matrix for improved YOLOv8s model for Cis-Test set}
    \label{fig:improved-cis-confusion-matrix}
\end{figure}

\begin{figure}[H]
    \centering
    \includegraphics[scale=0.45]{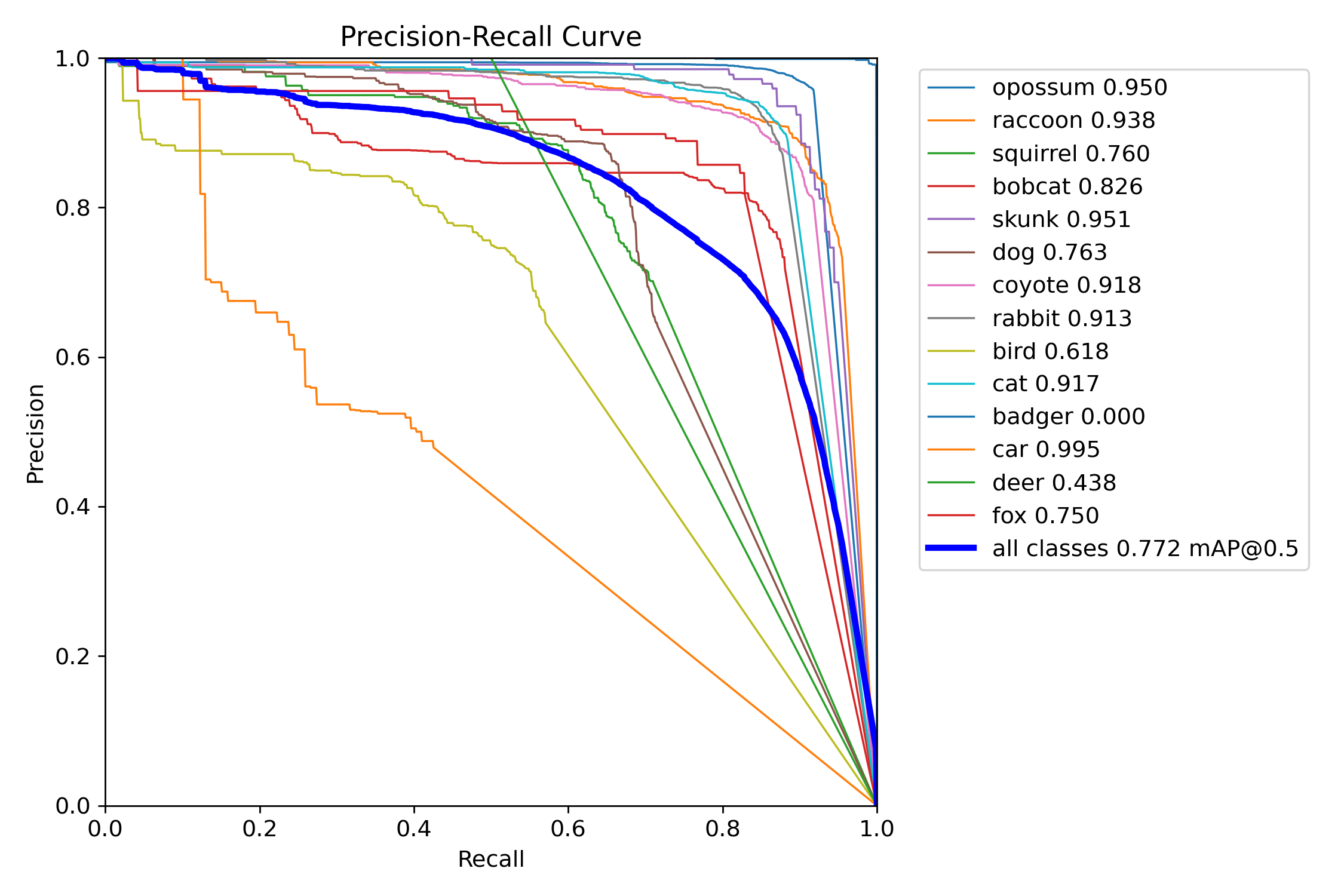}
    \caption{Precision-Recall curve for improved YOLOv8s model for Cis-Test set}
    \label{fig:improved-cis-pr-curve}
\end{figure}

C. Trans-Test Evaluation Plots 

\begin{figure}[H]
    \centering
    \includegraphics[scale=0.5]{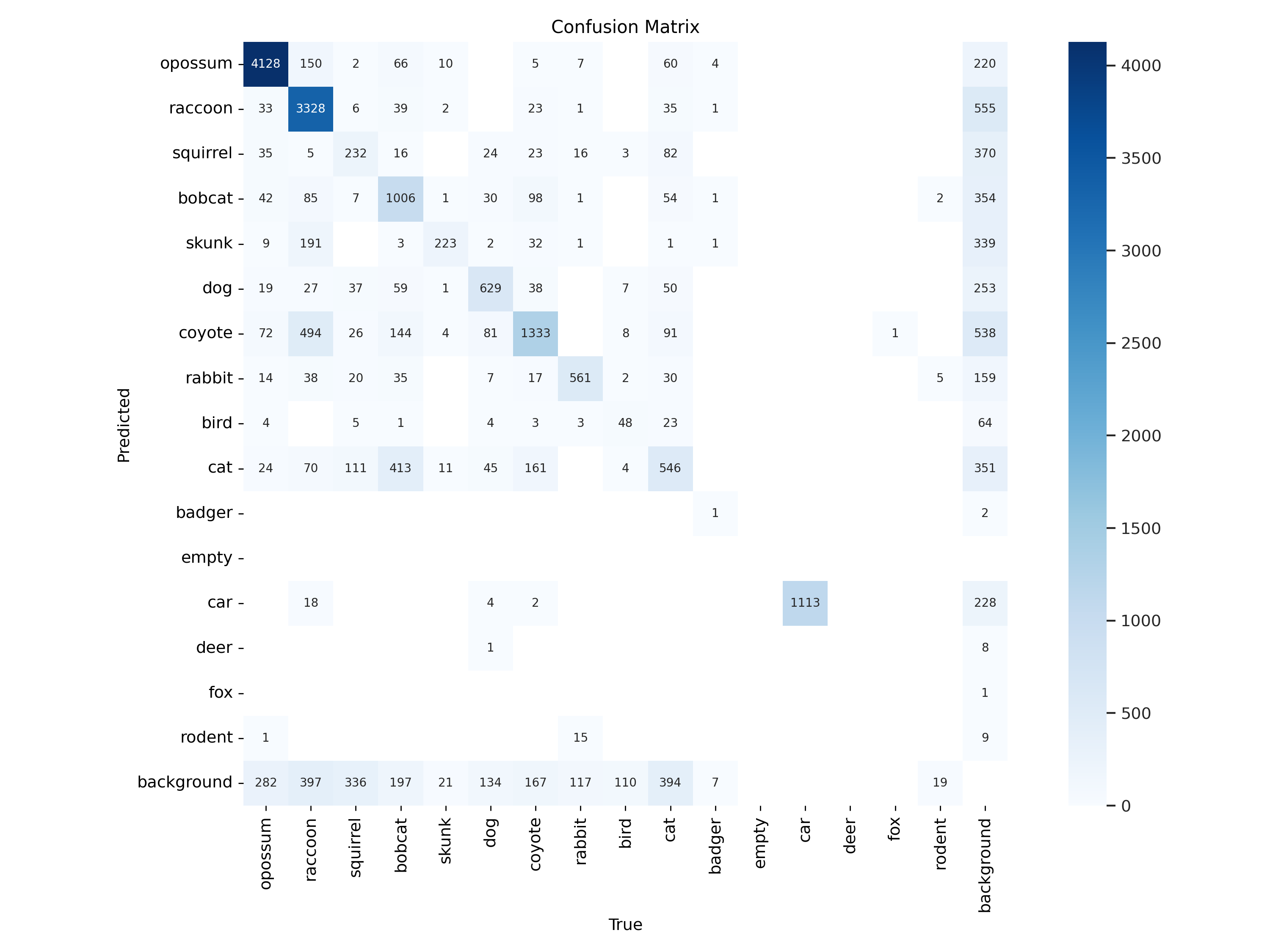}
    \caption{Confusion Matrix for improved YOLOv8s model for Trans-Test set}
    \label{fig:improved-trans-confusion-matrix}
\end{figure}

\begin{figure}[H]
    \centering
    \includegraphics[scale=0.45]{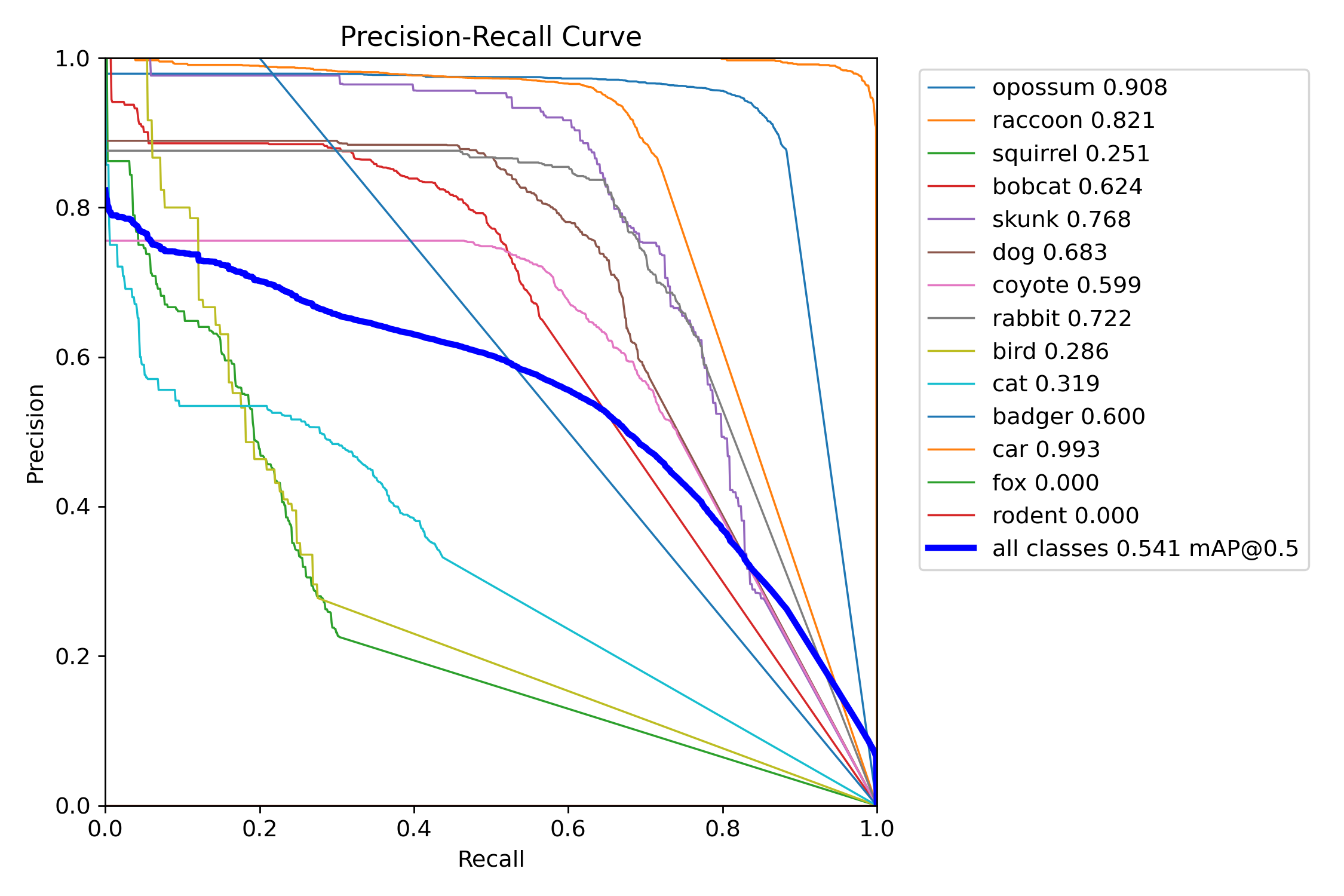}
    \caption{Precision-Recall curve for improved YOLOv8s model for Trans-Test set}
    \label{fig:improved-trans-pr-curve}
\end{figure}
\section{Significance of plots}
\begin{enumerate}
    \item The \textbf{train/box\_loss} plot in \ref{fig:baseline-different-curve} and \ref{fig:improved-different-curve} illustrates the relationship between box loss and epochs during the training phase on the dataset. It visually signifies the model's learning process in accurately identifying object locations within images. The declining trend in the loss across epochs suggests ongoing improvement in the model's proficiency in pinpointing the precise locations of objects.
    
    \item The \textbf{train/class\_loss} plot in \ref{fig:baseline-different-curve} and \ref{fig:improved-different-curve} displays the relationship between class loss and epochs during the training of the model on the dataset. It visually represents how well the model is performing in terms of classification as it undergoes training. The decreasing trend in class loss on the plot indicates that the model is getting better at accurately classifying the objects within the bounding boxes.
    
    \item The \textbf{train/dfl\_loss} plot in \ref{fig:baseline-different-curve} and \ref{fig:improved-different-curve} illustrates the relationship between distribution focal loss (DFL) and epochs during the training of the model on the dataset. It monitors the model's performance, with a specific emphasis on using Distribution Focal Loss to address imbalanced datasets often encountered in object detection tasks. The decreasing trend in dfl loss on the plot indicates that the model is getting better at handling differences in class label distributions.
    
    \item The \textbf{metrics/precision} plot in \ref{fig:baseline-different-curve} and \ref{fig:improved-different-curve} depicts the precision values of the model over various confidence thresholds for each epoch. The plot reveals an upward trend, suggesting that the model is becoming more proficient at accurately predicting both the location and class category of the detected objects.
    
    \item The \textbf{metrics/recall} plot in \ref{fig:baseline-different-curve} and \ref{fig:improved-different-curve} illustrates the recall values of the model over different confidence thresholds for each epoch. The plot exhibits a rising trend, signifying that the model is improving its capability to accurately predict the location and class category of the objects truly present in the image.

    \item The \textbf{val/box\_loss} plot in \ref{fig:baseline-different-curve} and \ref{fig:improved-different-curve}, showcasing box loss versus epochs on the validation set, offers insights into how well the model localizes objects during the validation phase. The decreasing trend in the loss indicates improving performance in accurately determining the location of objects.

    \item The \textbf{val/cls\_loss} plot in \ref{fig:baseline-different-curve} and \ref{fig:improved-different-curve}, depicting class loss versus epochs on the validation set, offers insights into the model's classification performance during validation. The decreasing trend in the loss suggests an improvement in the model's ability to correctly categorize objects during this validation phase. 

    \item The \textbf{val/dfl\_loss} plot in \ref{fig:baseline-different-curve} and \ref{fig:improved-different-curve}, which tracks distribution focal loss (DFL) against epochs on the validation set, is designed to address imbalanced datasets frequently encountered in object detection tasks. It provides a specific focus on the application of Distribution Focal Loss during the validation phase.

    \item The \textbf{metrics/mAP50} plot in \ref{fig:baseline-different-curve} and \ref{fig:improved-different-curve} illustrates the progression of mAP50 values across epochs, offering a visual representation of the mean Average Precision at an IoU threshold of 0.5 during the evaluation phase.

    \item The \textbf{metrics/mAP50-95} plot in \ref{fig:baseline-different-curve} and \ref{fig:improved-different-curve} charts the increase in mAP50-95 values over epochs, delivering a thorough evaluation of object detection performance by encompassing a spectrum of IoU thresholds.\\ \\
    Both mAP50 and mAP50-95 are on the rising trend indicating the ability of the model to localize and categorize objects in the image effectively across different confidence thresholds.

    \item \textbf{Precision-Recall curve}\\
     Precision-Recall curve in \ref{fig:baseline-cis-pr-curve} and \ref{fig:baseline-trans-pr-curve} for Cis-Test and Trans-Test dataset using the baseline model and \ref{fig:improved-cis-pr-curve} and \ref{fig:improved-trans-pr-curve} for Cis-Test and Trans-Test dataset using the improved model provides a more nuanced view of a model's trade-off between precision and recall at different confidence thresholds. The area under the PR curve (AUC-PR) provides a single scalar value summarizing the model's overall performance across various confidence thresholds. A higher AUC-PR indicates better performance.

\end{enumerate}

\section{Heatmaps Visualization}
For the baseline YOLOv8s model, on evaluating the heatmaps specifically generated for the end layer 21 of the model using the Grad-CAM technique, it revealed that during the model’s predictions, there is an apparent activation of background features in addition to the targeted object-related activations. This insight implies that, at the specified layer, the model might be responding not only to the features associated with the objects of interest but also to the background elements.  Layer 21 is selected for heatmap visualization because it is the end layer and has greater feature diversity. \\

Few images of the Trans-Test dataset is selected to overlay the heatmap and find out the regions of focus during model predictions.

\begin{figure}[H]
    \begin{minipage}{0.5\textwidth}
        \centering
        \includegraphics[width=\linewidth]{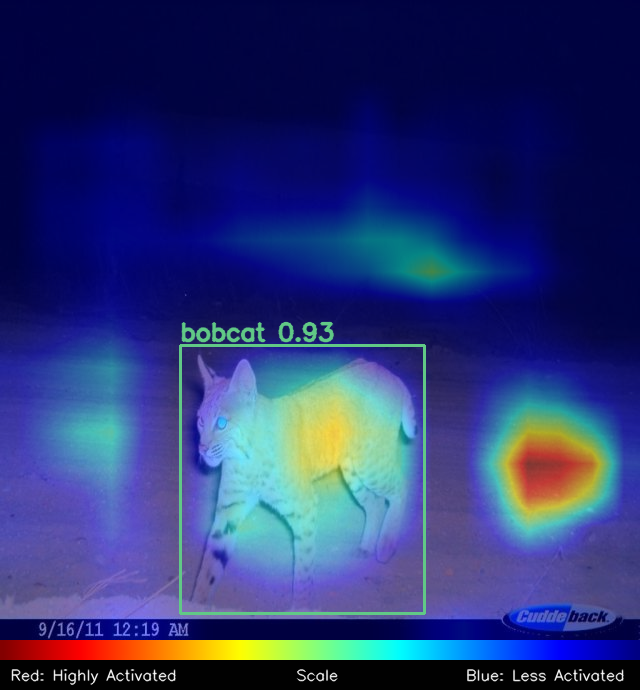}
    \end{minipage}%
    \begin{minipage}{0.5\textwidth}
        \centering
        \includegraphics[width=\linewidth]{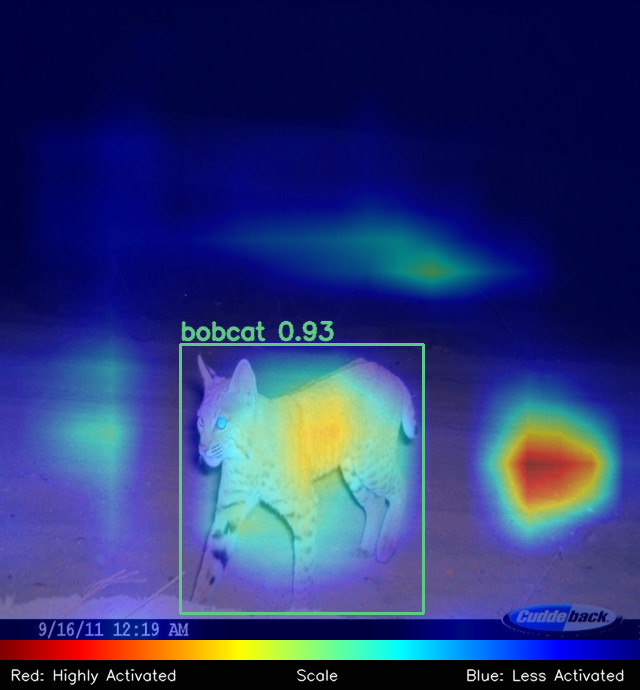}
    \end{minipage}
    
    \medskip
    
    \begin{minipage}{0.5\textwidth}
        \centering
        \includegraphics[width=\linewidth]{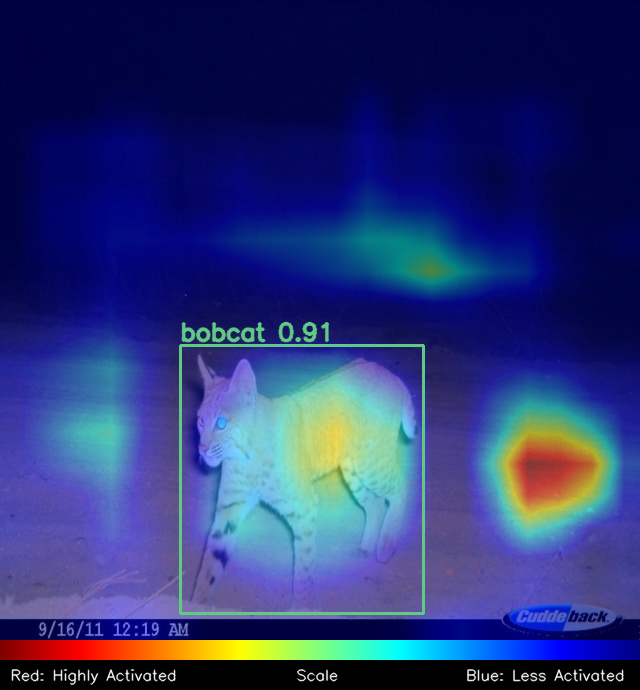}
    \end{minipage}%
    
    \caption{Heatmap visualization of Layer 21 feature maps in baseline YOLOv8s model for a bobcat image from Trans-Test set}
\end{figure}

\begin{figure}[H]
    \begin{minipage}{0.52\textwidth}
        \centering
        \includegraphics[width=\linewidth]{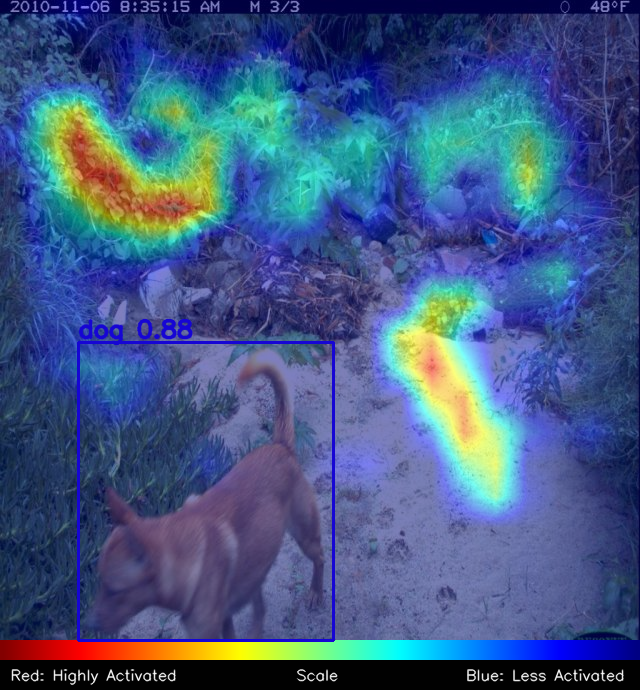}
    \end{minipage}%
    \begin{minipage}{0.52\textwidth}
        \centering
        \includegraphics[width=\linewidth]{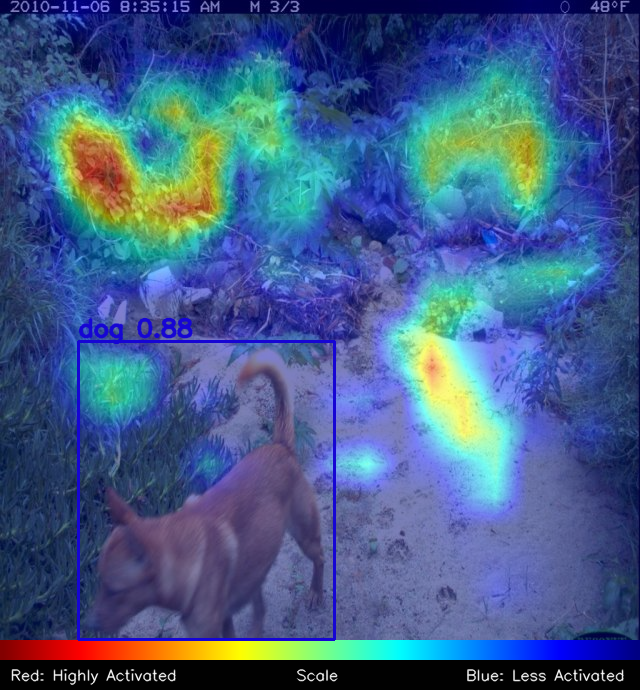}
    \end{minipage}
    
    \medskip
    
    \begin{minipage}{0.52\textwidth}
        \centering
        \includegraphics[width=\linewidth]{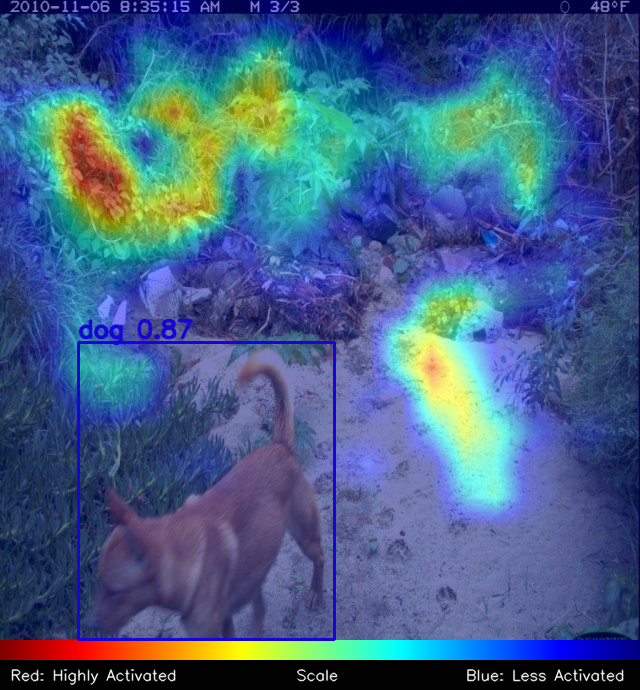}
    \end{minipage}%
    \begin{minipage}{0.52\textwidth}
        \centering
        \includegraphics[width=\linewidth]{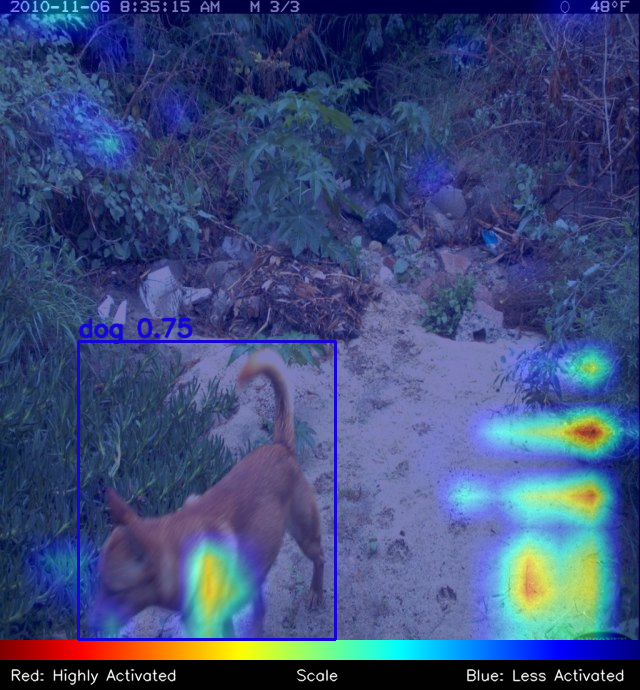}
    \end{minipage}
    
    \caption{Heatmap visualization of Layer 21 feature maps in baseline YOLOv8s model for a dog image from Trans-Test set}
\end{figure}

\begin{figure}[H]
    \begin{minipage}{0.52\textwidth}
        \centering
        \includegraphics[width=\linewidth]{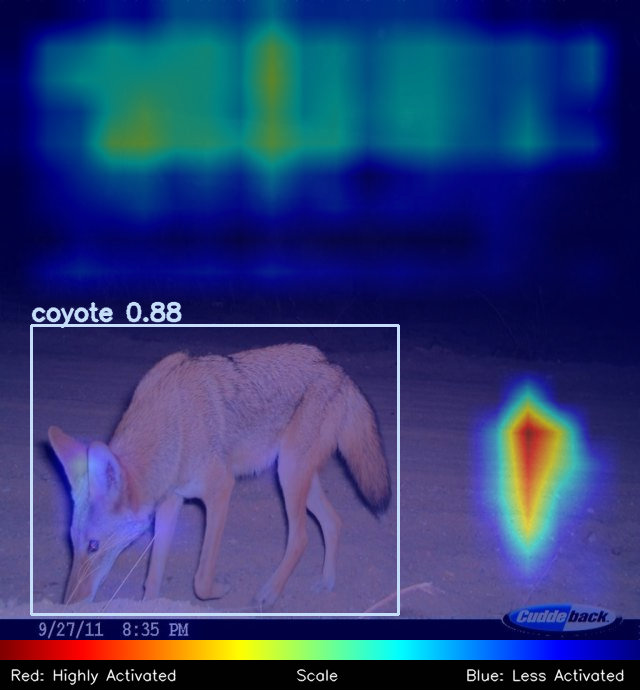}
    \end{minipage}%
    \begin{minipage}{0.52\textwidth}
        \centering
        \includegraphics[width=\linewidth]{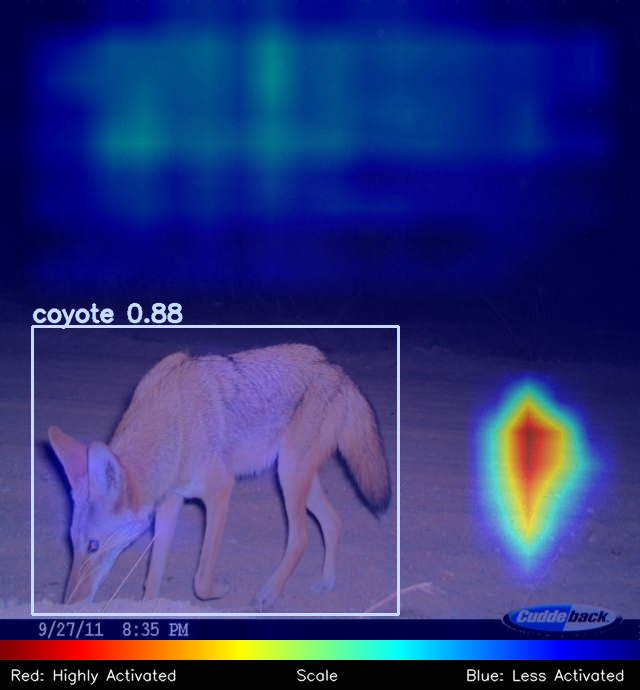}
    \end{minipage}
    
    \medskip
    
    \begin{minipage}{0.52\textwidth}
        \centering
        \includegraphics[width=\linewidth]{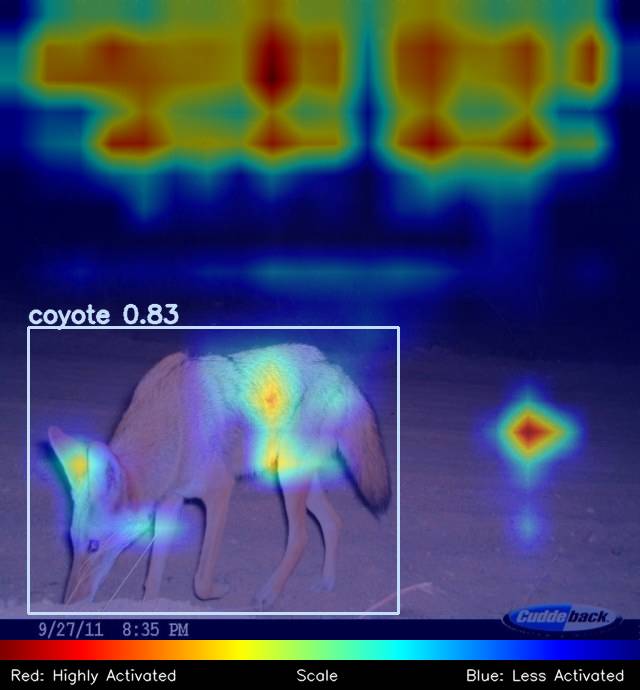}
    \end{minipage}%
    \begin{minipage}{0.52\textwidth}
        \centering
        \includegraphics[width=\linewidth]{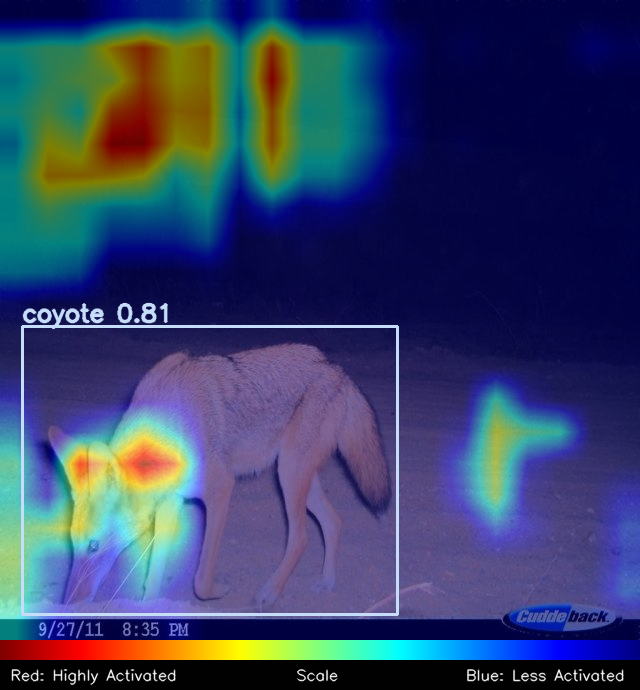}
    \end{minipage}
    
    \caption{Heatmap visualization of Layer 21 feature maps in baseline YOLOv8s model for a coyote image from Trans-Test set}
\end{figure}

\vspace{3cm}
The dual activation discussed above suggested a potential area for refinement in the model’s architecture to enhance its capacity for distinguishing foreground objects from background noise. After the introduction of the Global Attention Mechanism (GAM) module, the improved model is able to suppress the background clutters and is able to focus on the object property which is further validated by the heatmap images produced by Grad-CAM for layer 28 of the improved model. The choice of layer 28 for heatmap visualization is based on its position as the final layer in the network and has greater feature diversity. \\

\begin{figure}[H]
    \begin{minipage}{0.5\textwidth}
        \centering
        \includegraphics[width=\linewidth]{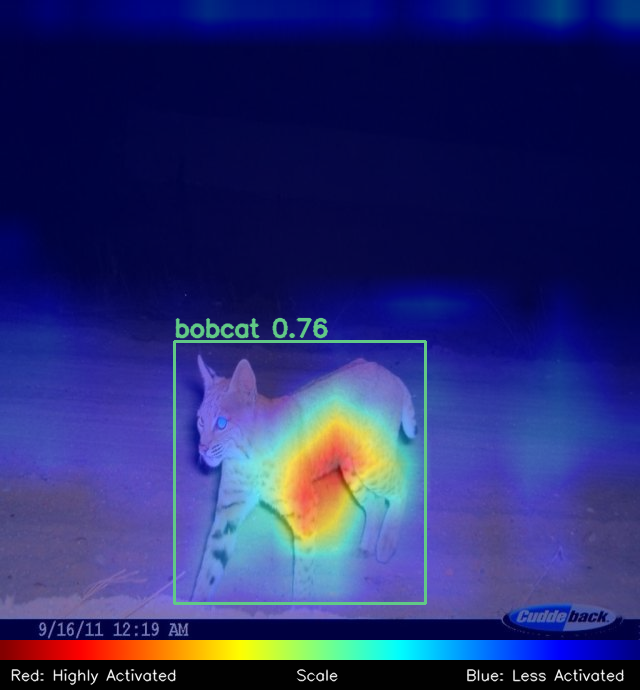}
    \end{minipage}%
    \begin{minipage}{0.5\textwidth}
        \centering
        \includegraphics[width=\linewidth]{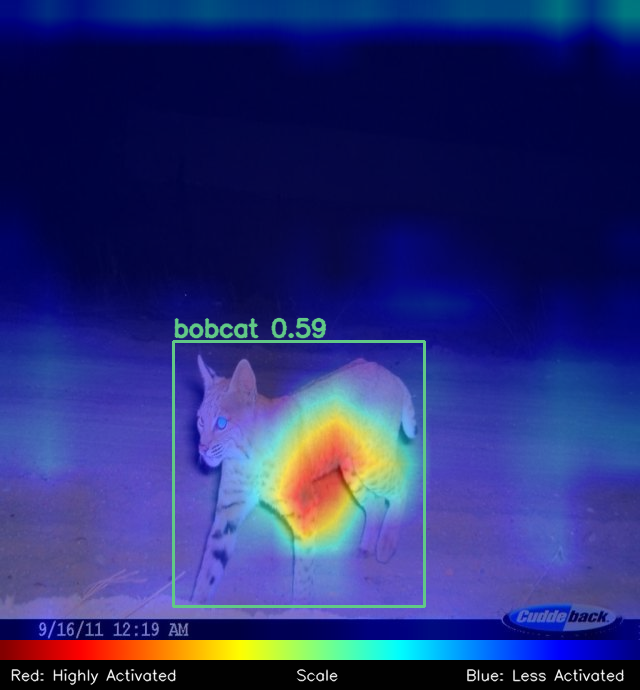}
    \end{minipage}
    
    \medskip
    \begin{minipage}{0.5\textwidth}
        \centering
        \includegraphics[width=\linewidth]{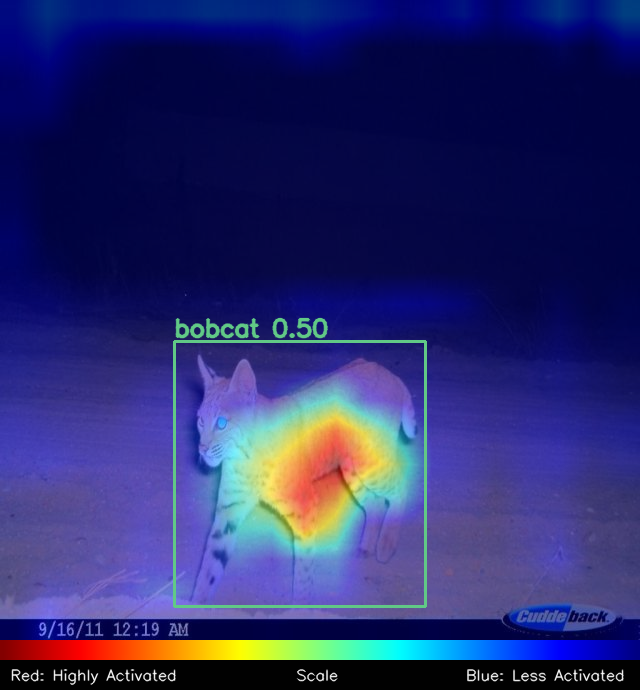}
    \end{minipage}%
    
    \caption{Heatmap visualization of Layer 28 feature maps in improved YOLOv8s model for a bobcat image from Trans-Test set}
\end{figure}

\begin{figure}[H]
    \begin{minipage}{0.5\textwidth}
        \centering
        \includegraphics[width=\linewidth]{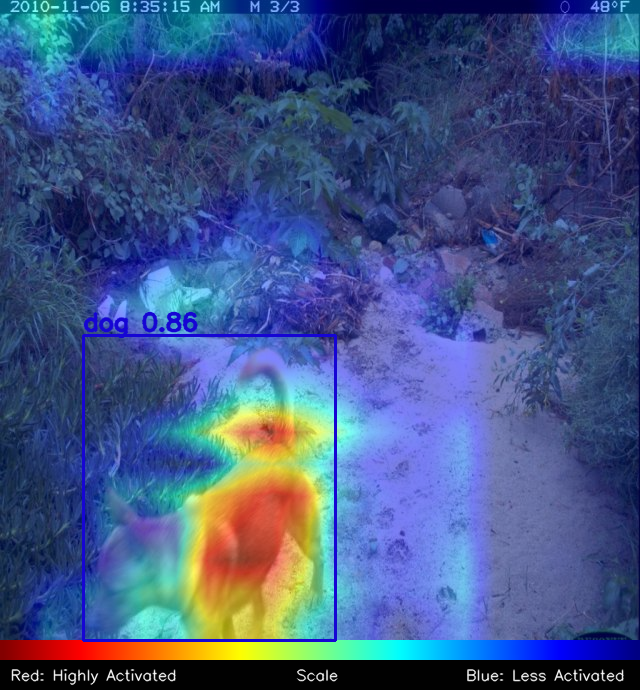}
    \end{minipage}%
    \begin{minipage}{0.5\textwidth}
        \centering
        \includegraphics[width=\linewidth]{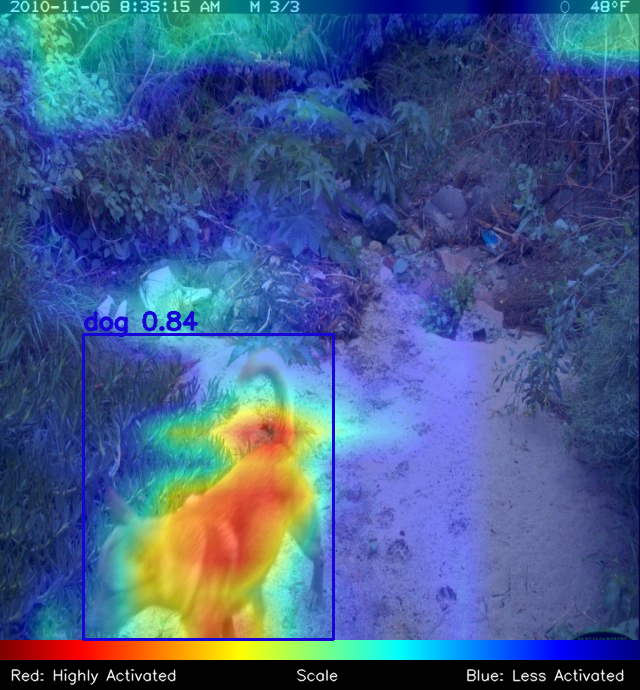}
    \end{minipage}
    
    \medskip
    \begin{minipage}{0.5\textwidth}
        \centering
        \includegraphics[width=\linewidth]{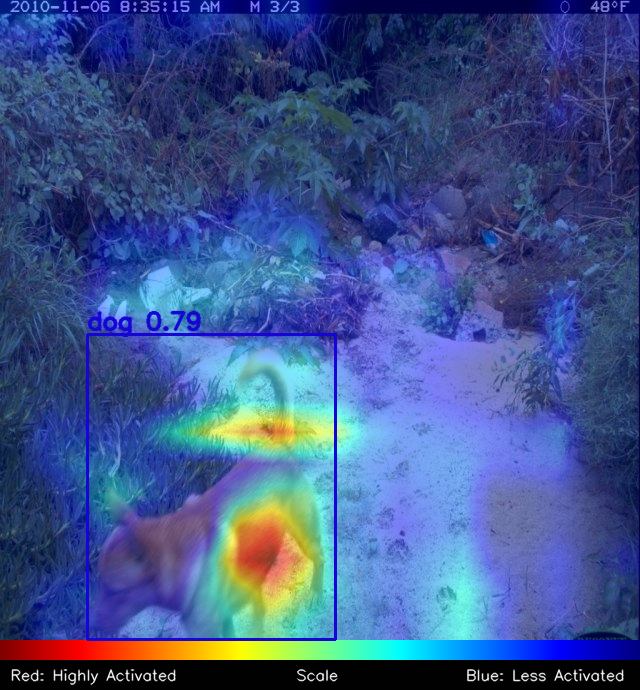}
    \end{minipage}%
    \begin{minipage}{0.5\textwidth}
        \centering
        \includegraphics[width=\linewidth]{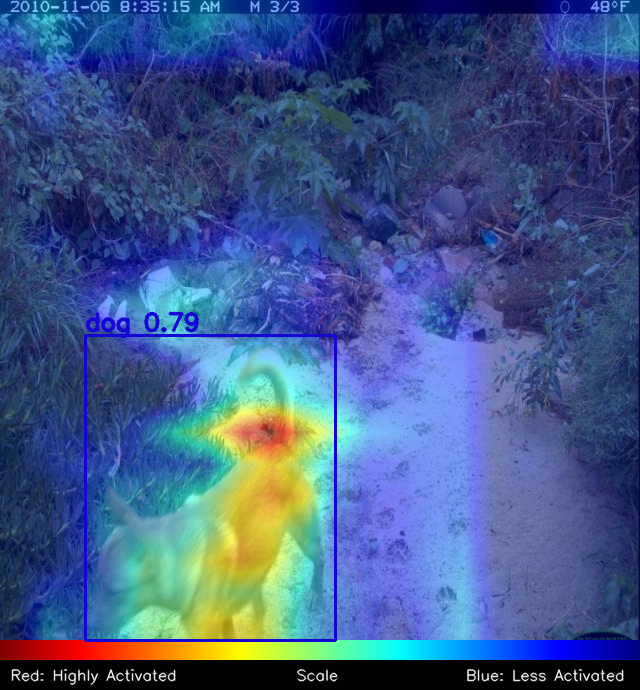}
    \end{minipage}
    
    \caption{Heatmap visualization of Layer 28 feature maps in improved YOLOv8s model for a dog image from Trans-Test set}
\end{figure}

\begin{figure}[H]
    \begin{minipage}{0.5\textwidth}
        \centering
        \includegraphics[width=\linewidth]{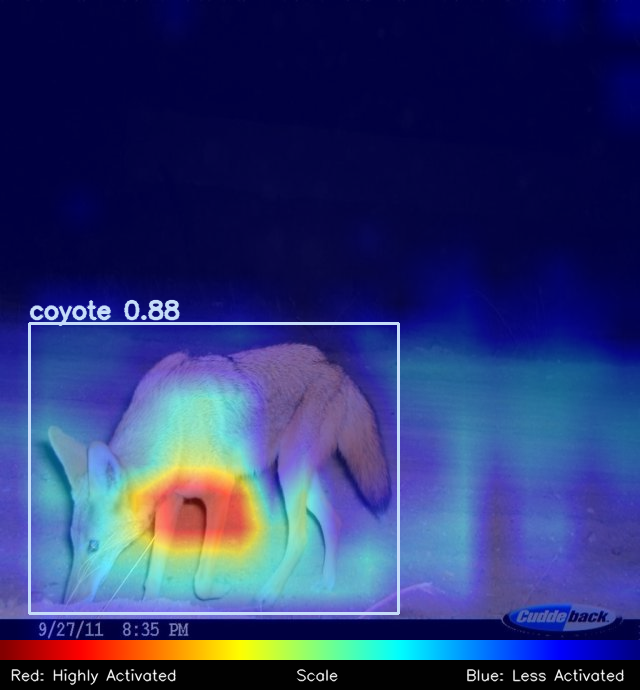}
    \end{minipage}%
    \begin{minipage}{0.5\textwidth}
        \centering
        \includegraphics[width=\linewidth]{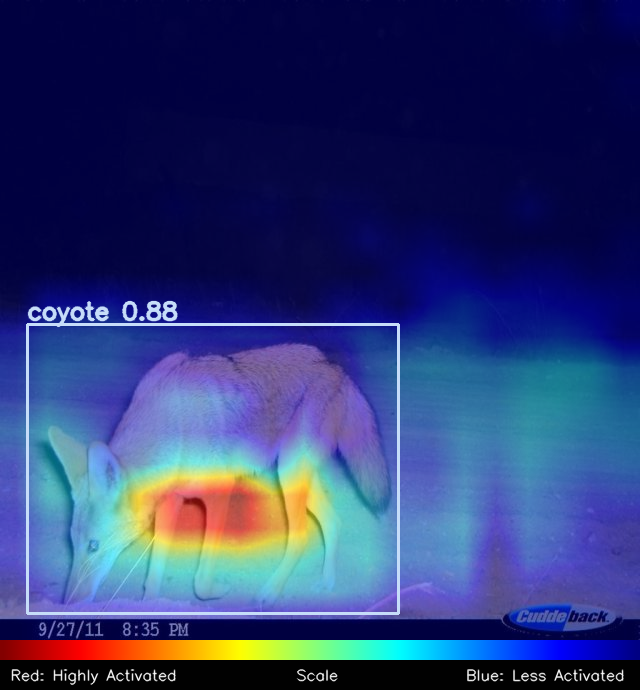}
    \end{minipage}
    
    \medskip
    \begin{minipage}{0.5\textwidth}
        \centering
        \includegraphics[width=\linewidth]{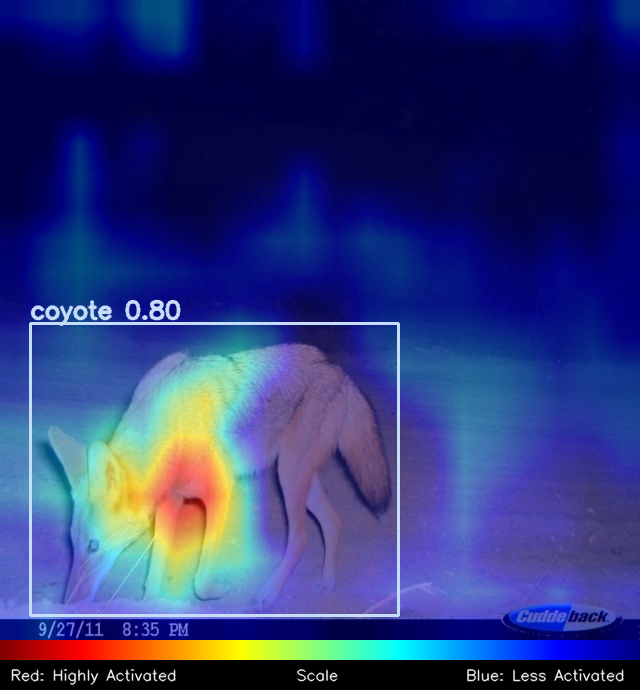}
    \end{minipage}%
    \begin{minipage}{0.5\textwidth}
        \centering
        \includegraphics[width=\linewidth]{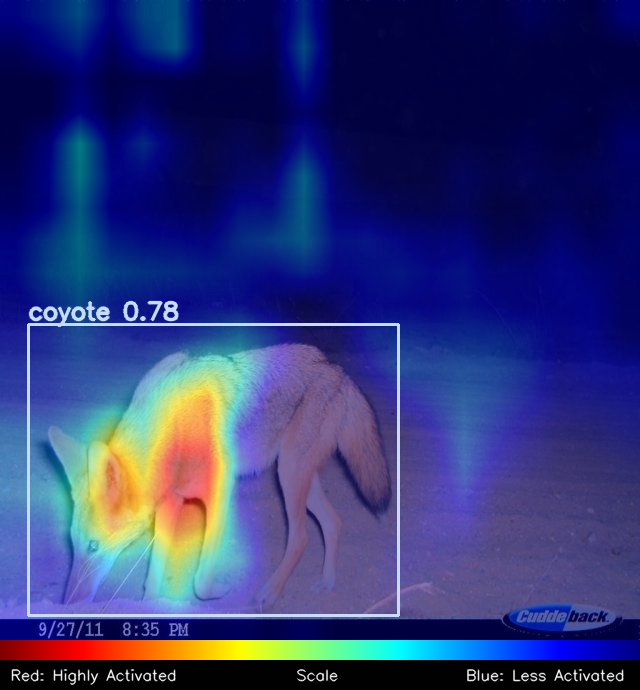}
    \end{minipage}
    
    \caption{Heatmap visualization of Layer 28 feature maps in improved YOLOv8s model for a coyote image from Trans-Test set}
\end{figure}

\section{Ablation Experiments}
To understand the impact of each specific enhancement done to the baseline YOLOv8s model, an ablation experiment was performed to understand the contribution of individual components. The components were systematically disabled to observe how it affected the model’s accuracy, precision, recall, and other performance metrics and it helped in understanding the importance of each specific component and their role in the overall improvement. \\

\begin{table}[H]
  \centering
  \resizebox{0.80\textwidth}{!}{%
  \begin{tabular}{|c|c|c|c|}
    \hline
     & Cis-Validation & Cis-Test & Trans-Test \\
    \hline
    YOLOv8s Baseline  & \textbf{0.889} & 0.813 & 0.52 \\
    YOLOv8s + WIoUv3 & 0.888 & \textbf{0.814} & 0.528\\
    YOLOv8s + GAM Attention & 0.872 & 0.777 & 0.496\\
    YOLOv8s + GAM Attention + WIoUv3 & 0.877 & 0.772 & \textbf{0.541}\\
    \hline
  \end{tabular}
  }
  \caption{Ablation experiments results}
  \label{tab:ablation_exp}
\end{table}

\begin{figure}[H]
    \centering
    \includegraphics[scale=0.60]{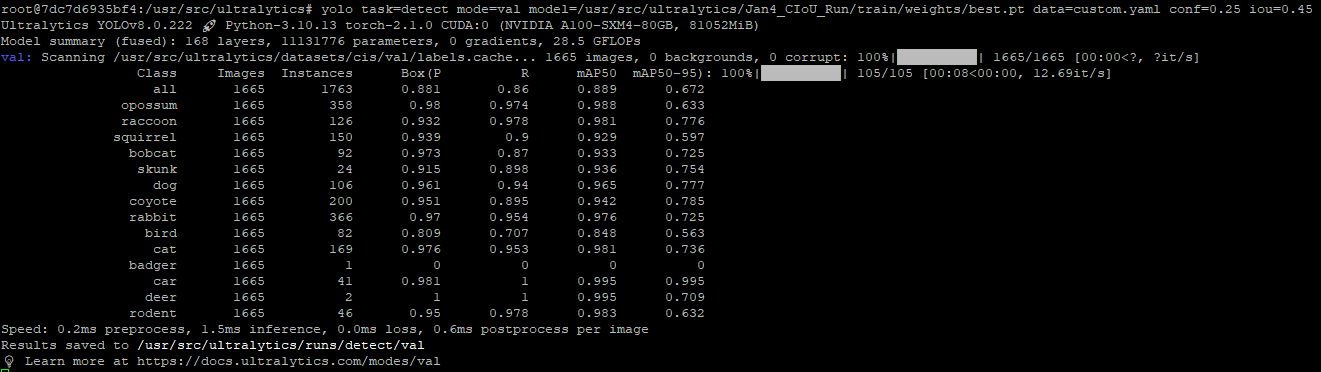}
    \caption{Screenshot of baseline YOLOv8s model performance on the Cis-validation set}
    \label{fig:yolov8-baseline-cis-val}
\end{figure}

\begin{figure}[H]
    \centering
    \includegraphics[scale=0.60]{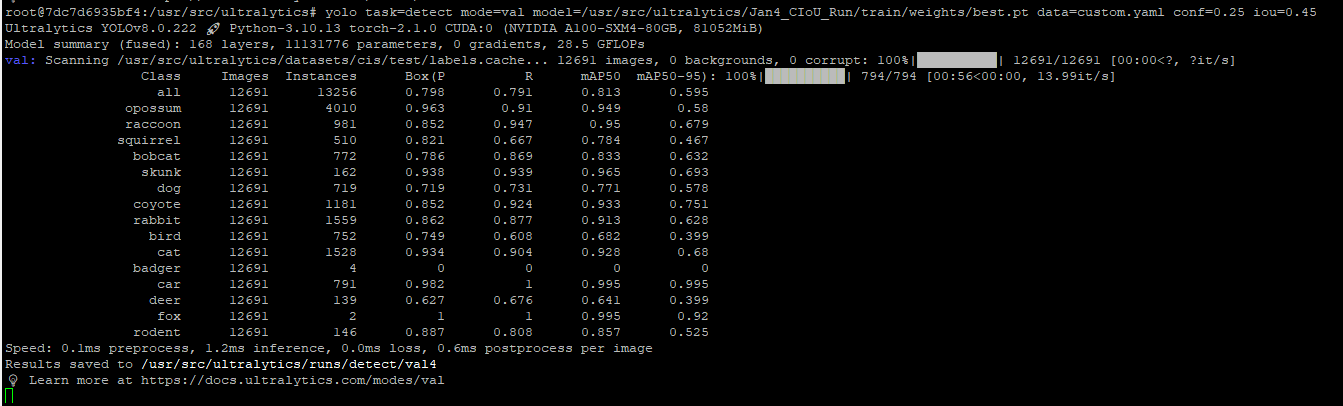}
    \caption{Screenshot of baseline YOLOv8s performance on the Cis-test set}
    \label{fig:yolov8-baseline-cis-test}
\end{figure}

\begin{figure}[H]
    \centering
    \includegraphics[scale=0.60]{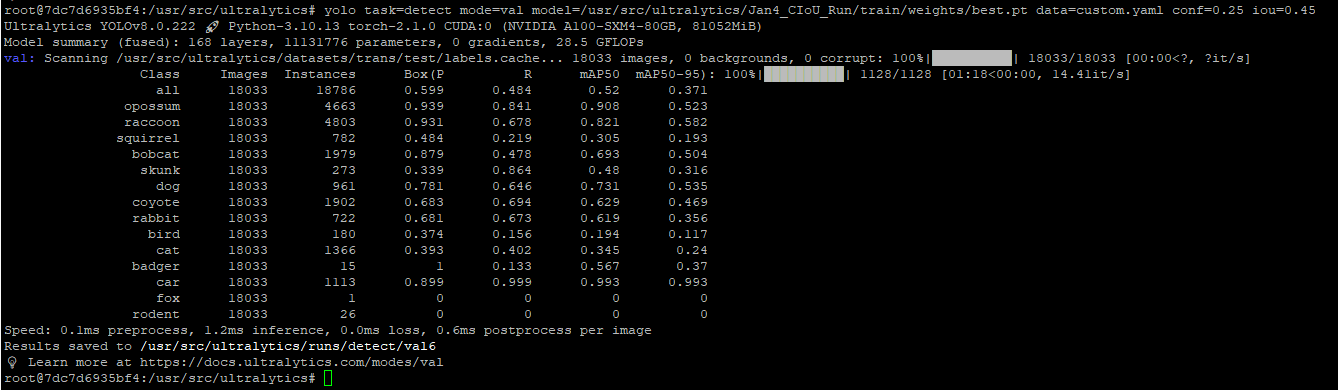}
    \caption{Screenshot of baseline YOLOv8s performance on the Trans-test set}
    \label{fig:yolov8-baseline-trans-test}
\end{figure}

\begin{figure}[H]
    \centering
    \includegraphics[scale=0.58]{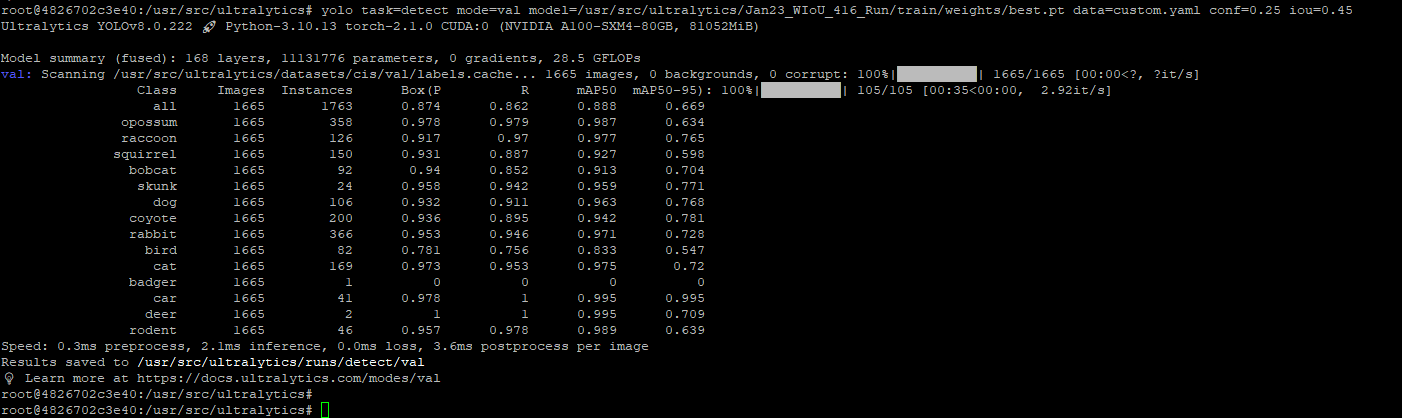}
    \caption{Screenshot of YOLOv8s + WIoUv3 performance on the Cis-validation set}
    \label{fig:yolov8-wiouv3-cis-val}
\end{figure}

\begin{figure}[H]
    \centering
    \includegraphics[scale=0.59]{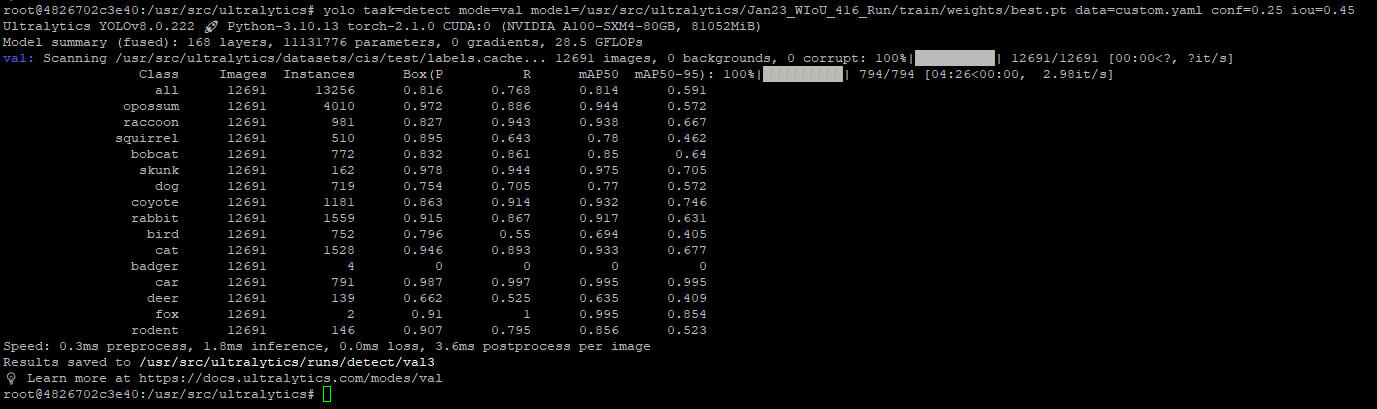}
    \caption{Screenshot of YOLOv8s + WIoUv3 performance on the Cis-test set}
    \label{fig:yolov8-wiouv3-cis-test}
\end{figure}

\begin{figure}[H]
    \centering
    \includegraphics[scale=0.58]{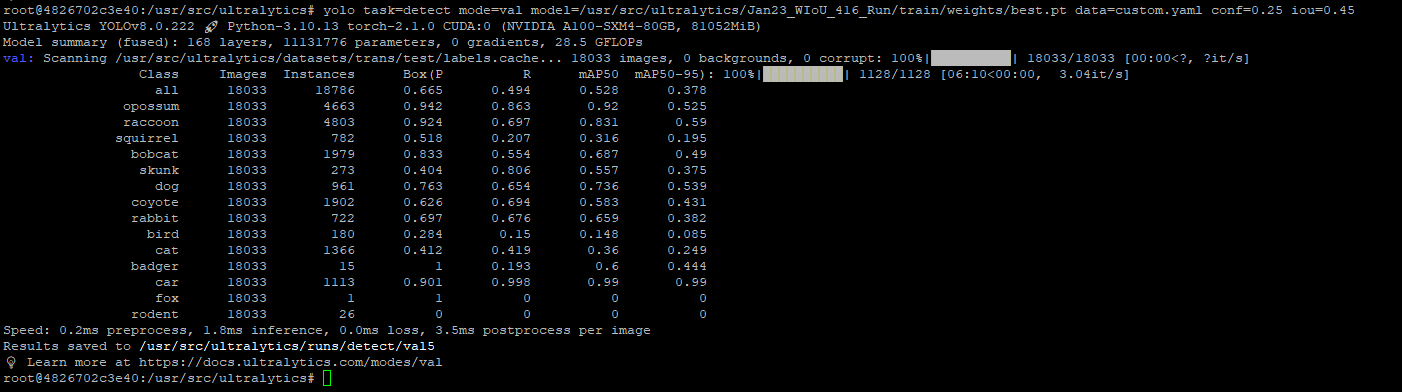}
    \caption{Screenshot of YOLOv8s + WIoUv3 performance on the Trans-test set}
    \label{fig:yolov8-wiouv3-trans-test}
\end{figure}

\begin{figure}[H]
    \centering
    \includegraphics[scale=0.57]{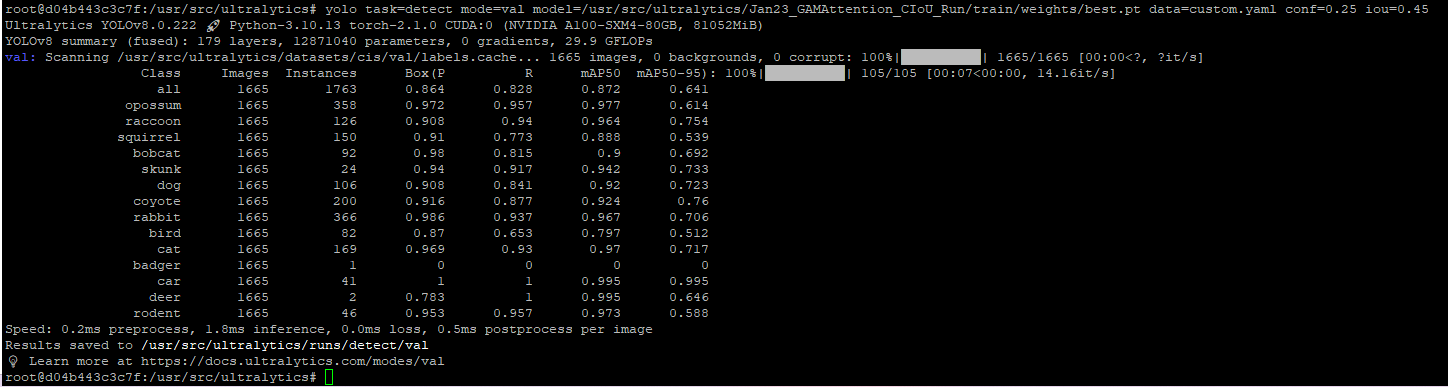}
    \caption{Screenshot of YOLOv8s + GAM Attention performance on the Cis-validation set}
    \label{fig:yolov8-gam-cis-val}
\end{figure}

\begin{figure}[H]
    \centering
    \includegraphics[scale=0.57]{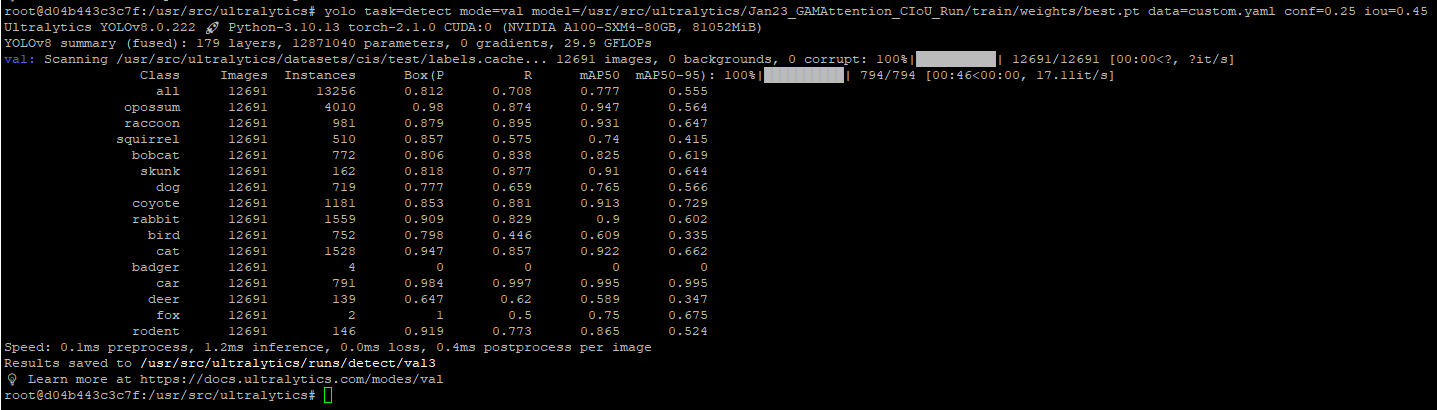}
    \caption{Screenshot of YOLOv8s + GAM Attention performance on the Cis-test set}
    \label{fig:yolov8-gam-cis-test}
\end{figure}

\begin{figure}[H]
    \centering
    \includegraphics[scale=0.57]{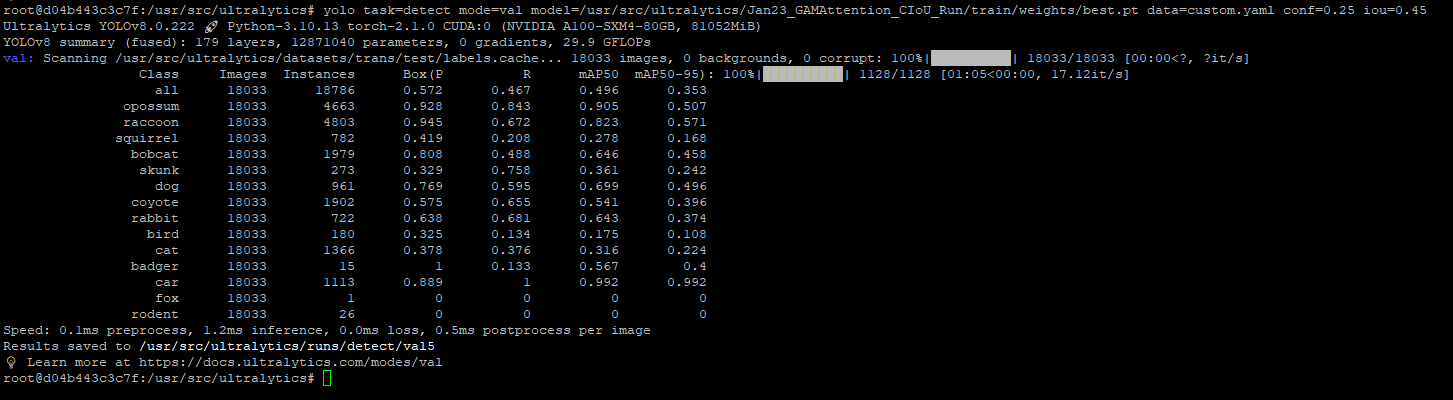}
    \caption{Screenshot of YOLOv8s + GAM Attention performance on the Trans-test set}
    \label{fig:yolov8-gam-trans-test}
\end{figure}

\begin{figure}[H]
    \centering
    \includegraphics[scale=0.58]{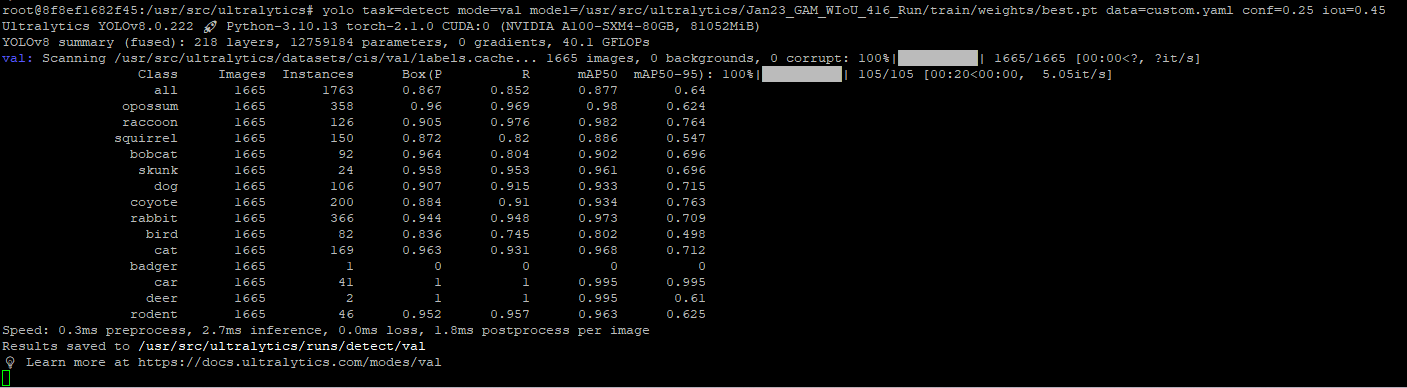}
    \caption{Screenshot of YOLOv8s + GAM Attention + WIoUv3  performance on the Cis-validation set}
    \label{fig:yolov8-combined-cis-val}
\end{figure}

\begin{figure}[H]
    \centering
    \includegraphics[scale=0.58]{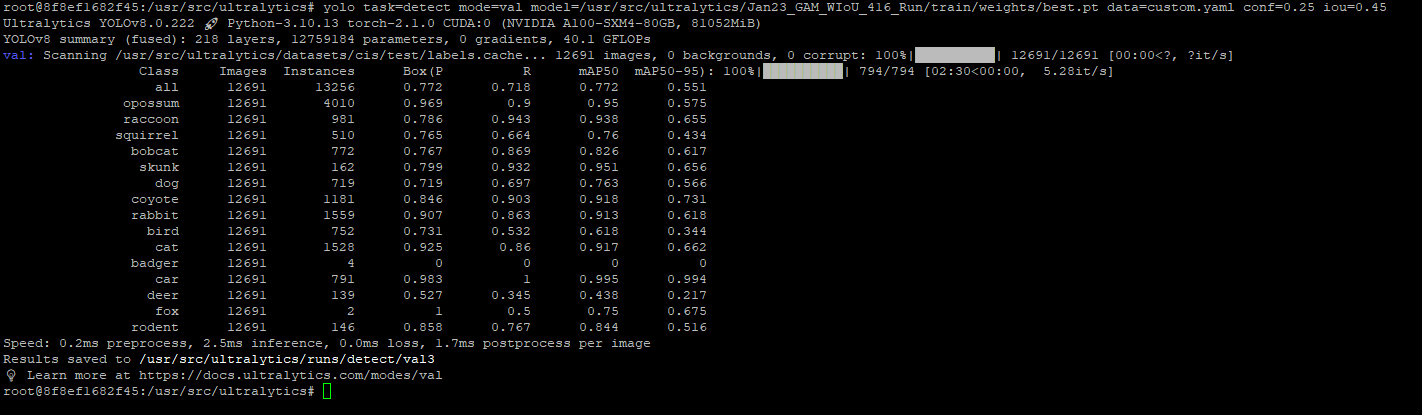}
    \caption{Screenshot of YOLOv8s + GAM Attention + WIoUv3  performance on the Cis-test set}
    \label{fig:yolov8-combined-cis-test}
\end{figure}

\begin{figure}[H]
    \centering
    \includegraphics[scale=0.58]{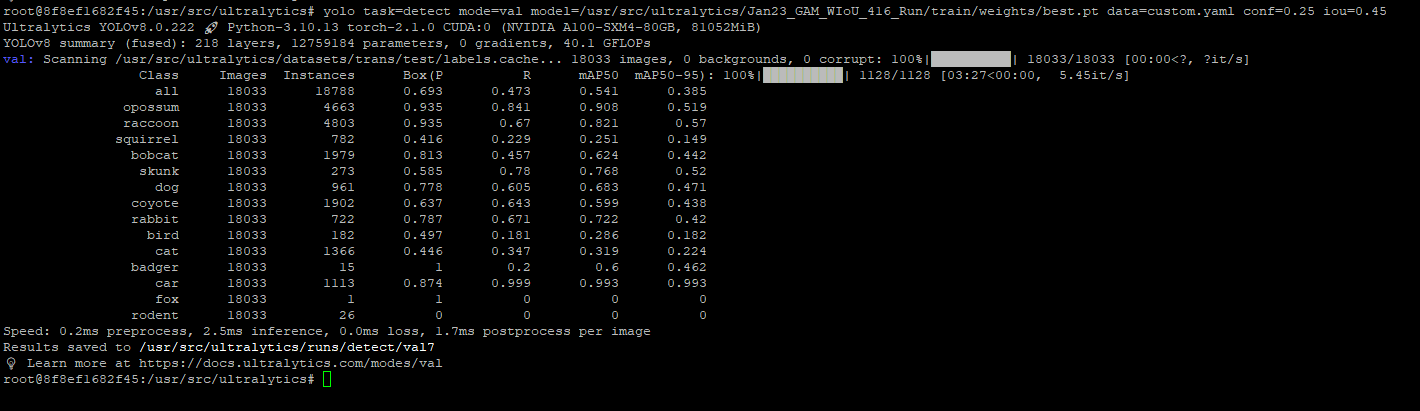}
    \caption{Screenshot of YOLOv8s + GAM Attention + WIoUv3 performance on the Trans-test set}
    \label{fig:yolov8-combined-trans-test}
\end{figure}

\section{Discussions}
From the above table \ref{tab:ablation_exp}, it can be deduced that the baseline model works better for the validation set, the baseline model with integrated WIoUv3 as bounding box regression loss function works better for the Cis-Test set while the improved YOLOv8s model with integrated WIoUv3 as bounding box regression loss function and the Global Attention Mechanism (GAM) module works better in the case of Trans-Test dataset. Although the improved model has a little lower mAP50 evaluation metric for the Cis-Test location, the improved model performs much better in the Trans-Test location’s data and achieves an improvement of 4\% than that of the baseline YOLOv8s model. \\

Given the model performs superiorly on the Trans-Test location dataset i.e. the set of data not encountered during the training and the validation stage, it can be reasonably inferred that this improved model possesses a commendable ability to apply its learned features or learnings to the never-before-seen dataset. This noteworthy observation is indicative of the model’s robust generalization capabilities, wherein it not only excels on familiar training and validation sets but also exhibits the capacity to adapt and perform well on entirely novel data instances. This further affirms the potential of this improved model for broader applicability and utility across diverse and unexplored scenarios. \\

One can argue that the evaluation metric in the Trans-Test location is much less compared to the evaluation metric in the Cis-Test location, and there are several reasons for this:

\begin{enumerate}
    \item \textbf{Number of images}\\
    There is a difference in the number of images in the Cis-Test data and Trans-Test data as shown in Table \ref{tab:dataset-numbers}. The number of images in the Trans-Test location is greater than 42\% compared to the Cis-Test location. Higher the number of images, the greater are the chances of increase in the false positives and incorrect predictions.

    \item \textbf{Uneven distribution of categories}\\
    The disparities in class distributions among the datasets are evident in Figures \ref{fig:train-class-distribution}, \ref{fig:cis-val-class-distribution}, \ref{fig:cis-test-class-distribution}, and \ref{fig:trans-test-class-distribution}. These visualizations highlight uneven category representation across different sets. Notably, certain class labels have fewer instances in the training set, while the test set exhibits a more substantial presence of these categories. If the model has not been sufficiently trained on these underrepresented labels, its ability to accurately detect those objects in real-world scenarios may be compromised.

    \item \textbf{Incorrect annotation present in the test data}\\
    Upon review, it was observed that certain images within the Trans-Test dataset had inaccuracies in their annotations. This discrepancy is likely one of the reasons for the significantly lower evaluation metrics observed on the Trans-Test data.

    \begin{figure}[H]
    \begin{minipage}{0.49\textwidth}
        \centering
        \includegraphics[width=\linewidth]{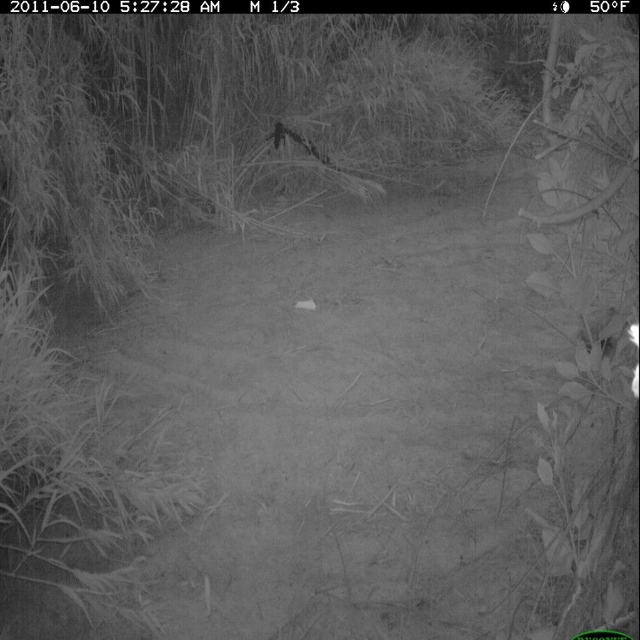}
    \end{minipage}%
    \begin{minipage}{0.5\textwidth}
        \centering
        \includegraphics[width=\linewidth]{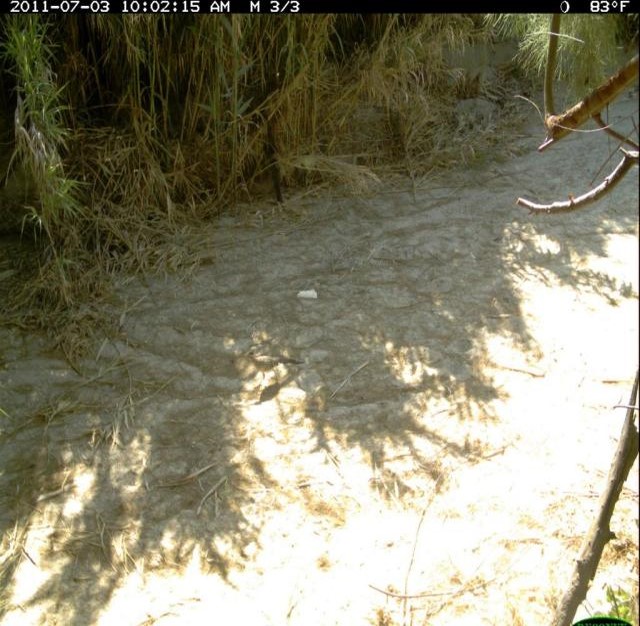}
    \end{minipage}
    
    \medskip
    \begin{minipage}{0.5\textwidth}
        \centering
        \includegraphics[width=\linewidth]{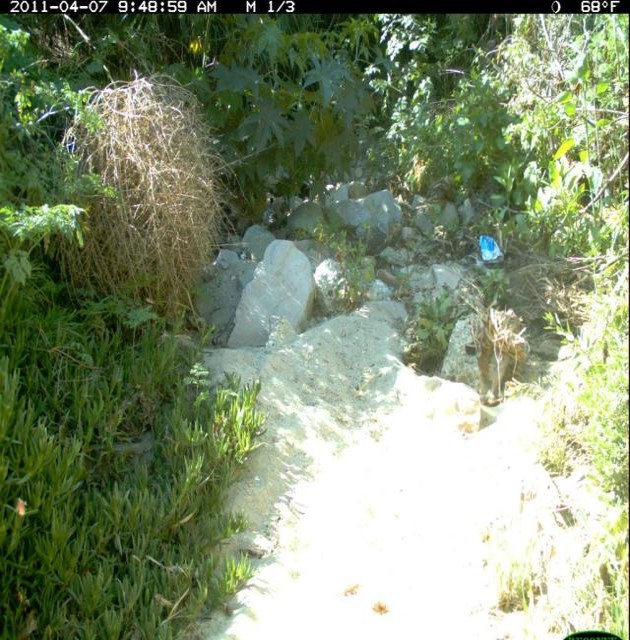}
    \end{minipage}%
    \begin{minipage}{0.5\textwidth}
        \centering
        \includegraphics[width=\linewidth]{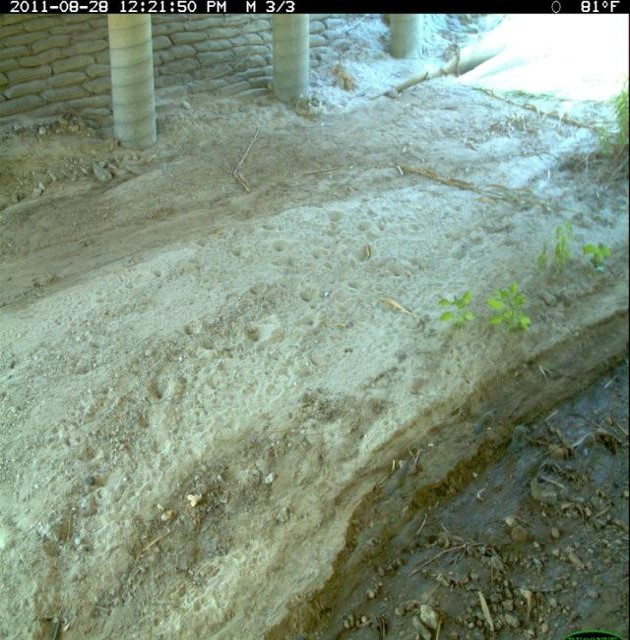}
    \end{minipage}
    
    \caption{Trans-Test data with incorrect annotations}
\end{figure}
    
\end{enumerate}

\section{Evaluation of custom dataset}
Some of the camera trap images of the animals for which the improved model is trained are collected from various sources including the internet and the trained model is then used for the prediction to test the generalization capability of the model.

\begin{figure}[H]
    \begin{minipage}{0.5\textwidth}
        \centering
        \includegraphics[width=\linewidth]{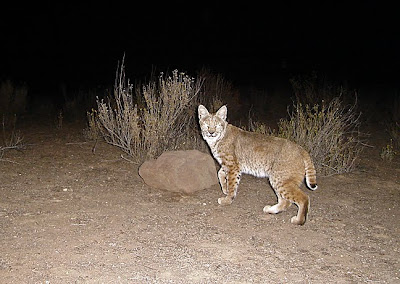}
    \end{minipage}%
    \begin{minipage}{0.5\textwidth}
        \centering
        \includegraphics[width=\linewidth]{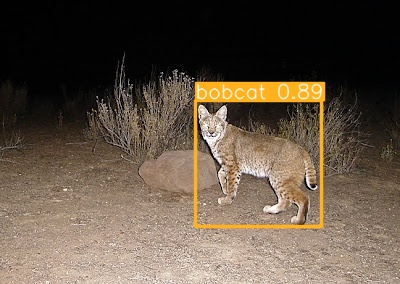}
    \end{minipage}
    
    \medskip
    \begin{minipage}{0.5\textwidth}
        \centering
        \includegraphics[width=\linewidth]{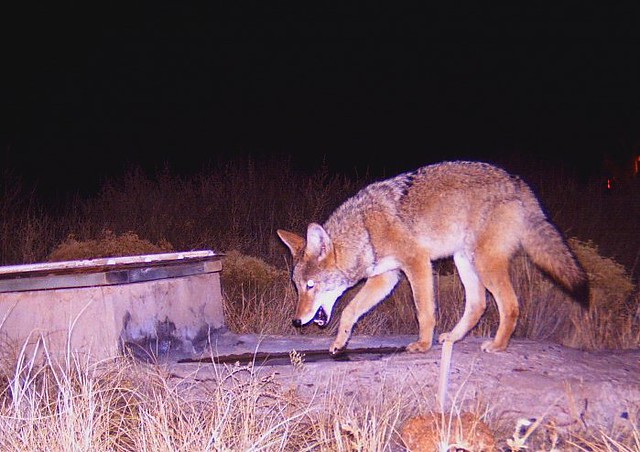}
    \end{minipage}%
    \begin{minipage}{0.5\textwidth}
        \centering
        \includegraphics[width=\linewidth]{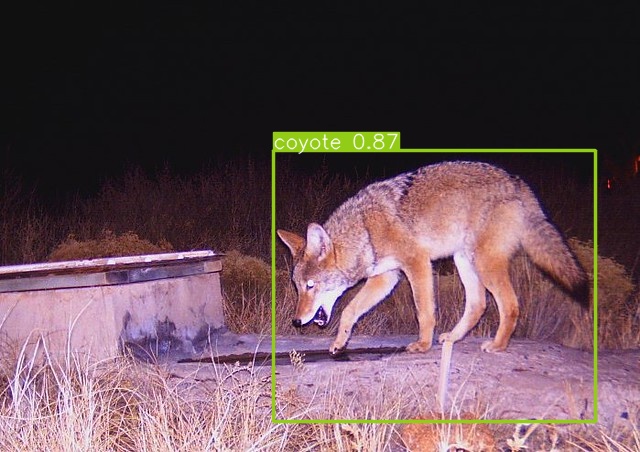}
    \end{minipage}

     \begin{minipage}{0.5\textwidth}
        \centering
        \includegraphics[width=\linewidth]{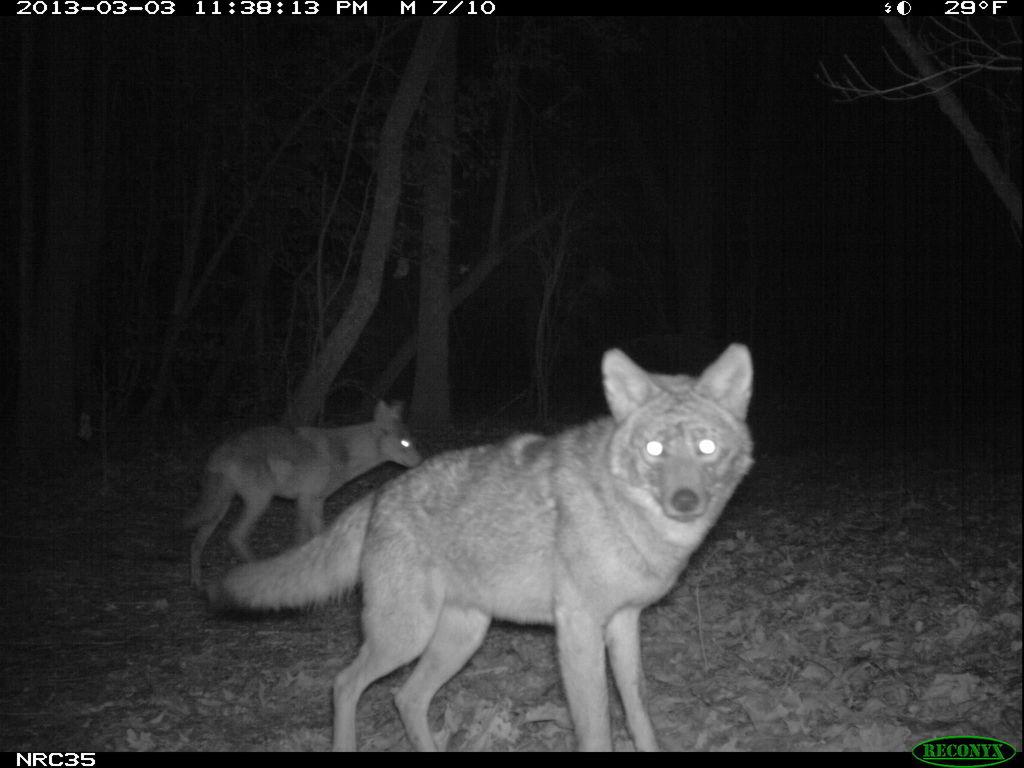}
    \end{minipage}%
    \begin{minipage}{0.5\textwidth}
        \centering
        \includegraphics[width=\linewidth]{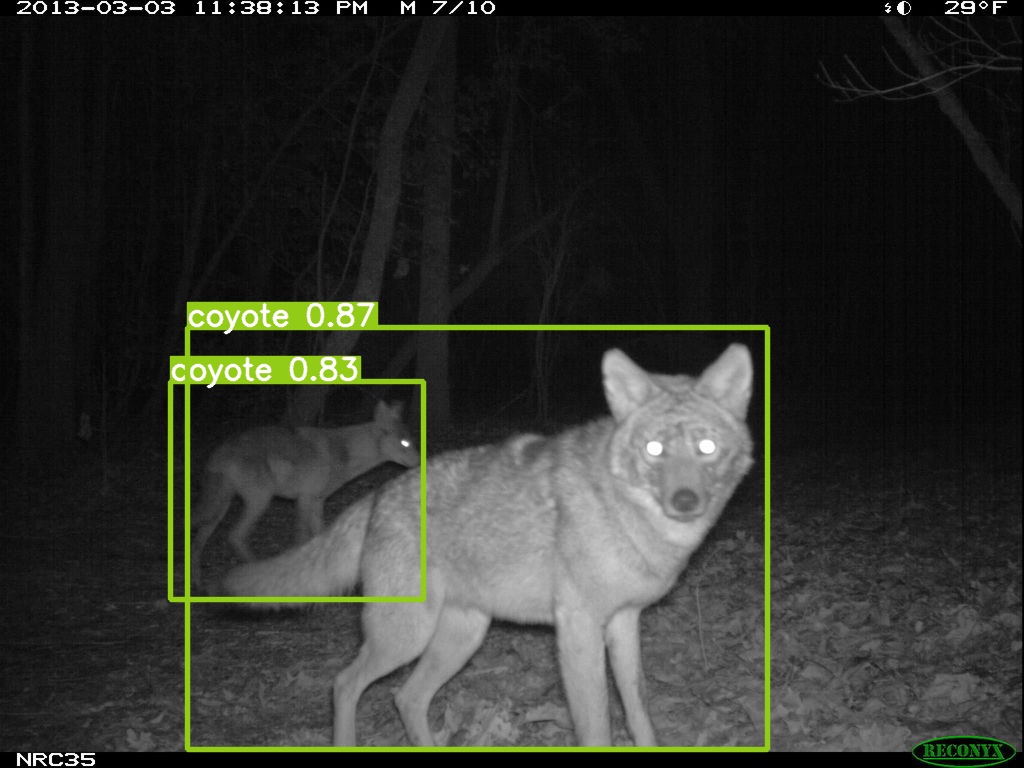}
    \end{minipage}
    
\end{figure}

\begin{figure}[H]
\label{fig:trans-test-improved-model-custom}
   \begin{minipage}{0.5\textwidth}
        \centering
        \includegraphics[width=\linewidth]{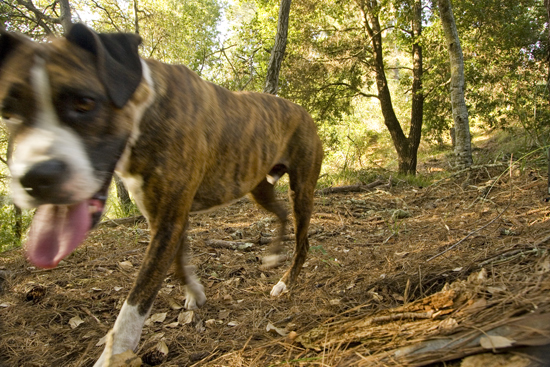}
    \end{minipage}%
    \begin{minipage}{0.5\textwidth}
        \centering
        \includegraphics[width=\linewidth]{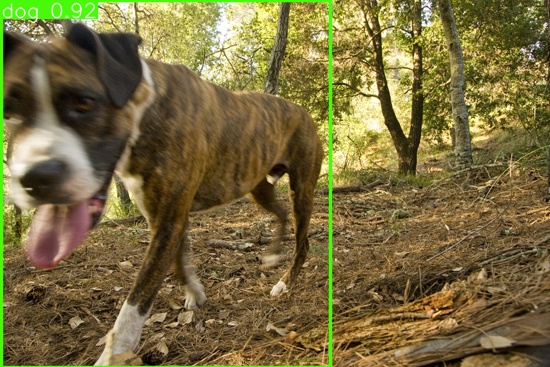}
    \end{minipage}
    \medskip
    \begin{minipage}{0.5\textwidth}
        \centering
        \includegraphics[width=\linewidth]{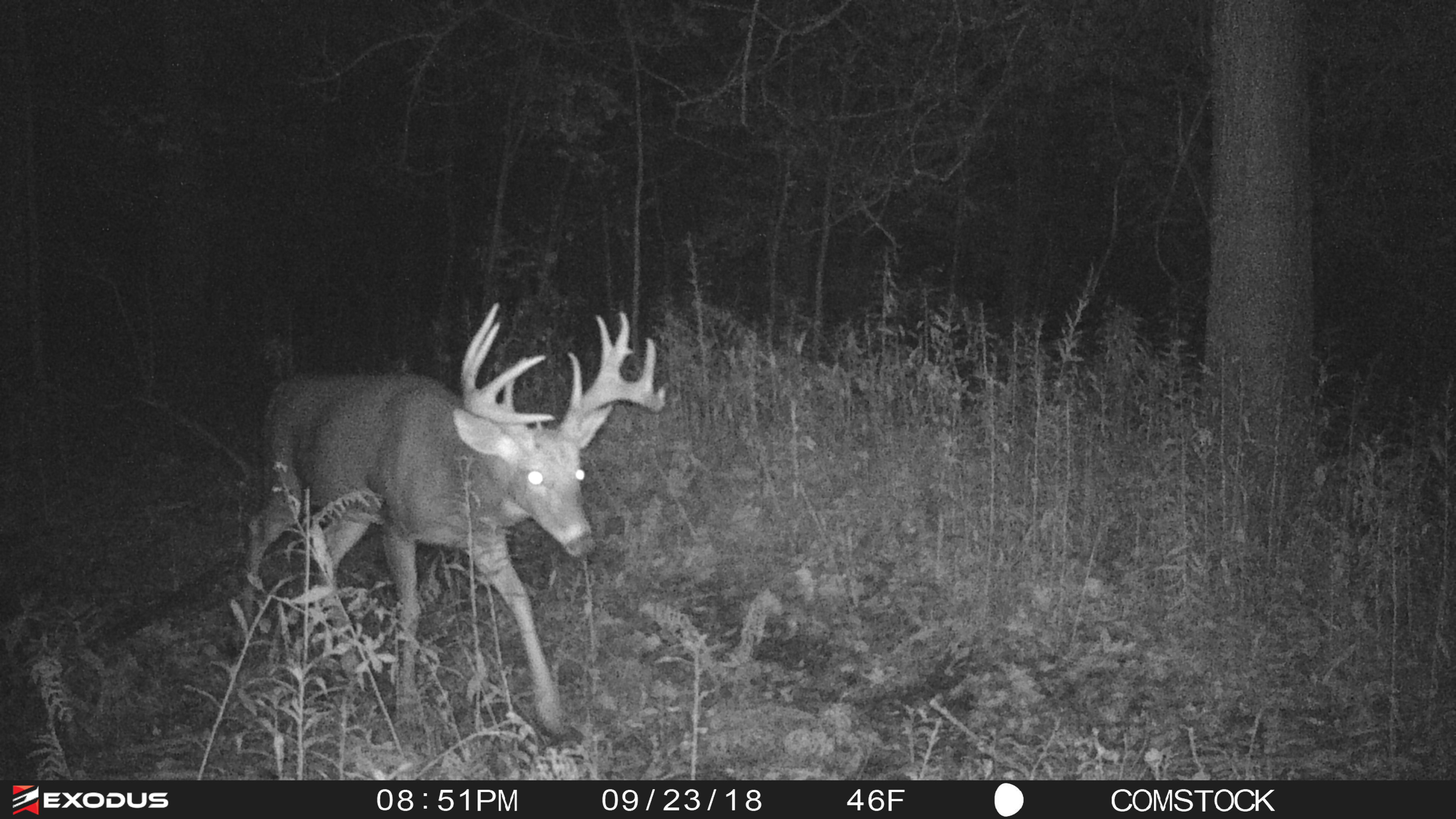}
    \end{minipage}%
    \begin{minipage}{0.5\textwidth}
        \centering
        \includegraphics[width=\linewidth]{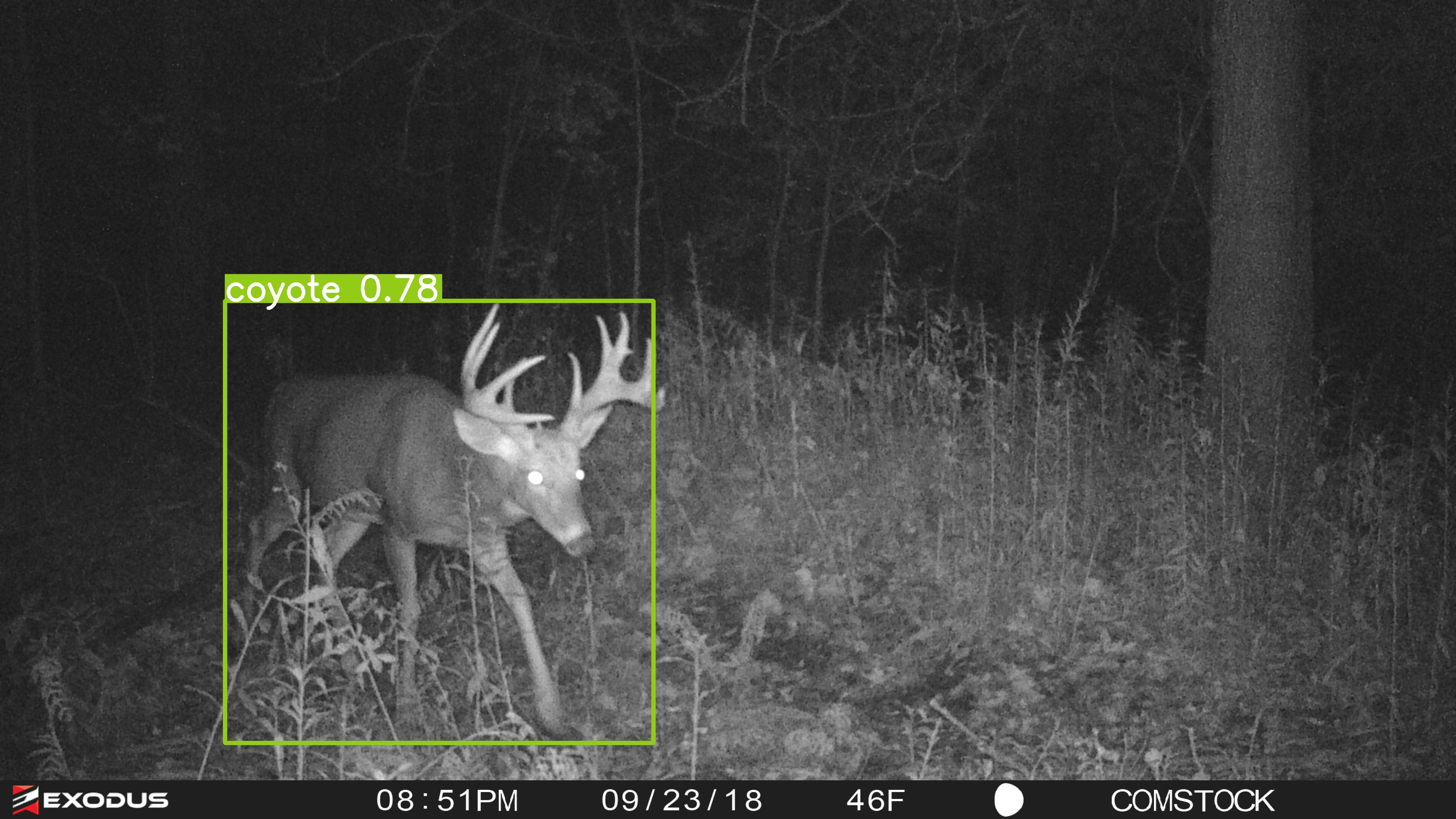}
    \end{minipage}
    
    \medskip
    \begin{minipage}{0.5\textwidth}
        \centering
        \includegraphics[width=\linewidth]{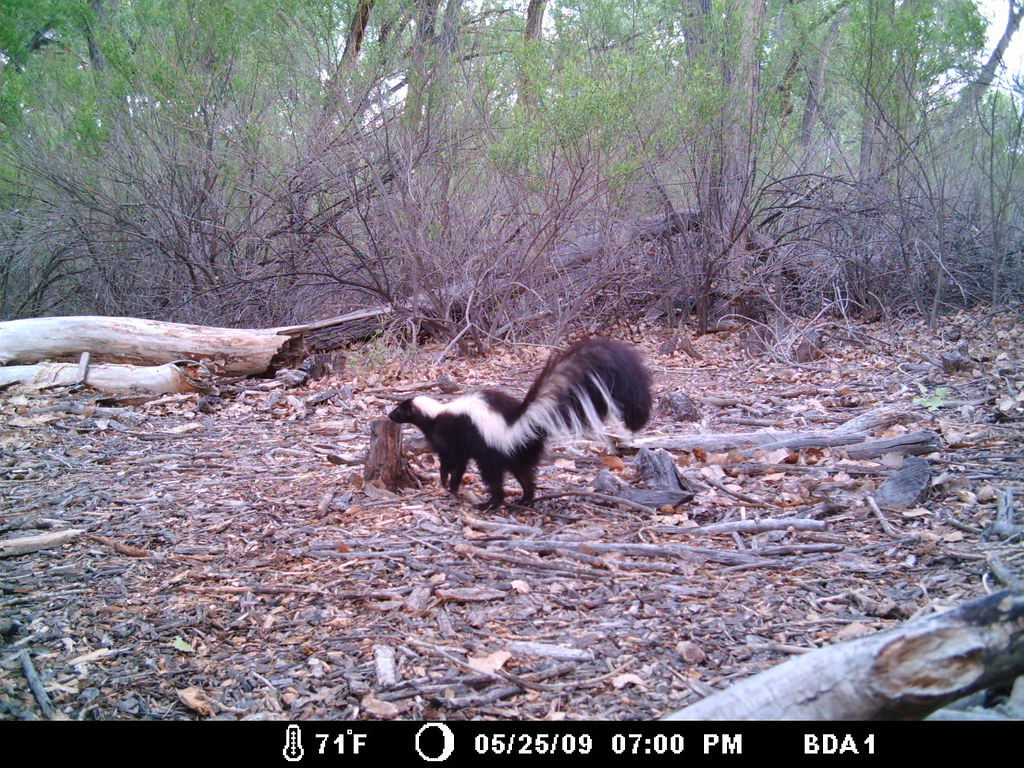}
    \end{minipage}%
    \begin{minipage}{0.5\textwidth}
        \centering
        \includegraphics[width=\linewidth]{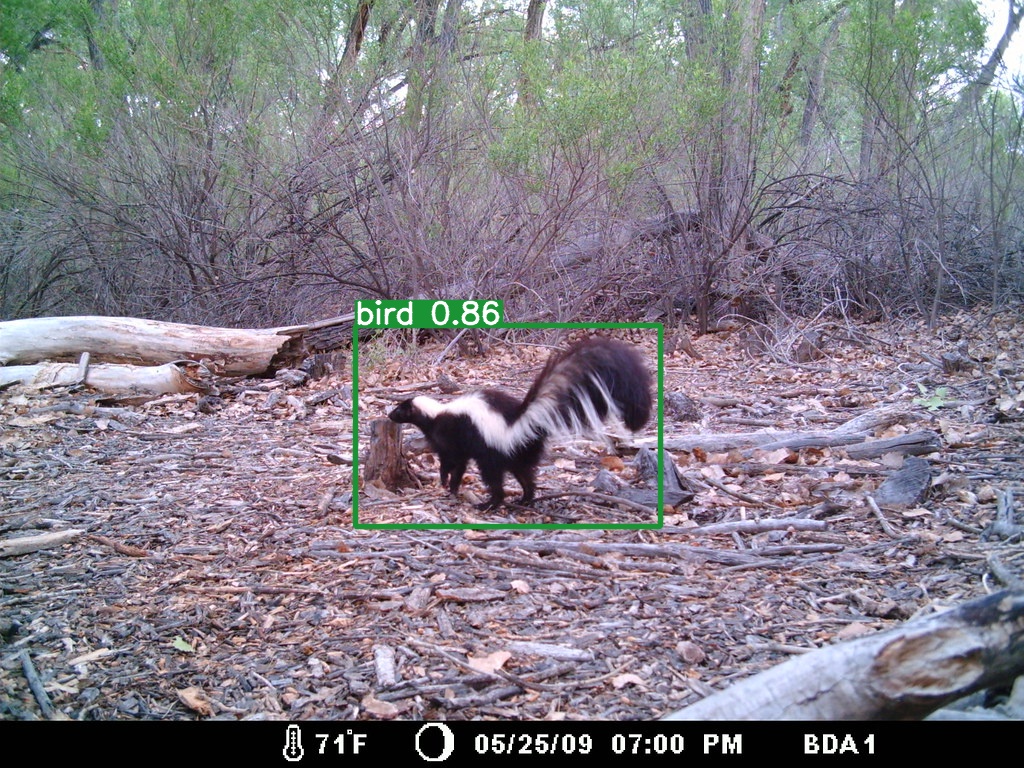}
    \end{minipage}
    \caption{Screenshots of predictions by improved YOLOv8s model on the custom camera trap dataset collected from the internet}
\end{figure}

The above collage of images and their corresponding prediction shows some of the correct predictions and incorrect predictions made by the improved trained model on the never-before-seen camera trap images collected from the internet.\\

In the context of the two inaccuracies highlighted in \ref{fig:trans-test-improved-model-custom}, I tested the performance of the trained baseline YOLOv8s model to determine its ability to make accurate predictions.
\begin{figure}[H]
   \begin{minipage}{0.5\textwidth}
        \centering
        \includegraphics[width=\linewidth]{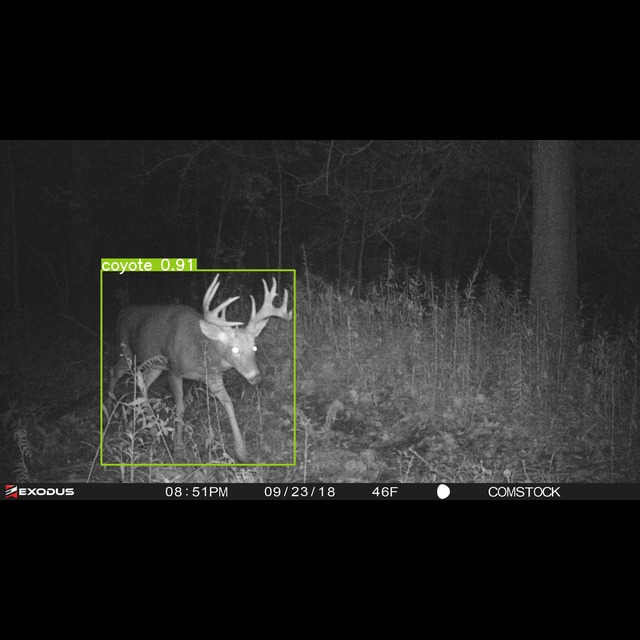}
    \end{minipage}%
    \begin{minipage}{0.5\textwidth}
        \centering
        \includegraphics[width=\linewidth]{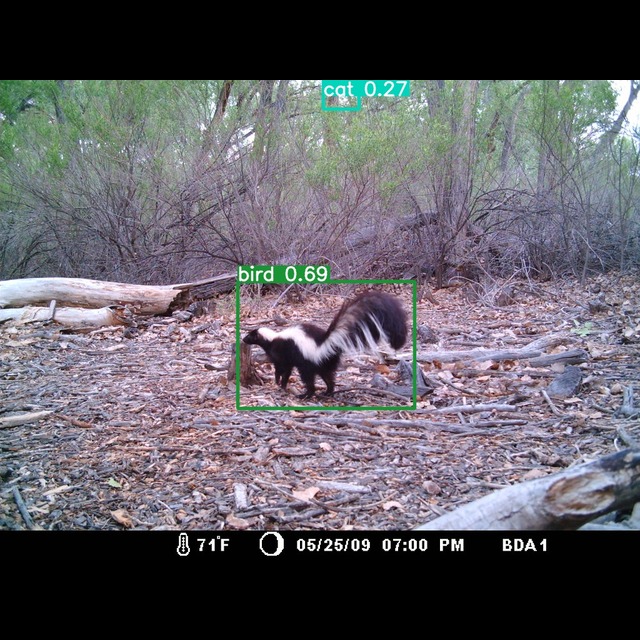}
    \end{minipage}
    \medskip
    \caption{Screenshots of predictions by baseline YOLOv8s model on the custom camera trap dataset collected from the internet which were incorrectly predicted by improved model}
\end{figure}
The trained model too was unable to make correct predictions for these two images. The above images contain a Deer and a Skunk respectively and have been classified incorrectly by both trained baseline YOLOv8s and the improved YOLOv8s model.\\

From the class distribution in Fig \ref{fig:train-class-distribution}, it is evident that the model has a lower number of images for these two categories so, the model hasn't been sufficiently trained to detect these object categories.  Insufficient representation of Deer and Skunk instances in the dataset has hindered the model's training for accurate identification of these labels. Therefore, both the baseline YOLOv8s model and the improved YOLOv8s model produce incorrect predictions for the images featuring the underrepresented categories in this case.
\chapter{Conclusion and Future Directions}

\section{Conclusion}
This thesis attempts to perform the comparative performance analysis between the baseline YOLOv8s and the improved YOLOv8s model in the context of object detection using a subset of the Caltech Camera Traps dataset, as outlined in the results and discussions section \ref{chap:results-and-analysis}. The baseline model demonstrated commendable efficacy on the validation set, while the baseline model with integrated WIoUv3 excelled on the Cis-Test set. Notably, the improved YOLOv8s model, augmented with both WIoUv3 and the Global Attention Mechanism (GAM) module, outperformed others on the Trans-Test dataset, showcasing its robust generalization capabilities. \\

The observed ability of the improved model to apply learned features to the never-before-seen dataset is a noteworthy highlight. This proficiency suggests a commendable adaptability, extending the model's utility beyond familiar training and validation sets. Such generalization capabilities are crucial for real-world applications, where the model must perform well across diverse and unexplored scenarios. \\

The findings discussed above indicate that the improved model holds promise for broader applicability, offering potential advantages in scenarios where adaptability and performance on novel datasets are paramount. However, it is essential to acknowledge the nuanced performance differences observed, such as the slight dip in the mAP50 evaluation metric for the Cis-Test location for the improved model. \\

In summary, the improved YOLOv8s model, with its integrated attention mechanism, modified multi-scale feature fusion, and an updated bounding box regression loss function presents a compelling case for its application in diverse and challenging environments. \\

\section{Future Directions}
It is pretty evident that there are several promising avenues for future research that can build upon the insights presented in this thesis study. Moving forward, researchers and practitioners alike can delve into the following key areas to expand their understanding and address pertinent questions in this field.

\begin{enumerate}
    \item Investigation of ensemble methods to leverage the strengths of different models by combining predictions from multiple models, including variations of YOLOv8s and other state-of-the-art object detection models that will possibly contribute to bridging the gap between the evaluation metric in the Cis-Test dataset and the Trans-Test dataset.

    \item Curation of a broad and comprehensive dataset that serves the purpose of testing the generalization capabilities and robustness of object detection models.

    \item Exploration of advanced transfer learning strategies, such as domain adaptation or few-shot learning, to further enhance the model’s ability to generalize to new and unseen environments.

\end{enumerate}

\bibliographystyle{plainnat}
\bibliography{References}

\appendix
\chapter{Code Repository}
In this appendix, you can find the link to the code repository for the thesis.\\
\\
The code for this thesis can be accessed at: \\
\url{https://github.com/arojsubedi/Improved-YOLOv8s}
\\ \\
Changes made to integrate each individual component discussed in this thesis can be found in the README file of the same repository and can be accessed at: \\
\url{https://github.com/arojsubedi/Improved-YOLOv8s/blob/main/README.md}
\\
\\
A subset of the processed CalTech Camera Traps dataset used in this project can be accessed at: \\
\url{https://github.com/arojsubedi/CCT20_Processed_Dataset}

\end{document}